\documentclass[times,letterpaper]{elsarticle}
\makeatletter
\def\ps@pprintTitle{%
 \let\@oddhead\@empty
 \let\@evenhead\@empty
 \def\@oddfoot{\thepage\hfill\footnotesize\itshape\today}
 \let\@evenfoot\@oddfoot}
\makeatother

\usepackage[margin=2.25cm]{geometry}
\usepackage[table]{xcolor}
\usepackage{mathtools}
\usepackage{graphicx}
\usepackage{amssymb}
\usepackage[labelfont=bf]{caption}
\usepackage{subcaption}
\usepackage{tcolorbox}
\usepackage{scrextend}
\usepackage{listings}
\usepackage{bm}
\usepackage{float}
\usepackage[colorlinks]{hyperref}
\usepackage{multirow}
\usepackage[linesnumbered,lined,boxed,ruled,vlined]{algorithm2e}
\usepackage{afterpage}
\usepackage{etoolbox}
\usepackage{dutchcal}
\usepackage{makecell}
\usepackage{cleveref}
\usepackage{enumitem}
\usepackage[makeroom,smaller]{cancel}
\usepackage{stmaryrd}
\usepackage{placeins}
\usepackage[pages=some,placement=top]{background}
\usepackage{lineno}

\backgroundsetup{
	scale=1,
	color=black,
	opacity=1.0,
	angle=0,
	contents={%
		\includegraphics[width=\paperwidth,height=2cm]{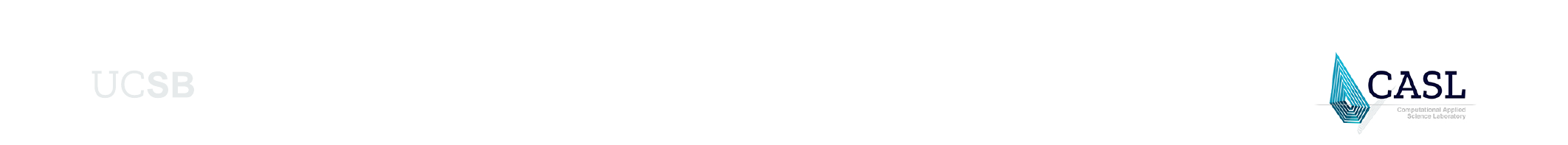}
	}%
}

\SetAlCapNameFnt{\small}	
\SetAlCapFnt{\small}
\SetAlFnt{\small}

\definecolor{navy}{RGB}{2, 48, 71}
\definecolor{aqua}{RGB}{4, 124, 145}
\definecolor{cloud0}{RGB}{249, 249, 249}
\definecolor{cloud1}{RGB}{230, 234, 235}
\definecolor{goodgreen}{RGB}{30, 135, 74}
\definecolor{darkcoral}{RGB}{196, 52, 36}
\definecolor{orange}{HTML}{FB8500}
\definecolor{bluematlab}{HTML}{0072BD}
\definecolor{yellowmatlab}{HTML}{EDB120}
\definecolor{redmatlab}{HTML}{A2142F}
\definecolor{orangematlab}{HTML}{D95319}
\definecolor{darkaqua}{RGB}{12, 103, 130}
\definecolor{darkgray}{RGB}{70, 70, 70}

\AtBeginDocument{%
\hypersetup{
	colorlinks=true,      		
	linkcolor=aqua,				
	citecolor=goodgreen,		
	filecolor=black,      		
	urlcolor=darkcoral,			
	bookmarks=false,
	pdffitwindow=true,
	pdfpagelayout=SinglePage,
	allcolors=aqua
}}

\newcommand{\colorsection}[1]{\sffamily\color{navy}\section{#1}\color{black}\rmfamily}
\newcommand{\colorsubsection}[1]{\sffamily\color{navy}\subsection{#1}\color{black}\rmfamily}
\newcommand{\colorsubsubsection}[1]{\sffamily\color{navy}\subsubsection{#1}\color{black}\rmfamily}

\patchcmd{\abstract}{Abstract}{\sffamily{Abstract}\rmfamily}{}{}
\abstracttitle{\sffamily\textcolor{navy}{Abstract}}

\newcommand{\etal}{et al.\ }

\newcommand{\vv}[1]{ \boldsymbol{\mathbf{#1}} }

\newcommand{\eten}[2]{ #1\times 10^{#2} }

\newcommand{\keyval}[2]{#1{:}\,{#2}}

\newcommand{\roundh}[1]{ \llparenthesis#1\rrparenthesis_h }

\let\OLDthebibliography\thebibliography
\renewcommand\thebibliography[1]{
  \OLDthebibliography{#1}
  \setlength{\parskip}{1pt}
  \setlength{\itemsep}{1pt plus 0.3ex}
}

\newcommand{\dist}[2]{\mathrm{dist}(#1, #2)}	
\newcommand{\qsn}{\mathrm{qsn}}					
\newcommand{\qhp}{\mathrm{qhp}}					
\newcommand{\elp}{\mathrm{elp}}					
\newcommand{\qgs}{\mathrm{qgs}}					
\newcommand{\qpr}{\mathrm{qpr}}					

\begin{document}

\title{\textcolor{aqua}{\sffamily\bfseries Machine learning algorithms for three-dimensional mean-curvature computation in the level-set method}}

\author[1]{\textcolor{gray}{Luis \'{A}ngel} Larios-C\'{a}rdenas\corref{cor1}}
\ead{lal@cs.ucsb.edu}

\author[1,2]{\textcolor{gray}{Fr\'{e}d\'{e}ric} Gibou}
\ead{fgibou@ucsb.edu}

\cortext[cor1]{Corresponding author}

\address[1]{\textcolor{darkgray}{Computer Science Department, University of California, Santa Barbara, CA 93106, USA}}
\address[2]{\textcolor{darkgray}{Mechanical Engineering Department, University of California, Santa Barbara, CA 93106, USA}}

\BgThispage


\begin{abstract}
We propose a data-driven mean-curvature solver for the level-set method.  This work is the natural extension to $\mathbb{R}^3$ of our two-dimensional error-correcting strategy presented in [Larios-C\'{a}rdenas and Gibou, \href{https://doi.org/10.1007/s10915-022-01952-2}{DOI: 10.1007/s10915-022-01952-2} (October 2022)] \cite{Larios;Gibou;KECNet2D;2022} and the hybrid inference system of [Larios-C\'{a}rdenas and Gibou, \href{https://doi.org/10.1016/j.jcp.2022.111291}{DOI: 10.1016/j.jcp.2022.111291} (August 2022)] \cite{Larios;Gibou;HybridCurvature;2021}.  However, in contrast to \cite{Larios;Gibou;HybridCurvature;2021, Larios;Gibou;KECNet2D;2022}, which built resolution-dependent dictionaries of curvature neural networks, here we develop a single pair of neural models in $\mathbb{R}^3$, regardless of the mesh size.  The core of our system comprises two feedforward neural networks.  These models ingest preprocessed level-set, gradient, and curvature information to fix numerical mean-curvature approximations for selected nodes along the interface.  To reduce the problem's complexity, we have used the local Gaussian curvature $\kappa_G$ to classify stencils and fit these networks separately to non-saddle (where $\kappa_G \gtrapprox 0$) and saddle (where $\kappa_G \lessapprox 0$) input patterns.  The Gaussian curvature is essential for enhancing generalization and simplifying neural network design.  For example, non-saddle stencils are easier to handle because they exhibit a mean-curvature error distribution characterized by monotonicity and symmetry.  While the latter has allowed us to train only on half the mean-curvature spectrum, the former has helped us blend the machine-learning-corrected output and the baseline estimation seamlessly around near-flat regions.  On the other hand, the saddle-pattern error structure is less clear; thus, we have exploited no latent characteristics beyond what is known.  Such a stencil distinction makes learning-tuple generation in $\mathbb{R}^3$ quite involving.  In this regard, we have trained our models on not only spherical but also sinusoidal and hyperbolic paraboloidal patches at various configurations.  Our approach to building their data sets is systematic but gleans samples randomly while ensuring well-balancedness.  Furthermore, we have resorted to standardization and dimensionality reduction as a preprocessing step and integrated layer-wise regularization to minimize outlying effects.  In addition, our strategy leverages mean-curvature rotation and reflection invariance to improve stability and precision at inference time.  The synergy among all these features has led to a performant mean-curvature solver that works for any grid resolution.  Experiments with several interfaces confirm that our proposed system can yield more accurate mean-curvature estimations in the $L^1$ and $L^\infty$ norms than modern particle-based interface reconstruction and level-set schemes around under-resolved regions.  Although there is ample room for improvement, this research already shows the potential of machine learning to remedy geometrical problems in well-established numerical methods.  Our neural networks are available online at \url{https://github.com/UCSB-CASL/Curvature_ECNet_3D}.
\end{abstract}


\maketitle


\colorsection{Introduction}
\label{sec:Introduction}

Curvature is a fundamental geometrical attribute related to minimal surfaces in differential geometry \cite{ModernDifferentialGeometry;2006} and surface tension in physics \cite{Osher1988, Popinet;NumModelsOfSurfTension;18}.  In particular, mean curvature plays a crucial role in solving free-boundary problems (FBP) \cite{Friedman10, CMGP16} with applications in multiphase flows \cite{Sussman;Smereka;Osher:94:A-Level-Set-Approach, Sussman;Fatemi;Smereka;etal:98:An-Improved-Level-Se, Gibou;Chen;Nguyen;etal:07:A-level-set-based-sh, Theillard:2019aa, Gibou:2019aa, Karnakov;etal;HybridParticleMthdVOFCuravture;2020, Egan;etal;DirNumSimIncompFlowsOctree;2021}, heat conduction and solidification \cite{Chen;Min;Gibou:09:A-numerical-scheme-f, Papac;Gibou;Ratsch:10:Efficient-symmetric-, Papac;Helgadottir;Ratsch;etal:13:A-level-set-approach, Mirzadeh;Gibou:14:A-conservative-discr, Theillard;Gibou;Pollock:14:A-Sharp-Computationa}, wildfire propagation \cite{Mallet;etal;ModelingWildlandFirePropagation;2009}, biological systems \cite{Macklin;Lowengrub;ImprovedCurvatureAppTumorGrowth;2006, Boudon;etal;3DPlantMorphogenesis;2015, Ocko;Heyde;Mahadevan;MorphTermiteMounds;2019, AliasBuenzli20},  and computer graphics and vision \cite{Chan;Vese:01:Active-Contour-Witho, Osher;Paragios:03:Geometric-Level-Set-, Losasso;Gibou;Fedkiw:04:Simulating-Water-and, Gibou;Levy;Cardenas;etal:05:Partial-Differential, Losasso:2006aa, LevelSetAsDeepRNN18, Zhang;ConstRobotImgSegPDEs;2022}.  In surface-tension modeling, for example, computing curvature accurately at the interface is essential for recovering specific equilibrium solutions to the FBP's partial differential equations (PDE) \cite{Popinet;NumModelsOfSurfTension;18}.  Well-balanced numerical schemes that do so should also work appropriately near regions of topological change to avoid erroneous pressure jumps that may affect breakup and coalescence \cite{Macklin;Lowengrub;ImprovedCurvatureAppTumorGrowth;2006, Lervag;CalcCurvatureLSM;2014, Ervik;Lervag;Munkejord;LOLEX;2014}.

Over the years, the research community has devised several numerical procedures for solving FBPs efficiently.  Among them, the level-set method \cite{Osher1988, Osher;Fedkiw:02:Level-Set-Methods-an, GFO18} has emerged as a popular choice for its simplicity and ability to capture and transport deformable surfaces.  The level-set framework belongs to the family of Eulerian formulations, alongside the phase-field \cite{QB10} and volume-of-fluid (VOF) schemes \cite{Hirt;Nichols:81:Volume-of-Fluid-VOF-}.  Unlike Lagrangian front-tracking methods \cite{Tryggvason;Bunner;Esmaeeli;etal:01:A-Front-Tracking-Met, Bo;Liu;Glimm;etal:11:A-robust-front-track}, they all represent free boundaries implicitly with higher-dimensional functions.  Thus, Eulerian formulations can handle complex topological changes naturally, with no need for overcomplicated interface-reconstruction (IR) algorithms \cite{Osher;Fedkiw:02:Level-Set-Methods-an}. 

The standard curvature-estimating method in the level-set framework is straightforward.  Yet, level-set field irregularities often undermine its accuracy, especially around under-resolved regions and near discontinuities  \cite{Macklin;Lowengrub;ImprovedCurvatureAppTumorGrowth;2006, Chene;Min;Gibou:08:Second-order-accurat, Lervag;CalcCurvatureLSM;2014, Ervik;Lervag;Munkejord;LOLEX;2014}.  To remedy these flaws, computational scientists have introduced different approaches.  In \cite{Chene;Min;Gibou:08:Second-order-accurat}, for example, du Ch\'{e}n\'{e}, Min, and Gibou have developed high-order reinitialization schemes that yield second-order accurate curvature computations on uniform and adaptive Cartesian grids.  These schemes extend the idea of redistancing the level-set field with iterative algorithms until it resembles a signed distance function \cite{Sussman;Smereka;Osher:94:A-Level-Set-Approach}.  When executed correctly, not only can reinitialization restore level-set smoothness, but it can also dampen the critical problem of mass loss \cite{Russo;Smereka:00:A-remark-on-computin}.  These efforts, however, do not always guarantee satisfactory results on coarse discretizations.  And, although we can refine the mesh next to the free boundary with adaptive grids efficiently \cite{Strain1999, Mirzadeh;etal:16:Parallel-level-set}, this only delays the onset of under-resolution issues.  Moreover, there are FBPs, such as high-$Re$ turbulent flows \cite{Pathak;etal;MLToAugCoarseGridCFD;2020, Egan;etal;DirNumSimIncompFlowsOctree;2021}, for which increasing the resolution in $\mathbb{R}^3$ is prohibitive even on multicore computing systems.  Here, we put forward a data-driven solution constructed on the above numerical schemes.  Nevertheless, our goal is to attain high mean-curvature precision around under-resolved regions while short-cutting the need for further mesh refinement.

Other practitioners have adopted a geometrical approach to reduce mean-curvature errors when handling level-set discontinuities.  Among them, we can find the schemes of Macklin and Lowengrub \cite{Macklin;Lowengrub;ImprovedCurvatureAppTumorGrowth;2006}.  They have approximated two-dimensional interfaces with least-squares quadratic polynomials and recalculated curvature on local sub-grids near level-set singularities (i.e., around kinks in the underlying level-set function emerging from free-boundary collisions).  Similarly, Lerv\r{a}g \cite{Lervag;CalcCurvatureLSM;2014} has devised a method to parametrize interfaces with Hermite splines and estimate curvature, spawning no additional sub-cell stencils.  While accurate, these systems are difficult to extend to three-dimensional space because of their surface reconstruction requirement.  However, their topology-aware strategy has motivated the emergence of sophisticated yet easy-to-integrate non-curve-fitting alternatives derived from insertion/extraction algorithms \cite{Salac;Lu;LocalSemiImpliciteLSM;2008}.  In this context, technologies, such as the local level-set extraction method (LOLEX) \cite{Ervik;Lervag;Munkejord;LOLEX;2014}, have recently outperformed any of their predecessors when computing geometrical attributes in extreme interfacial conditions (e.g., involving self-folding bodies).  In the same spirit as LOLEX, here, we have designed a mean-curvature solver that researchers can readily incorporate into existing level-set codebases.  Its inner workings are optimized to leverage distance and other geometrical information to model and fix errors in finite-difference-based mean-curvature estimations.

In this manuscript, we propose to address the level-set's curvature shortcomings from a data-driven perspective.  The latest developments of Qi \etal \cite{CurvatureML19} and Patel \etal \cite{VOFCurvature3DML19}, in particular, have inspired our machine- and deep-learning incursions for solving not only for mean-curvature \cite{LALariosFGibou;LSCurvatureML;2021, Larios;Gibou;HybridCurvature;2021, Larios;Gibou;KECNet2D;2022} but also passive transport \cite{Larios;Gibou;ECNetSemiLagrangian;2021}.  In the pioneering study of Qi \etal \cite{CurvatureML19}, the researchers fitted two-layered perceptrons to circular-interface samples to estimate curvature from area fractions in the VOF representation.  Later, in \cite{VOFCurvature3DML19}, Patel \etal extended this idea and trained mean-curvature feedforward networks with randomized spherical patches in three-dimensional VOF problems.  In both cases, their approaches showed promising, accurate results in two-phase flow simulations.  Like these, there have been other contemporary deep learning implementations devoted to solving geometrical difficulties in conventional numerical schemes.  These include, for instance, the curvature prediction from angular and normal-vector information \cite{Franca;Oishi;MLCurvatureFrontTracking;2022} in front-tracking and the neural calculation of linear piecewise interface construction (PLIC \cite{Youngs;VOF-PLIC;1982}) coefficients \cite{NPLIC20} in the VOF method.  Likewise, one can find advancements in the level-set framework, such as Buhendwa, Bezgin, and Adams' consistent algorithms for inferring area fractions and apertures for IR \cite{Buhendwa;Bezgin;Adams;IRinLSwithML;2021}.

Our research incorporates the notion of neural error quantification first introduced in computational fluid dynamics to fix artificial numerical diffusion and augment coarse-grid simulations \cite{Pathak;etal;MLToAugCoarseGridCFD;2020, Larios;Gibou;ECNetSemiLagrangian;2021}.  More importantly, our study is the natural extension to $\mathbb{R}^3$ of the preliminary two-dimensional error-correcting strategy presented in \cite{Larios;Gibou;KECNet2D;2022}.  In previous work, we optimized simple multilayer perceptrons (MLP) \cite{A18} with samples from circular and sinusoidal zero-isocontours to offset numerical curvature errors at the interface.  Then, we blended the neurally corrected and the baseline approximations selectively to improve precision around under-resolved regions and near steep\footnote{In this work, we use the term \textit{steep} to qualify a surface with regions where the dimensionless mean curvature is much greater than 0.1.} interface transitions.  Compared to a network-only approach \cite{LALariosFGibou;LSCurvatureML;2021} and an earlier hybrid inference system \cite{Larios;Gibou;HybridCurvature;2021}, error correction has proven more effective at closing the gap between the expected curvature and its numerical estimation.  Indeed, beyond-usual contextual information (e.g., level-set values, gradients, and the curvature itself), dimensionality reduction, and regularization have been important factors in achieving the results reported in \cite{Larios;Gibou;KECNet2D;2022}.

Here, we propose a hybrid solver to estimate mean-curvature in three-dimensional free-boundary problems.  At the core of our system, we have two shallow feedforward neural networks.  These models ingest preprocessed level-set, normal-vector, and curvature information to fix numerical mean-curvature approximations selectively along the interface.  To reduce the problem's complexity, we have used the local Gaussian curvature \cite{ModernDifferentialGeometry;2006}, $\kappa_G$, as an inexpensive stencil classifier and fitted these networks separately to non-saddle (where $\kappa_G \gtrapprox 0$) and saddle (where $\kappa_G \lessapprox 0$) input patterns.  Unlike \cite{VOFCurvature3DML19}, Gaussian curvature is essential in our strategy for enhancing generalization and simplifying neural architecture and design.  Non-saddle stencils, for example, are easier to handle because they exhibit a mean-curvature error distribution characterized by monotonicity and symmetry.  While the latter has allowed us to train only on half the mean-curvature spectrum, the former has helped us blend the machine-learning-corrected output and the baseline estimation seamlessly around near-flat regions.  On the other hand, the saddle-pattern error structure is less clear; thus, we have exploited no latent characteristics beyond what is known.  Such a distinction between stencils makes learning-tuple generation in $\mathbb{R}^3$ more involving than its planar counterpart \cite{Larios;Gibou;KECNet2D;2022}.  In this regard, we have trained our models on not only spherical but also sinusoidal and hyperbolic paraboloidal patches at various configurations.  Our approach to building their corresponding data sets is systematic but gleans samples in a randomized fashion while ensuring well-balancedness as much as possible.  In the same spirit as \cite{Larios;Gibou;ECNetSemiLagrangian;2021, Larios;Gibou;KECNet2D;2022}, we have resorted to standardization and dimensionality reduction as a preprocessing step and integrated layer-wise regularization to minimize outlying effects.  In addition, our strategy leverages mean-curvature rotation and reflection invariance, as in \cite{Buhendwa;Bezgin;Adams;IRinLSwithML;2021}, to improve stability and precision at inference time.  The synergy among all these features has led to a performant mean-curvature solver that works for any grid resolution.  And there is more to it.  Experiments with several geometries have confirmed that our proposed system yields more accurate mean-curvature estimations in the $L^1$ and $L^\infty$ norms than modern particle-based IR \cite{Karnakov;etal;HybridParticleMthdVOFCuravture;2020} and level-set schemes around under-resolved regions.  Although there is ample room for improvement, this research already shows the potential of machine learning and neural networks to remedy geometrical problems in well-established numerical methods.  Our results below and those given in preliminary studies \cite{Pathak;etal;MLToAugCoarseGridCFD;2020, Buhendwa;Bezgin;Adams;IRinLSwithML;2021, Larios;Gibou;ECNetSemiLagrangian;2021, Larios;Gibou;KECNet2D;2022} suggest that data-driven solutions perform better when they supplement conventional frameworks rather than when they try to replace them.

We have organized the contents of this manuscript into five sections.  First, \Cref{sec:TheLevelSetMethod,sec:AdaptiveCartesianGrids} summarize the level-set method and adaptive Cartesian grids, which are the only way of handling FBPs in $\mathbb{R}^3$ efficiently.  Then, \Cref{sec:Methodology} describes all the aspects related to our proposed hybrid inference system and the algorithms to generate its learning data and deploy its components.  After that, we evaluate our strategy through a series of experiments in \Cref{sec:Results}. Finally, we reflect on our findings and outline some future work in \Cref{sec:Conclusions}.


\FloatBarrier
\colorsection{The level-set method}
\label{sec:TheLevelSetMethod}

The level-set method \cite{Osher1988} is an implicit formulation for capturing and evolving interfaces represented as the zero-isocontour of a high-dimensional, Lipschitz relation $\phi(\vv{x}, t): \mathbb{R}^{n+1} \mapsto \mathbb{R}$ known as the \textit{level-set function}.  In this framework, we denote an interface by $\Gamma(t) \doteq \{\vv{x} : \phi(\vv{x}, t) = 0\}$, which partitions the computational domain into $\Omega^-(t) \doteq \{\vv{x} : \phi(\vv{x}, t) < 0\}$ and $\Omega^+(t) = \{\vv{x} : \phi(\vv{x}, t) > 0\}$\footnote{For compactness, we will omit functional dependence on time in the rest of this paper.}.  Deforming and convecting $\Gamma$ in level-set schemes require solving a Hamilton--Jacobi PDE known as the \textit{level-set equation},

\begin{equation}
\phi_t(\vv{x}) + \vv{u}(\vv{x})\cdot\nabla\phi(\vv{x}) = 0,
\label{eq:LevelSetEquation}
\end{equation}
where $\vv{u}(\vv{x}) : \mathbb{R}^n \mapsto \mathbb{R}^n$ is an application-dependent velocity field.  Traditionally, we solve \cref{eq:LevelSetEquation} with high-order accurate schemes, such as HJ-WENO \cite{Jiang;Peng:00:Weighted-ENO-Schemes}, or with unconditionally stable semi-Lagrangian approaches \cite{Courant;Isaacson;Rees:52:On-the-Solution-of-N, Wiin-Nielsen;SemiLagrangian;1958, Strain1999} in passive-transport FBPs.

The implicit nature of the level-set framework makes it straightforward to calculate normal vectors and mean curvature values at virtually any point $\vv{x} \in \Omega \subseteq \mathbb{R}^n$.  We define both attributes with

\begin{equation}
\hat{\vv{n}}(\vv{x}) \doteq \frac{\nabla\phi(\vv{x})}{\|\nabla\phi(\vv{x})\|} \quad \textrm{and} \quad 
\kappa(\vv{x}) \doteq \frac{1}{n-1} \nabla\cdot\frac{\nabla\phi(\vv{x})}{\|\nabla\phi(\vv{x})\|},
\label{eq:NormalAndMeanCurvature}
\end{equation}
where the latter can be expanded for the two- and three-dimensional cases \cite{Osher;Fedkiw:02:Level-Set-Methods-an,CurvatureImplicitSurfaces;Albin-etal;2017} as

\begin{subequations}
\begin{align}
\kappa(\vv{x} \in \mathbb{R}^2) &\equiv \frac{\phi_x^2 \phi_{yy} - 2\phi_x\phi_y\phi_{xy} + \phi_y^2 \phi_{xx}}{(\phi_x^2 + \phi_y^2)^{3/2}} \quad \textrm{and} \label{eq:MeanCurvature.2d} \\
\kappa(\vv{x} \in \mathbb{R}^3) &\equiv \frac{\phi_x^2(\phi_{yy}+\phi_{zz}) + \phi_y^2(\phi_{xx}+\phi_{zz}) + \phi_z^2(\phi_{xx}+\phi_{yy})}{2(\phi_x^2 + \phi_y^2 + \phi_z^2)^{3/2}} - \frac{\phi_x\phi_y\phi_{xy} + \phi_x\phi_z\phi_{xz} + \phi_y\phi_z\phi_{yz}}{(\phi_x^2 + \phi_y^2 + \phi_z^2)^{3/2}} \label{eq:MeanCurvature.3d}.
\end{align}
\label{eq:MeanCurvature}
\end{subequations}

\begin{figure}[!t]
	\centering
	\includegraphics[width=7.5cm]{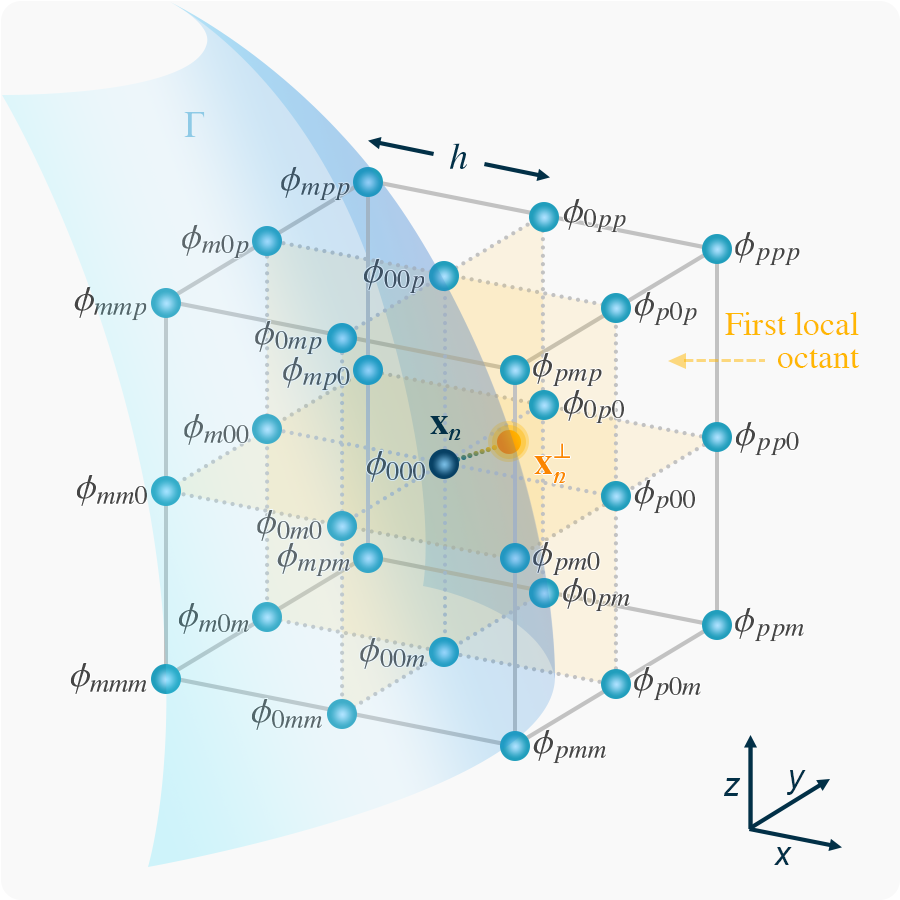}
	\caption{The $h$-uniform, 27-point stencil of an interface node $\mathcal{n}$ located at $\vv{x}_\mathcal{n}$ (\textcolor{navy}{$\bullet$}) and its normal projection $\vv{x}_\mathcal{n}^\perp$ (\textcolor{orange}{$\bullet$}) onto $\Gamma$.  Each stencil point holds its level-set value $\phi_{ijk}$.  The subscripts $i, j, k \in \{m,0,p\}$ help enumerate the nodes, starting from the left-back-bottommost corner.  (Color online.)}
	\label{fig:Stencil}
\end{figure}

Likewise, one can compute the Gaussian curvature at any $\vv{x} \in \Omega \subseteq \mathbb{R}^3$ using Goldman's derivation \cite{CurvatureFormulasForImplicit;Goldman;2005} and the extended relation for implicit surfaces provided by \cite{CurvatureImplicitSurfaces;Albin-etal;2017} as

\begin{subequations}
\begin{align}
\kappa_G(\vv{x}) &\doteq \frac{\nabla\phi(\vv{x})^T H^*(\phi(\vv{x})) \nabla\phi(\vv{x})}{\|\nabla\phi(\vv{x})\|^4} \label{eq:GaussianCurvature.Compact} \\
\begin{split}
				 &\equiv 2\left(\frac{\phi_x\phi_y(\phi_{xz}\phi_{yz}-\phi_{xy}\phi_{zz}) + \phi_x\phi_z(\phi_{xy}\phi_{yz}-\phi_{xz}\phi_{yy}) + \phi_y\phi_z(\phi_{xy}\phi_{xz}-\phi_{yz}\phi_{xx})}{(\phi_x^2 + \phi_y^2 + \phi_z^2)^{3/2}}\right) \\
				 &\qquad+ \frac{\phi_x^2(\phi_{yy}\phi_{zz}-\phi_{yz}^2) + \phi_y^2(\phi_{xx}\phi_{zz}-\phi_{xz}^2) + \phi_z^2(\phi_{xx}\phi_{yy}-\phi_{xy}^2)}{(\phi_x^2 + \phi_y^2 + \phi_z^2)^{3/2}},
\end{split} \label{eq:GaussianCurvature.Ext}
\end{align}
\label{eq:GaussianCurvature}
\end{subequations}
where $H^*(\phi)$ is the adjoint of $\phi(\vv{x})$'s Hessian matrix.  In practice, we employ stencils of level-set values (see \cref{fig:Stencil}) and first- and second-order central differences to discretize \cref{eq:MeanCurvature,eq:GaussianCurvature.Ext} and approximate the gradient in \cref{eq:NormalAndMeanCurvature} \cite{Min;Gibou:07:A-second-order-accur, Chene;Min;Gibou:08:Second-order-accurat}.  \Cref{eq:GaussianCurvature.Ext} is critical for our current research since it allows us to characterize grid points\footnote{We use the terms \textit{grid point} and \textit{node} interchangeably to refer to the discrete locations in a Eulerian mesh.} for downstream regression.

Also, in this work, we consider the geometrical interpretation of \cref{eq:MeanCurvature,eq:GaussianCurvature} given by

\begin{equation}
\kappa = \tfrac{1}{2}(\kappa_1 + \kappa_2) \quad \textrm{and} \quad \kappa_G = \kappa_1 \kappa_2,
\label{eq:PrincipalCurvatures}
\end{equation}
where $\kappa_1$ and $\kappa_2$ are the principal curvatures \cite{ModernDifferentialGeometry;2006}.

Curvature computations in the level-set representation are often noisy because of the second derivatives in \cref{eq:MeanCurvature,eq:GaussianCurvature.Ext}.  However, it is possible to estimate $\kappa$ and $\kappa_G$ accurately if $\phi(\vv{x})$ guarantees sufficient smoothness and regularity \cite{Chene;Min;Gibou:08:Second-order-accurat}.  On this point, signed-distance functions are an excellent choice because they are beneficial not only for curvature calculations but also for mass preservation.  Unfortunately, even if one chooses $\phi(\vv{x})$ as a signed-distance function at the onset, we cannot ensure that solving \cref{eq:LevelSetEquation} under an arbitrary velocity field will preserve this feature throughout the simulation.  For this reason, it is typical to redistance $\phi(\vv{x})$ periodically by solving the pseudo-transient \textit{reinitialization equation} \cite{Sussman;Smereka;Osher:94:A-Level-Set-Approach}

\begin{equation}
\phi_\tau(\vv{x}) + \texttt{sgn}(\phi^0(\vv{x}))(\|\nabla\phi(\vv{x})\| - 1) = 0
\label{eq:Reinitialization}
\end{equation}
to a steady state (e.g., for $\nu$ iterations) using a Godunov spatial discretization and high-order temporal TVD Runge--Kutta schemes \cite{Shu;Osher:88:Efficient-Implementa, Shu;Osher:89:Efficient-Implementa}.  In \cref{eq:Reinitialization}, $\tau$ represents fictitious time, $\texttt{sgn}(\cdot)$ is a smoothed-out signum function, and $\phi^0(\vv{x})$ is the arbitrary level-set function before the reshaping process.  Furthermore, when $\phi(\vv{x})$ resembles a signed-distance function, $\|\nabla\phi\| \approx 1$, several calculations simplify, and one can produce more robust estimates of \crefrange{eq:NormalAndMeanCurvature}{eq:GaussianCurvature} \cite{Chene;Min;Gibou:08:Second-order-accurat}.  Similarly, preliminary research \cite{LALariosFGibou;LSCurvatureML;2021, Larios;Gibou;HybridCurvature;2021, Larios;Gibou;KECNet2D;2022} has shown that redistancing can help remove unstructured patterns that may undermine the performance of deep learning curvature models in $\mathbb{R}^2$.

In this study, we have resorted to the level-set algorithms developed by Du Ch\'{e}n\'{e}, Min, and Gibou \cite{Min;Gibou:07:A-second-order-accur} and their parallel implementation on {\tt p4est} \cite{Burstedde;Wilcox;Ghattas:11:p4est:-Scalable-Algo}, as described in \cite{Mirzadeh;etal:16:Parallel-level-set}.  For a more comprehensive analysis, we refer the interested reader to the level-set texts by Osher, Fedkiw, and Sethian \cite{Sethian:99:Level-set-methods-an, Osher;Fedkiw:02:Level-Set-Methods-an} and the recent review by Gibou \etal \cite{GFO18}.


\FloatBarrier
\colorsection{Adaptive Cartesian grids}
\label{sec:AdaptiveCartesianGrids}

Solving FBPs on uniform grids is particularly expensive when the underlying phenomena exhibit a wide range of length scales.  High-$Re$ channel-flow simulations \cite{Egan;etal;DirNumSimIncompFlowsOctree;2021}, for example, require a high grid resolution along the boundary layers to capture wake vortices and other flow characteristics.  In such cases, working with adaptive Cartesian grids is more cost-effective because we only refine the mesh next to $\Gamma$, where accuracy is needed the most \cite{Strain1999}.

\begin{figure}[!t]
	\centering
	\includegraphics[width=16cm]{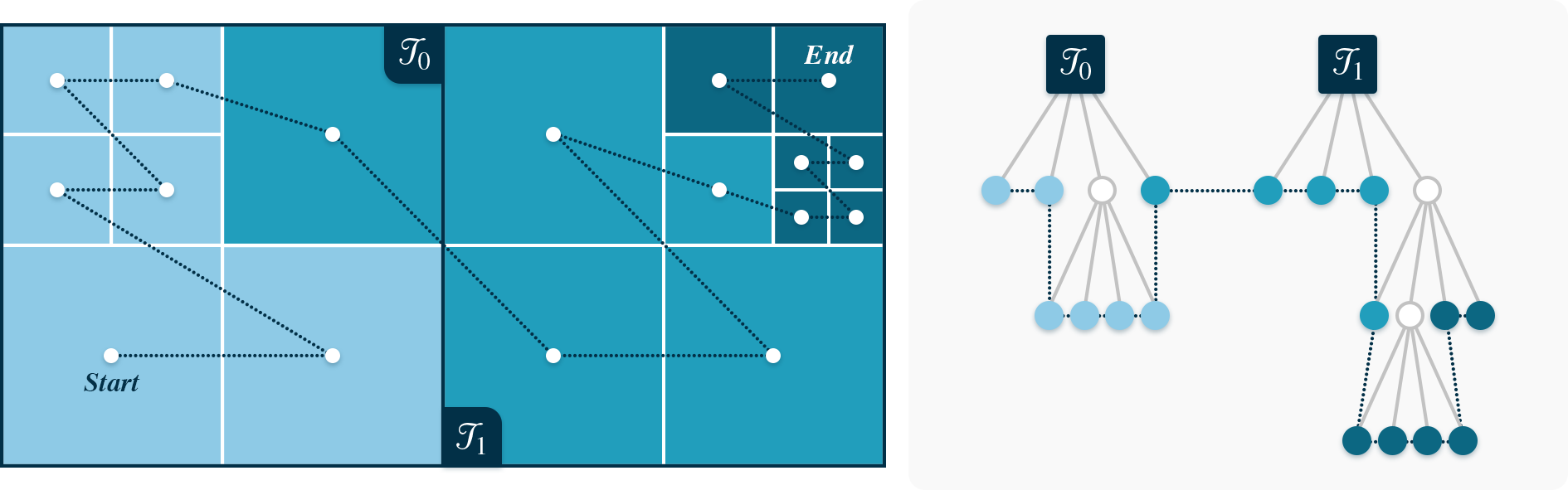}
	\caption{A macromesh $\mathcal{G}$ with two square quadtrees: $\mathcal{T}_0$ and $\mathcal{T}_1$.  Also displayed is a $z$-curve (dotted line) that joins the quadtree leaves from \textbf{\textit{start}} to \textbf{\textit{end}}.  This $z$-curve helps partition $\mathcal{G}$ across three processes for parallel computations.  The cells assigned to different processes appear in different shades of blue with white dots at the cell centers.  The grid points lie, essentially, at the \textit{junctions} between neighboring cells.  On the right, we illustrate the tree data structures, where $\mathcal{T}_0$ and $\mathcal{T}_1$ have maximum levels of refinement $\eta_0 = 2$ and $\eta_1 = 3$.  Adapted from \cite{Mirzadeh;etal:16:Parallel-level-set}.  (Color online.)}
	\label{fig:Forest}
\end{figure}

In this context, quadtree- ($\mathbb{R}^2$) and octree-based ($\mathbb{R}^3$) Cartesian grids have grown popular among level-set practitioners because they are easy to implement with signed distance functions \cite{Strain1999, Min;Gibou:07:A-second-order-accur, Mirzadeh;etal:16:Parallel-level-set}.  Quad/octrees are rooted data structures covering a region $B \subseteq \Omega$.  They are composed of cells organized recursively into levels, where each cell $C$ either has four (eight) children (i.e., quadrants (octants)) or is a leaf \cite{BKOS00}.  Once constructed, a quad/octree with maximum depth $\eta \geqslant 0$ can support several operations efficiently, such as searching in $\mathcal{O}(\eta)$ and sorting \cite{Strain1999}.  \Cref{fig:Forest} illustrates two squared quadtrees, where $\mathcal{T}_0$ and $\mathcal{T}_1$ have $\eta_0=2$ and $\eta_1=3$ maximum levels of refinement.  \Cref{fig:Forest} also describes a straightforward characterization of a domain's discretization in $\mathbb{R}^2$ from the {\tt p4est} library's perspective.  

The {\tt p4est} framework \cite{Burstedde;Wilcox;Ghattas:11:p4est:-Scalable-Algo} is a family of parallel algorithms for grid management.  This library provides scalable methods for parallelizing, partitioning, and load-balancing spatial tree data structures across heterogeneous computing systems.  Recently, {\tt p4est} has been shown to handle up to 458,752 cores \cite{Isaac;etal;RecursiveAlgDistForestOctrees;2015} in various resource-intensive tasks where working with the memory and processing power of a single computing node is unfeasible (e.g., see \cite{Egan;etal;DirNumSimIncompFlowsOctree;2021}).  In {\tt p4est}, we build an adaptive Cartesian grid $\mathcal{G}$ with non-overlapping trees rooted in a coarse scaffolding known as the \textit{macromesh}.  This coarse grid can adopt different configurations, but we always assume it to be small enough to be replicated on every process.  For simplicity, here, we have limited ourselves to uniform brick-like macromesh layouts, as depicted on the left diagram of \cref{fig:Forest}.

Each tree in a {\tt p4est} macromesh first undergoes coarsening and refinement based, for instance, on the interface's location.  In this study, we have opted for Min's extension \cite{Min2004} to the Whitney decomposition criteria \cite{Strain1999}, as presented in \cite{Min;Gibou:07:A-second-order-accur, Mirzadeh;etal:16:Parallel-level-set}, to guide the discretization process.  In particular, we mark a cell $C \in \mathcal{G}$ for refinement if the criterion

\begin{equation}
\min_{v\in \textrm{vertices}(C)}|\phi(v)| \leqslant \tfrac{1}{2}\textrm{Lip}(\phi)\cdot \texttt{diag}(C)
\label{eq:RefinementCriterion}
\end{equation}
holds, where $v$ is a cell vertex (i.e., a grid node), $\textrm{Lip}(\phi)$ is $\phi(\cdot)$'s Lipschitz constant (set to $1.2$ across $\Omega$), and $\texttt{diag}(C)$ is $C$'s diagonal.  Conversely, we tag a cell for coarsening if the condition

\begin{equation}
\min_{v\in \textrm{vertices}(C)}|\phi(v)| > \textrm{Lip}(\phi)\cdot \texttt{diag}(C)
\label{eq:CoarseningCriterion}
\end{equation}
is valid.  Besides these criteria, we can enforce a uniform band of $b$ smallest cells around $\Gamma$ by subdividing quadrants (octants) for which the inequality

\begin{equation}
\min_{v\in \textrm{vertices}(C)}|\phi(v)| < b\max( \Delta x_{\textrm{finest}},\, \Delta y_{\textrm{finest}},\, \Delta z_{\textrm{finest}} )
\label{eq:UniformBandCriterion}
\end{equation}
is true.  Our research considers only cubic cells to reduce the number of degrees of freedom in the problem at hand.  Hence, $h \equiv \Delta x_{\textrm{finest}} = \Delta y_{\textrm{finest}} = \Delta z_{\textrm{finest}}$ in \cref{eq:UniformBandCriterion} (see \cref{fig:Stencil}), where $h \doteq 2^{-\eta}$ is $\mathcal{G}$ 's mesh size, and $\eta$ is the maximum depth of the unit-hexahedron octrees in $\mathcal{G}$.

After fitting $\mathcal{G}$ to $\Gamma$ using \crefrange{eq:RefinementCriterion}{eq:UniformBandCriterion}, {\tt p4est} linearizes the trees using a $z$-curve that traverses their leaf cells, as shown in \cref{fig:Forest}.  Such a space-filling curve \cite{Aluru;Svilgen;ParallelDomDecomp;1997} helps partition and load-balance $\mathcal{G}$ among processes while ensuring that leaves with close $z$-indices remain close to each other, on average.  In practice, this property is beneficial because it reduces {\tt MPI} communications \cite{MPI14} and leads to efficient interpolation and finite difference procedures \cite{Mirzadeh;etal:16:Parallel-level-set}.  For additional {\tt p4est} details and algorithms, the reader may consult \cite{Burstedde;Wilcox;Ghattas:11:p4est:-Scalable-Algo}.

As remarked at the end of \Cref{sec:TheLevelSetMethod}, the present study is also based on Mirzadeh and coauthors' parallel level-set schemes \cite{Mirzadeh;etal:16:Parallel-level-set}.  Mirzadeh \etal have exploited {\tt p4est}'s global-indexing and ghost-layering features to devise a simple reinitialization procedure and a scalable interpolation algorithm for semi-Lagrangian advection.  In the upcoming sections, we will use their redistancing scheme and Algorithm \href{https://www.sciencedirect.com/science/article/pii/S002199911630242X\#fg0050}{2} to solve \cref{eq:Reinitialization} and estimate $\kappa$ at $\Gamma$.  Meanwhile, we refer the reader to \cite{Mirzadeh;etal:16:Parallel-level-set, Strain1999, Min;Gibou:07:A-second-order-accur} for further information about parallelizable level-set procedures on non-uniform meshes.


\colorsection{Methodology}
\label{sec:Methodology}

In this study, we investigate the possibility of extending the error-correcting approach introduced in \cite{Larios;Gibou;ECNetSemiLagrangian;2021, Larios;Gibou;KECNet2D;2022} to the three-dimensional mean-curvature problem.  Our goal is to endow the level-set framework with data-driven mechanisms that compensate for its lack of inbuilt features to calculate $\kappa$ \textit{at} the interface \cite{Popinet;NumModelsOfSurfTension;18}.  Traditionally, one circumvents this limitation by first evaluating \cref{eq:MeanCurvature.3d} at the grid nodes.  Then, we interpolate these results at their nearest locations on $\Gamma$.  This procedure, however, does not always yield accurate estimations around under-resolved regions and steep surface portions.  For this reason, we propose a methodology that computes on-the-fly corrections to such numerical approximations using deep learning.  More succinctly, if $h\kappa$ represents the \textit{dimensionless mean curvature}\footnote{A few notation pointers: $h\kappa$ denotes the numerical baseline estimation to the dimensionless mean curvature, $h\kappa_\mathcal{F}$ is the neurally corrected value, and $h\kappa^\star$ is the best approximation from our proposed hybrid solver to the true $h\kappa^*$.} interpolated at node $\mathcal{n}$'s perpendicular projection

\begin{equation}
\vv{x}_\mathcal{n}^\perp \approx \vv{x}_\mathcal{n}^\Gamma = \vv{x}_\mathcal{n} - \phi(\vv{x}_\mathcal{n})\frac{\nabla\phi(\vv{x}_\mathcal{n})}{\|\nabla\phi(\vv{x}_\mathcal{n})\|},
\label{eq:NormalProjection}
\end{equation}
then the expression

\begin{equation}
h\kappa^* = h\kappa + \bar{\varepsilon}
\label{eq:MeanCurvatureError}
\end{equation}
realizes the numerical deviation from the true $h\kappa^*$ by some $\bar{\varepsilon}$.

We claim neural networks trained on various stencil configurations can model $\bar{\varepsilon}$ and fix $h\kappa$ selectively.  Here, we plan to devise a \textit{hybrid curvature solver} by coupling these networks and the schemes from \Cref{sec:TheLevelSetMethod} to furnish better estimations denoted by $h\kappa^\star$.  In particular, one should restrict such an inference system to \textit{interface nodes}, where $\phi(\vv{x}_\mathcal{n})\cdot\phi(\vv{y}) \leqslant 0$ holds at least for one of the six neighbors at $\vv{y} = \vv{x}_\mathcal{n} \pm h\hat{\vv{e}}_i$, in the direction of the standard $\mathbb{R}^3$ basis vectors $\hat{\vv{e}}_i$.


\FloatBarrier
\colorsubsection{A hybrid mean-curvature solver enhanced by error-correcting neural networks}
\label{subsec:HybridMeanCurvatureSolverAndECNets}

\begin{figure}[!t]
	\centering
	\begin{subfigure}[b]{7.25cm}
		\includegraphics[width=\textwidth]{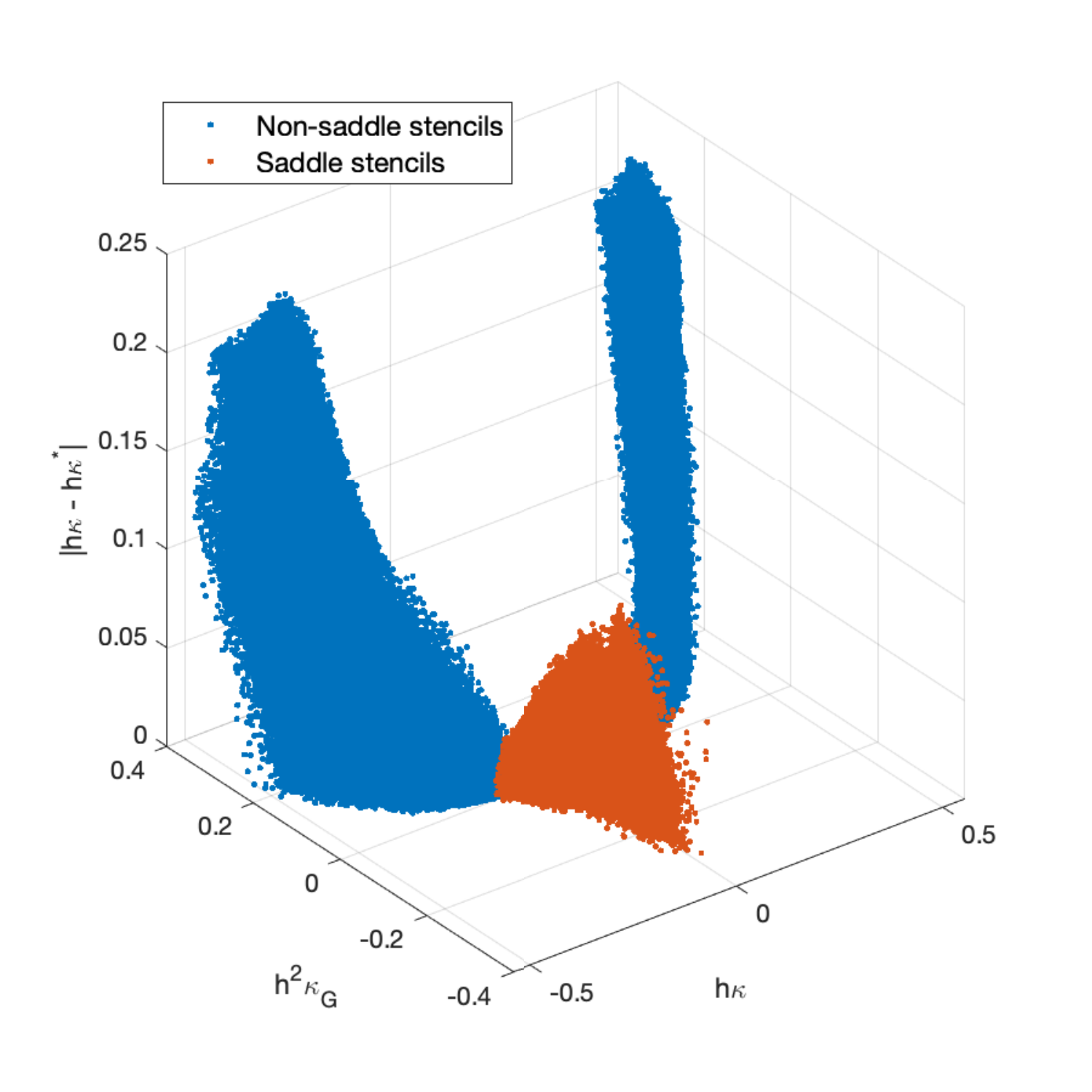}
		\caption{\footnotesize Three-dimensional $h\kappa$-$h^2\kappa_G$-error perspective}
		\label{fig:PrepAnalyses.Perspective}
	\end{subfigure}
	~
	\begin{subfigure}[b]{7.25cm}
		\includegraphics[width=\textwidth]{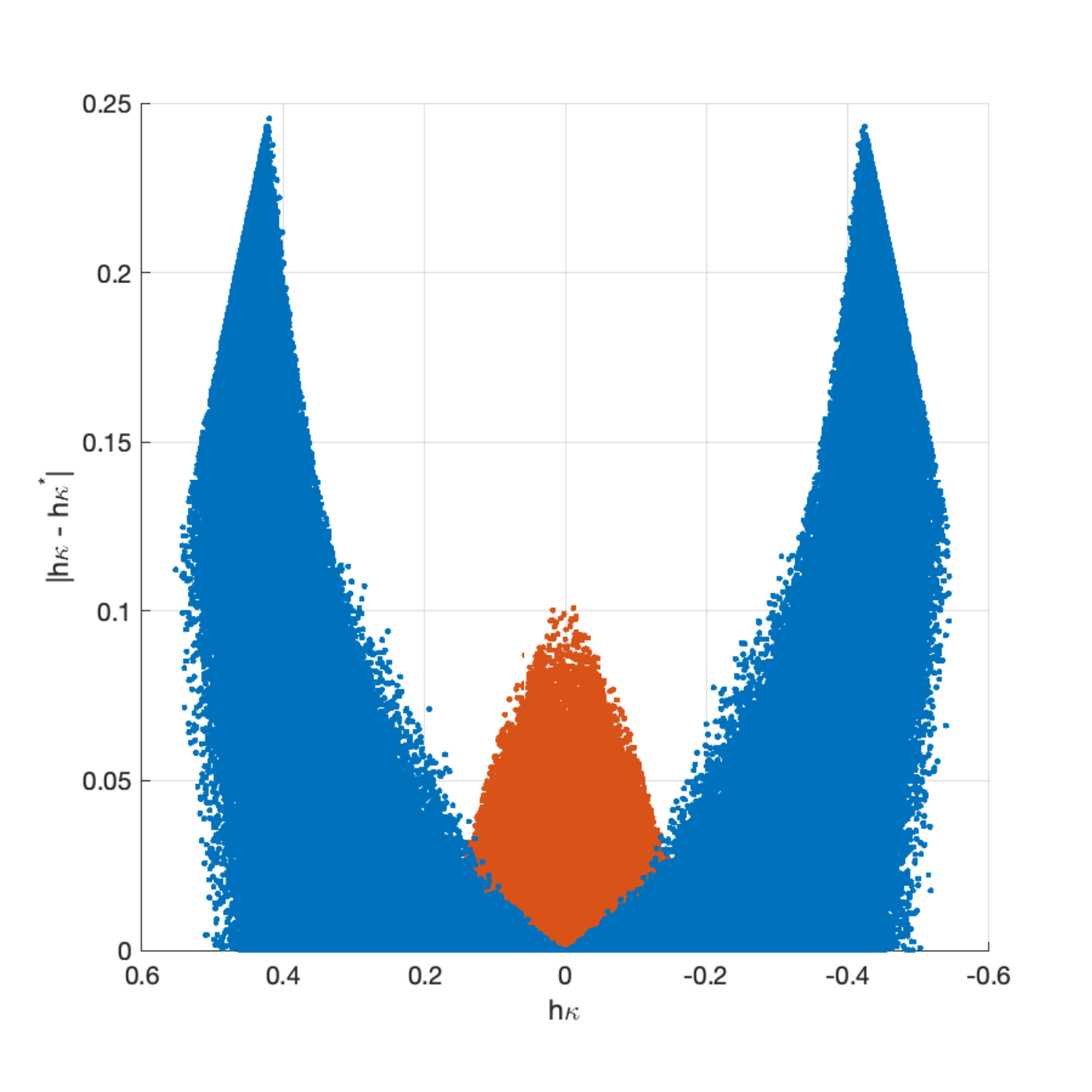}
		\caption{\footnotesize $h\kappa$-error planar view}
		\label{fig:PrepAnalyses.hk_error_plane}
	\end{subfigure}
	\\
	\begin{subfigure}[b]{7.25cm}
		\includegraphics[width=\textwidth]{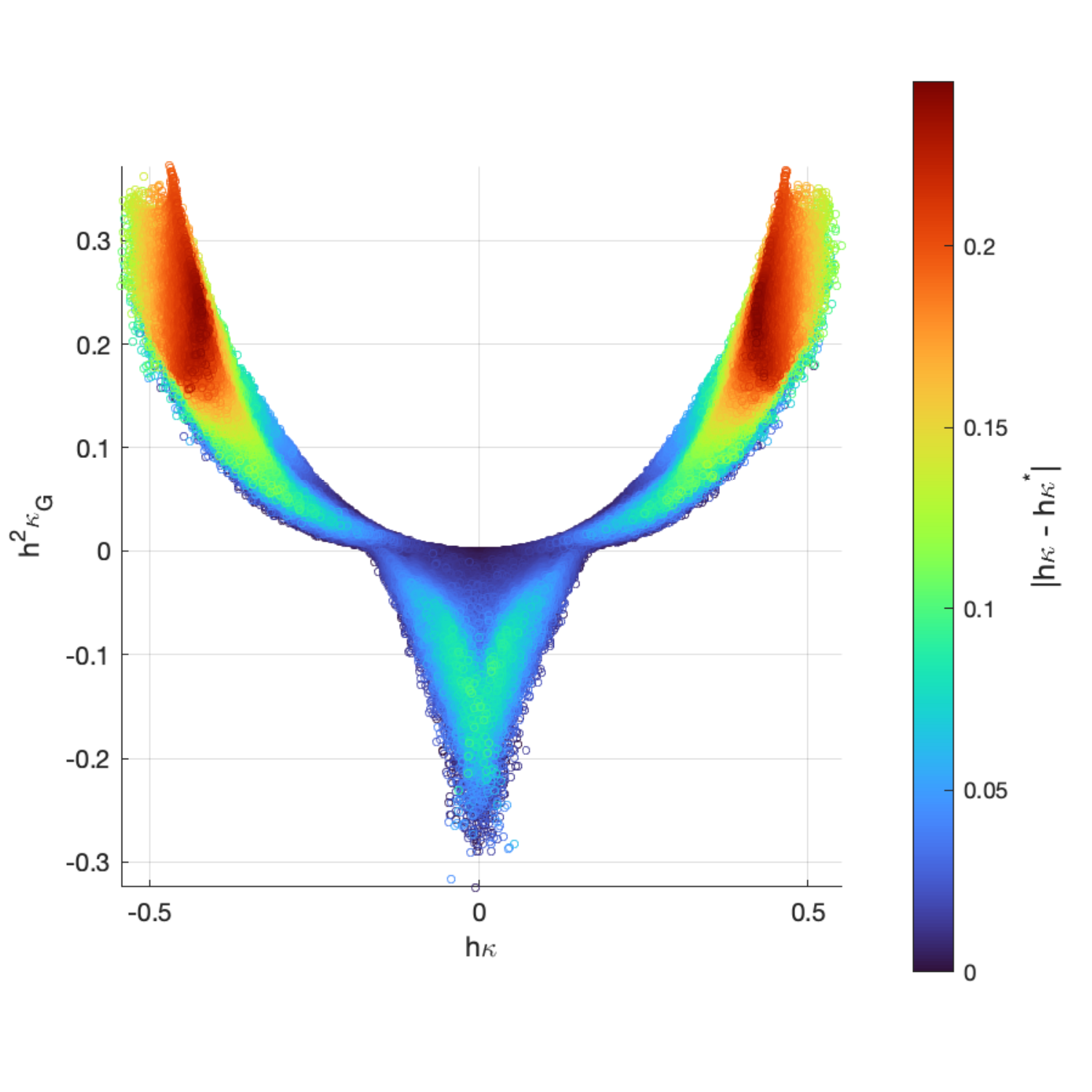}
		\caption{\footnotesize $h\kappa$-$h^2\kappa_G$ planar view}
		\label{fig:PrepAnalyses.hk_h2kg_plane}
	\end{subfigure}
	\caption{Error $|\bar{\varepsilon}|$ as a function of non-dimensionalized mean and Gaussian curvatures from sinusoidal data.  The top row provides a three-dimensional perspective in (a) and a planar view in (b), where non-saddle-region interface nodes appear in blue (\textcolor{bluematlab}{$\bullet$}) and saddle-region ones in dark orange (\textcolor{orangematlab}{$\bullet$}).  The bottom picture in (c) shades the error projection onto the curvature plane as a heat map.  (Color online.)}
	\label{fig:PrepAnalyses}
\end{figure}

In analogy to the two-dimensional problem in \cite{Larios;Gibou;KECNet2D;2022}, we propose using a neural function, $\mathcal{F}_\kappa(\cdot)$, to characterize $\bar{\varepsilon}$ in \cref{eq:MeanCurvatureError}.  However, building this neural network requires further analysis in $\mathbb{R}^3$ than simply deciding on what features it should ingest.  For example, \cref{fig:PrepAnalyses} plots the distribution of $|\bar{\varepsilon}| \equiv |h\kappa - h\kappa^*|$ for a collection of nodes adjacent to randomized sinusoidal surfaces.  Sinusoidal waves are essential because they host a variety of stencil patterns determined by a wide range of Gaussian curvature values associated with their center nodes.  \Cref{subsubsec:SinusoidalInterfaceDataSetConstruction} discusses this class of interfaces from the training point of view, but we introduce them at this point to motivate neural network design.  Let $h\kappa$ and $h^2\kappa_G$ be the linearly interpolated, non-dimensionalized mean and Gaussian curvatures at $\Gamma$.  Then, a swift inspection of \cref{fig:PrepAnalyses} reveals the following key observations:

\begin{enumerate}[label=OB.\arabic*]
	\item \label{item:OBNonSaddle} If $h^2\kappa_G \gtrapprox 0$, we have that $|\bar{\varepsilon}| \to 0$ as $h\kappa \to 0$, and we can safely disregard $h^2\kappa_G$.
	\item \label{item:OBSaddle} If $h^2\kappa_G \lessapprox 0$, $|\bar{\varepsilon}|$ grows nonlinearly with $|h\kappa|$ and $|h^2\kappa_G|$ in a less separable fashion.
\end{enumerate}

These observations suggest incorporating $h^2\kappa_G$ as an inexpensive stencil classifier.  In particular, \ref{item:OBNonSaddle} corresponds to the typical behavior alluded to in \cite{Larios;Gibou;HybridCurvature;2021, Larios;Gibou;KECNet2D;2022} for the two-dimensional curvature problem.  Here, we refer to patterns described by this observation as \textit{non-saddle stencils}.  As seen in \cref{fig:PrepAnalyses.hk_error_plane,fig:PrepAnalyses.hk_h2kg_plane}, there is an intrinsic symmetry about the plane $h\kappa = 0$ that we can exploit to simplify $\mathcal{F}_\kappa(\cdot)$'s topology.  In this case, we can train $\mathcal{F}_\kappa(\cdot)$ only on the negative mean-curvature spectrum and rely on numerical schemes to determine whether the stencil belongs to a convex or concave region.

Observation \ref{item:OBSaddle}, on the other hand, depicts a more complex use case that prevents us from exploiting symmetry like before. For instance, $|\bar{\varepsilon}|$ can be large, even for \textit{saddle stencils}\footnote{We will extend the usage of the \textit{non-saddle} and \textit{saddle} qualifiers to distinguish not only stencils but also patterns, regions, samples, and data---all of them characterized by observations \ref{item:OBNonSaddle} and \ref{item:OBSaddle}.} whose $h\kappa$ value at $\Gamma$ is close to zero (e.g., see \cref{fig:PrepAnalyses.hk_h2kg_plane}).  This non-saddle- and saddle-pattern dichotomy prompts us to build two neural functions: $\mathcal{F}_\kappa^{ns}(\cdot)$ and $\mathcal{F}_\kappa^{sd}(\cdot)$.  At inference time, the solver would choose one or the other depending on $h^2\kappa_G$.  To distinguish non-saddle from saddle objects, sets, or artifacts, we will continue to use the superscripts $ns$ and $sd$ in the remaining sections of this manuscript.

\begin{figure}[!t]
	\centering
	\includegraphics[width=15.2cm]{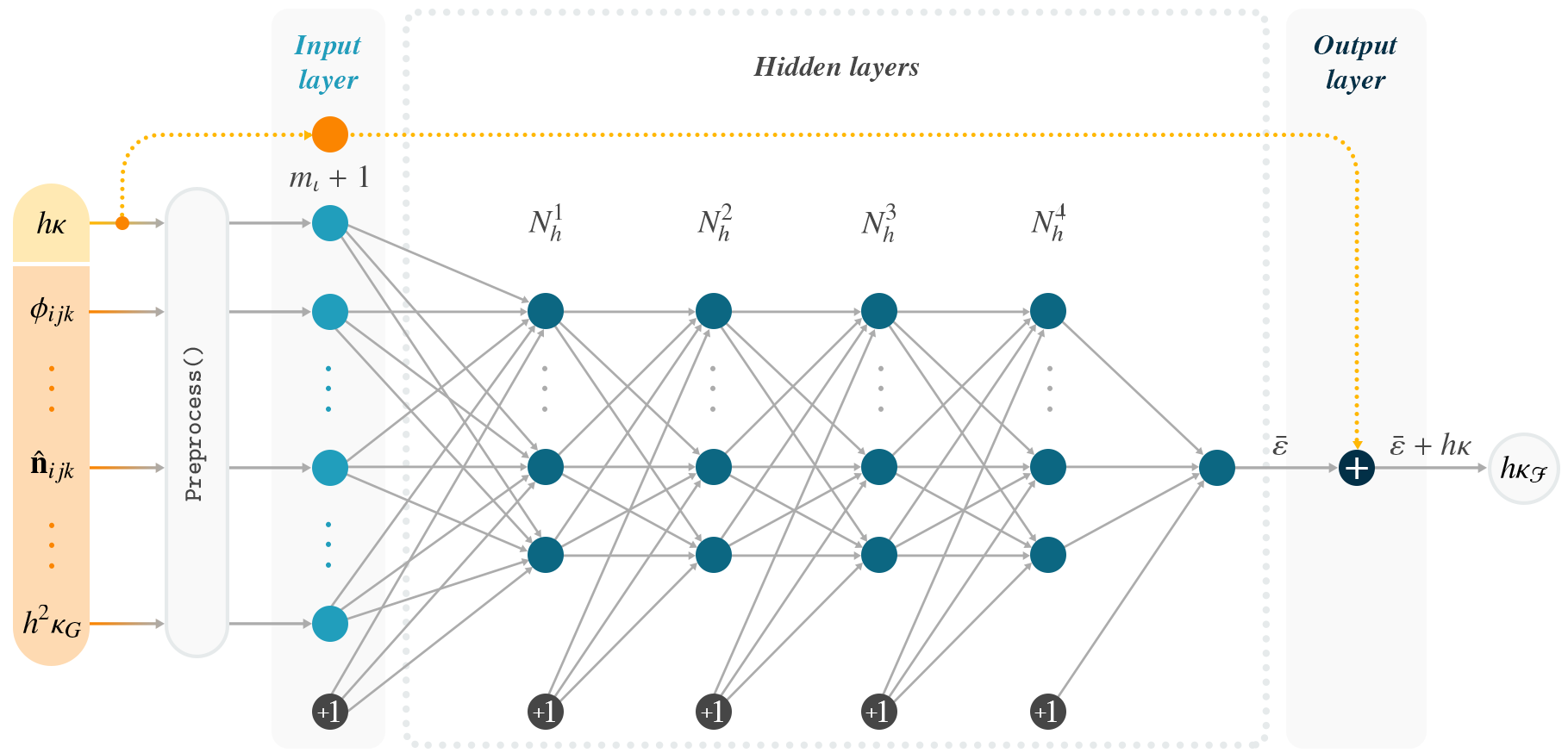}
	\caption{The mean-curvature error-correcting neural network and its preprocessing module.  $\mathcal{F}_\kappa^{ns}(\cdot)$ and $\mathcal{F}_\kappa^{sd}(\cdot)$ exhibit this feedforward architecture.  The raw 110-feature input vectors include level-set values ($\phi_{ijk}$), unit normal vectors ($\hat{\vv{n}}_{ijk}$), and dimensionless mean ($h\kappa$) and Gaussian ($h^2\kappa_G$) curvatures.  The hidden, nonlinear units responsible for computing $\bar{\varepsilon}$ appear in aqua (\textcolor{darkaqua}{$\bullet$}) with incoming $(+1)$-labeled biases shown in dark gray (\textcolor{darkgray}{$\bullet$}).  After processing $m_\iota$ transformed inputs, the MLP yields $h\kappa_\mathcal{F} \doteq \bar{\varepsilon} + h\kappa$ through a (dotted, yellow) skip-connection and an additive output neuron shaded in navy blue (\textcolor{navy}{$\bullet$}).  (Color online.)}
	\label{fig:ECNet}
\end{figure}

$\mathcal{F}_\kappa^{ns}(\cdot)$ and $\mathcal{F}_\kappa^{sd}(\cdot)$ are as simple as MLPs.  \Cref{fig:ECNet} overviews their distinctive cascading architecture organized into five hidden layers.  This diagram also illustrates the preprocessing module and a skip-connection that carries the $h\kappa$ signal from the input layer to a non-adaptable output neuron.  The latter completes the error-correcting task by adding $h\kappa$ to the $\bar{\varepsilon}$ value computed internally to produce $h\kappa_\mathcal{F}$.  As stated above, we expect $h\kappa_\mathcal{F}$ to be closer to $h\kappa^*$ than $h\kappa$, especially around under-resolved regions.

The {\tt Preprocess()} subroutine in \cref{fig:ECNet} transforms incoming raw feature vectors into a more suitable representation for training and evaluating our downstream regressor \cite{scikit-learn11}.  In this work, we have constrained ourselves to evaluate data extracted only from regular $h$-uniform 27-point stencils.  Later, in \Cref{subsec:TechnicalAspects}, we will discuss the {\tt Preprocess()} module's specific implementation alongside the technical corners of the project.  To provide $\mathcal{F}_\kappa^{ns}(\cdot)$ and $\mathcal{F}_\kappa^{sd}(\cdot)$ with as much information as possible, we consider more field data than just level-set values \cite{LALariosFGibou;LSCurvatureML;2021, Larios;Gibou;HybridCurvature;2021} (or volume fractions in VOF technologies \cite{CurvatureML19, VOFCurvature3DML19}).  Similar to the two-dimensional approach \cite{Larios;Gibou;KECNet2D;2022}, our hybrid solver and models take in stencil level-set and gradient data, as well as the Gaussian and mean curvatures projected onto $\Gamma$ from the center grid point (see \cref{fig:Stencil}).  We have organized all these features into \textit{data packets} of the form

\begin{equation}
\mathcal{p} = \left(\begin{array}{rl}
			\phi_{ijk}\!: & \textrm{27-point stencil level-set values} \\
	\hat{\vv{n}}_{ijk}\!: & \textrm{27-point stencil unit normal vectors} \\
			   h\kappa\!: & \textrm{dimensionless mean curvature interpolated at } \vv{x}_\mathcal{n}^\Gamma \\
		   h^2\kappa_G\!: & \textrm{dimensionless Gaussian curvature interpolated at } \vv{x}_\mathcal{n}^\Gamma
\end{array}\right) \in \mathbb{R}^{110},
\label{eq:DataPacket}
\end{equation}
where $\psi_{ijk} \doteq \psi(x_0 + ih,\, y_0 + jh,\, z_0 + kh)$, with $i,j,k \in \{\keyval{m}{-1},\, \keyval{0}{0},\, \keyval{p}{+1}\}$\footnote{This notation resembles the Pythonic representation of a {\tt dict} data type.  We also use $m$, $0$, and $p$ as subscripts to enumerate grid points in a stencil.}, is a nodal attribute within $\mathcal{n}$'s stencil centered at $\vv{x}_\mathcal{n} = [x_\mathcal{n},\, y_\mathcal{n},\, z_\mathcal{n}]^T$.

\begin{figure}[!t]
	\centering
	\begin{subfigure}[b]{6.5cm}
		\includegraphics[width=\textwidth]{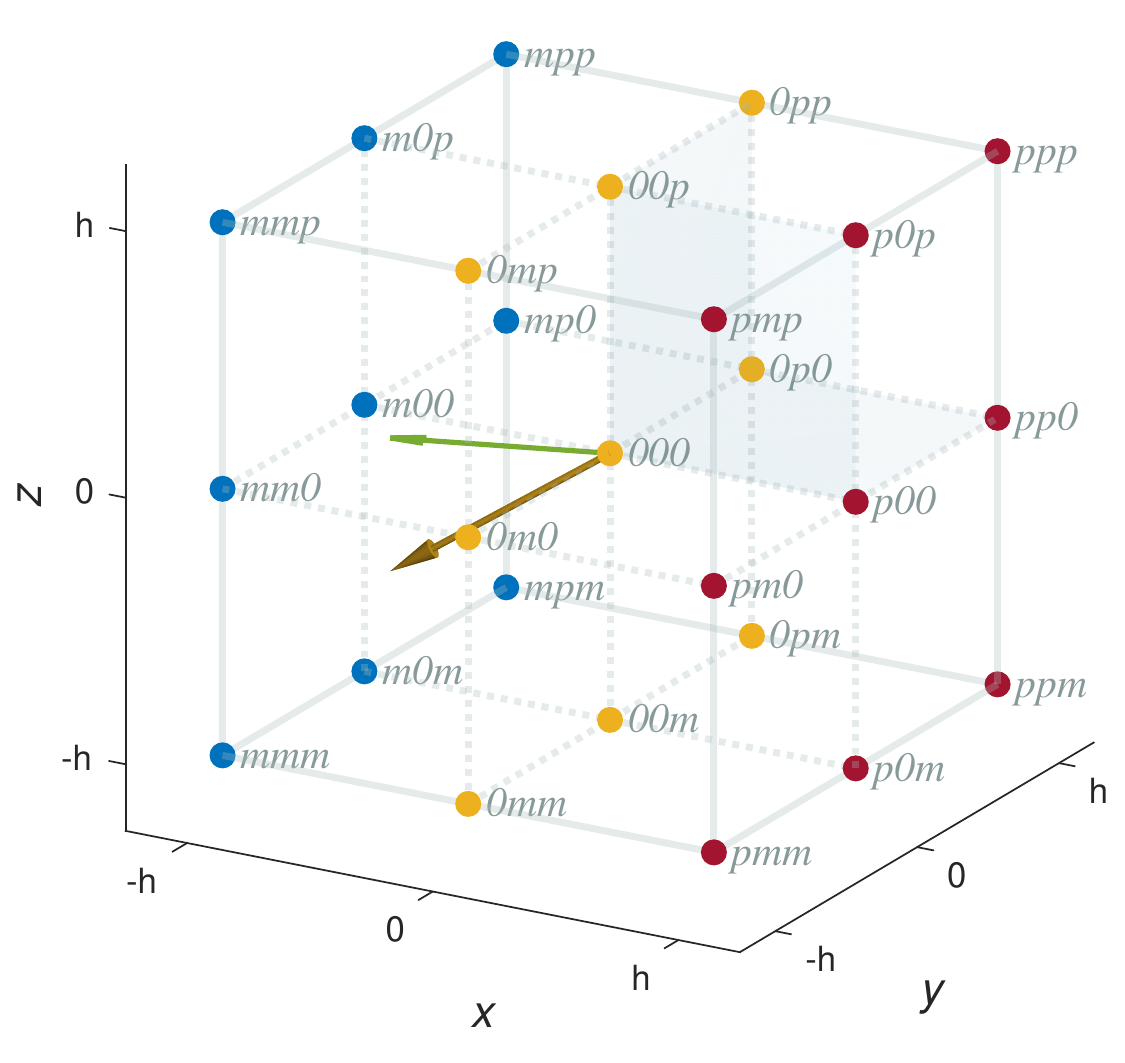}
		\caption{\footnotesize (Original) $\mathcal{p}_0$}
		\label{fig:DataPackets.Original}
	\end{subfigure}
	~
	\begin{subfigure}[b]{6.5cm}
		\includegraphics[width=\textwidth]{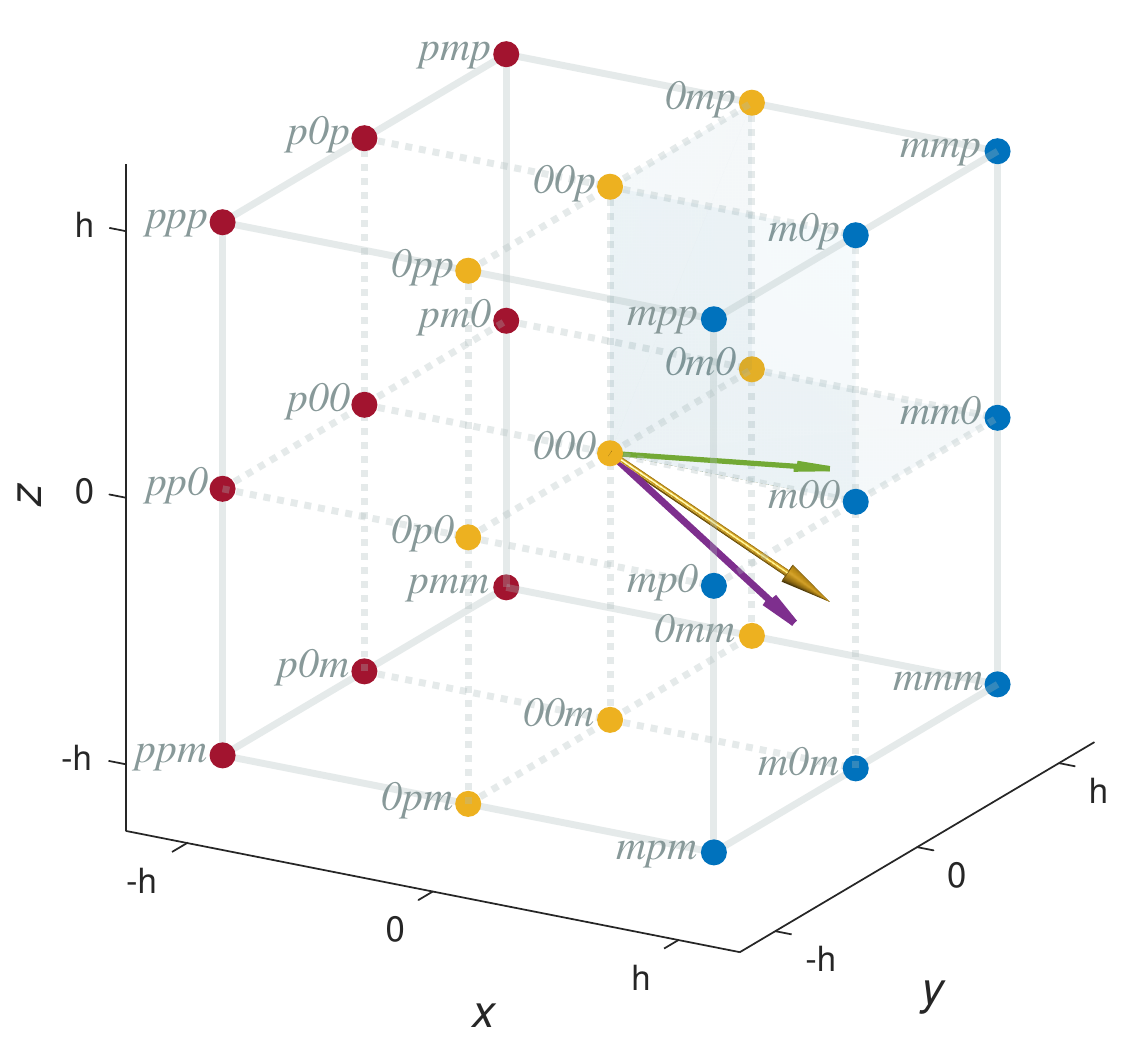}
		\caption{\footnotesize Intermediate rotation about the $z$-axis}
		\label{fig:DataPackets.Intermediate}
	\end{subfigure}
	\\
	\begin{subfigure}[b]{6.5cm}
		\includegraphics[width=\textwidth]{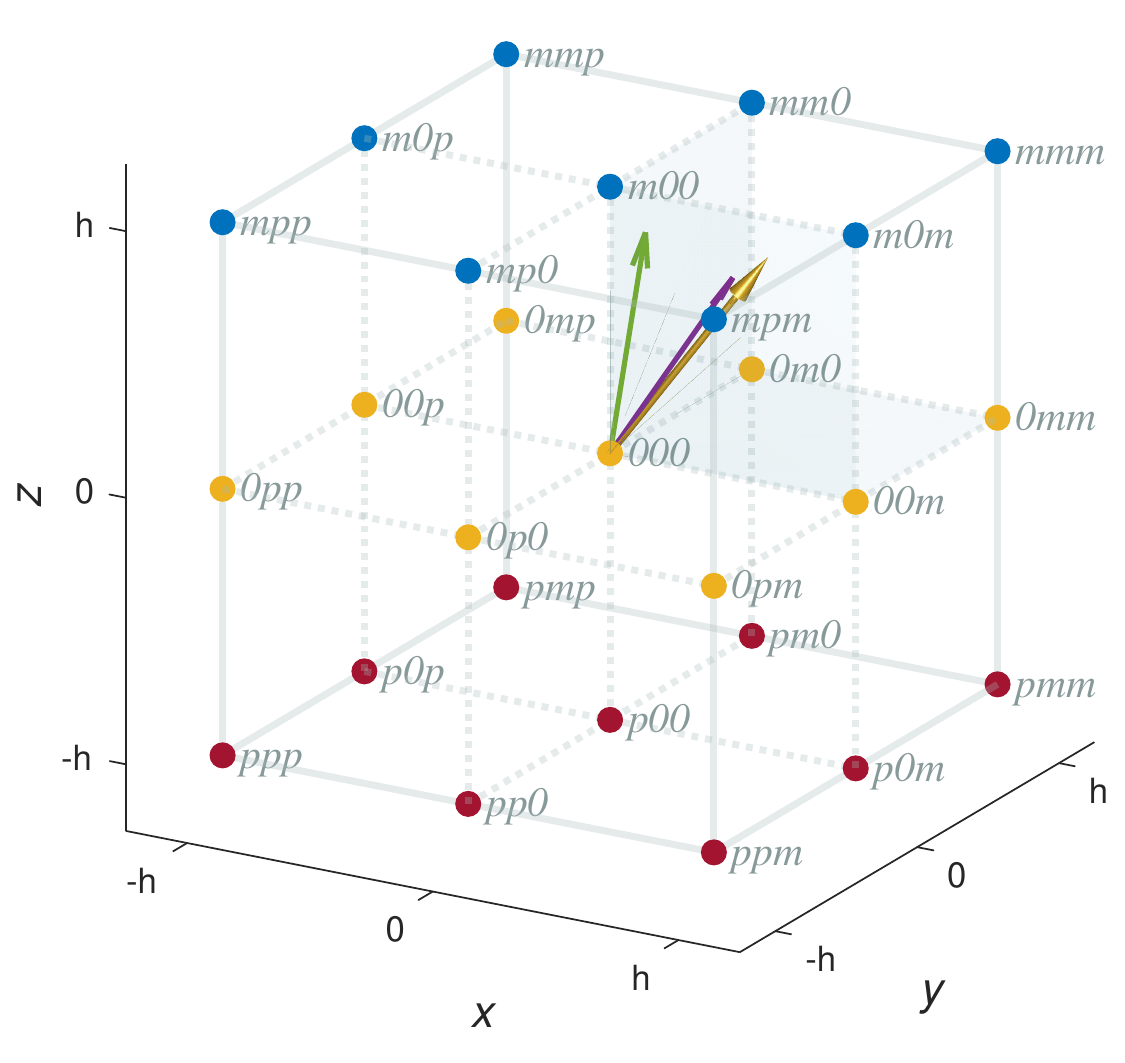}
		\caption{\footnotesize (Reoriented) $\mathcal{p}_1$ {\color{gray}($\alpha\hat{\vv{n}}_{000} = [0.5,\, 0.25,\, 0.75]^T$)}}
		\label{fig:DataPackets.StdForm}
	\end{subfigure}
	~
	\begin{subfigure}[b]{6.5cm}
		\includegraphics[width=\textwidth]{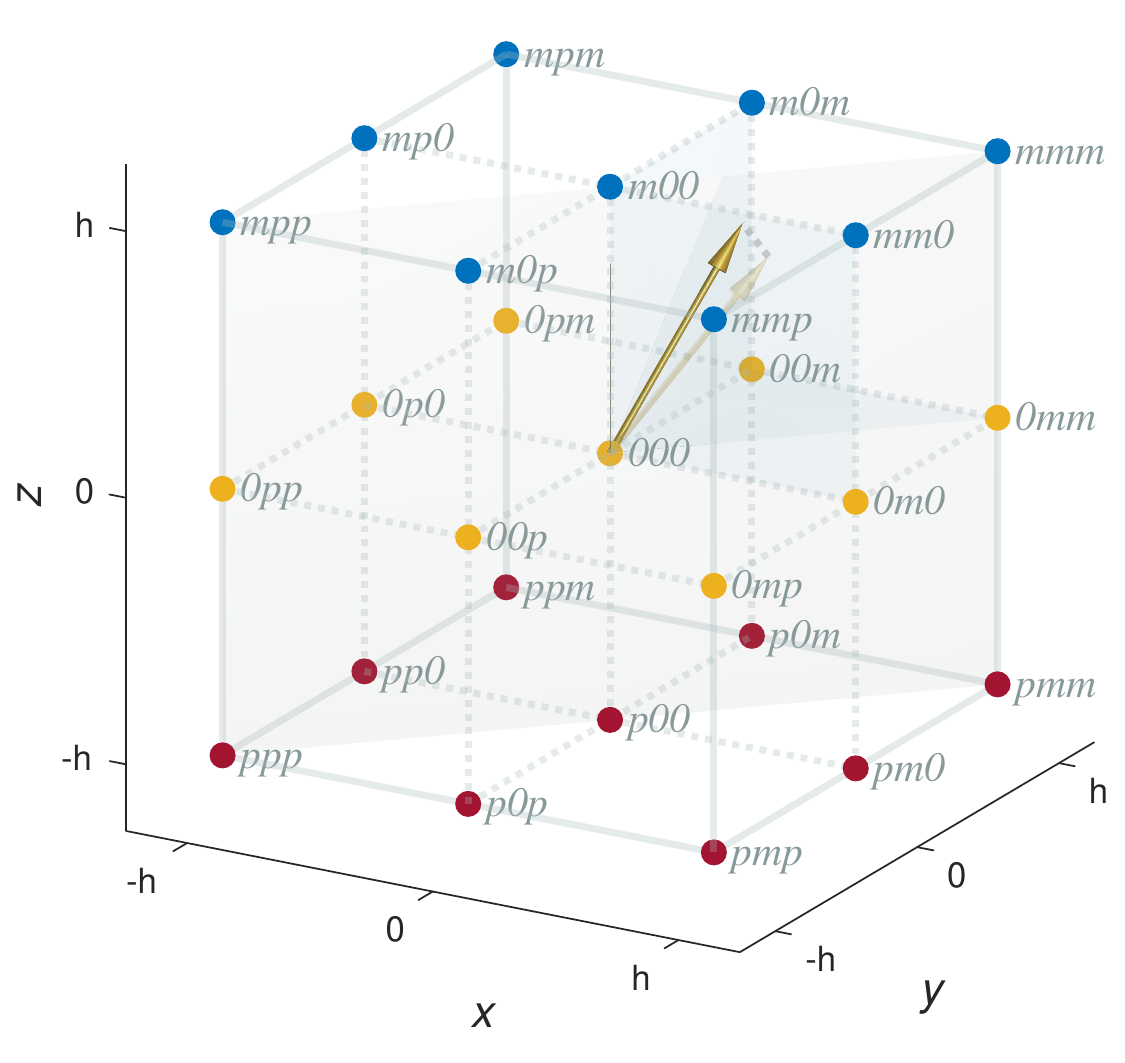}
		\caption{\footnotesize (Reflected) $\mathcal{p}_2$ {\color{gray}($\alpha\hat{\vv{n}}_{000} = [0.25,\, 0.5,\, 0.75]^T$)}}
		\label{fig:DataPackets.RefForm}
	\end{subfigure}
	\caption{Tracking the reorientation of a data packet associated with an arbitrary interface node $\mathcal{n}$.  The grid points are enumerated in the same order as \cref{fig:Stencil}, where $\vv{x}_\mathcal{n}$ is $\mathcal{n}$'s location at the center of the stencil ($000$).  A three-dimensional golden arrow represents $\alpha\hat{\vv{n}}_{000}$, where $\hat{\vv{n}}_{000} \equiv \hat{\vv{n}}_\mathcal{n}$ and $\alpha > 0$.  To ease visualization, we have colored the nodes resting on the same initial $x$-layer with blue (\textcolor{bluematlab}{$\bullet$}) ($x = x_\mathcal{n}-h$), yellow (\textcolor{yellowmatlab}{$\bullet$}) ($x = x_\mathcal{n}$), and red (\textcolor{redmatlab}{$\bullet$}) ($x = x_\mathcal{n}+h$).  The original data packet, $\mathcal{p}_0$, appears in (a) alongside $\alpha\hat{\vv{n}}_{000}$'s green-shaded $xy$-projection.  After a $z$-axis rotation, we produce the intermediate state in (b), where the green arrow now lies in the first local octant.  Also in (b), a purple arrow indicates $\alpha\hat{\vv{n}}_{000}$'s intermediate projection onto the $xz$-plane.  A $-\frac{\pi}{2}$ rotation about the $y$-axis finally makes the auxiliary projections' and all of $\hat{\vv{n}}_{000}$'s components nonnegative and leaves the data packet in its (first) standard form, $\mathcal{p}_1$, in (c).  (d) depicts $\mathcal{p}_1$'s reflection about the (light-gray) plane $(x-x_\mathcal{n})-(y-y_\mathcal{n})=0$ going through $\mathcal{n}$ to produce $\mathcal{p}_2$.  Notice that we have evolved the node labels as we transform $\mathcal{n}$'s data packet.  For consistency, however, one must relabel the 27-point stencil in (c) and (d) so that the $mmm$th entry always refers to the left-back-bottom-most corner.  (c) and (d) provide the example coordinates of $\alpha\hat{\vv{n}}_{000}$ for their corresponding reoriented and reflected data packets.  See \Cref{alg:GenerateStdDataPackets} and \cref{fig:AdditionalDataPackets} for more details about producing the remaining four standard forms.  (Color online.)}
	\label{fig:DataPackets}
\end{figure}

Likewise, we exploit curvature rotation and reflection invariance to reduce the problem's degrees of freedom and improve prediction stability in two ways.  First, we train $\mathcal{F}_\kappa^{ns}(\cdot)$ and $\mathcal{F}_\kappa^{sd}(\cdot)$ with a single class of normalized stencil configuration.  In particular, we can always find a sequence of 90-degree rotations about the Cartesian-coordinate axes to leave $\mathcal{p}$ into its first \textit{standard form}, $\mathcal{p}_1$, where $\hat{\vv{n}}_\mathcal{n} \equiv \hat{\vv{n}}(\vv{x}_\mathcal{n}) \equiv \mathcal{p}.\hat{\vv{n}}_{000}$ has all its components nonnegative.  Not only does rotating stencils in this manner preserve $h\kappa$ and $h^2\kappa_G$, but it also prevents our MLPs from diverting neural capacity to unnecessary data-packet orientations.  This type of normalization is a staple preprocessing step in eigenface computation \cite{Turk;Pentland;Eigenfaces;1991, Parker;CS170A;2016}. And recently, it has been incorporated into data-driven semi-Lagrangian advection \cite{Larios;Gibou;ECNetSemiLagrangian;2021} and two-dimensional curvature estimation \cite{Larios;Gibou;KECNet2D;2022}.  \Crefrange{fig:DataPackets.Original}{fig:DataPackets.StdForm} exemplify the reorientation process on the raw data packet, $\mathcal{p}_0$, associated with an arbitrary interface node $\mathcal{n}$.  The reader may track $\mathcal{p}_0$'s transformation into $\mathcal{p}_1$ by following the evolution of its nodal labels and colors.

\begin{figure}[!t]
	\centering
	\begin{subfigure}[b]{6.5cm}
		\includegraphics[width=\textwidth]{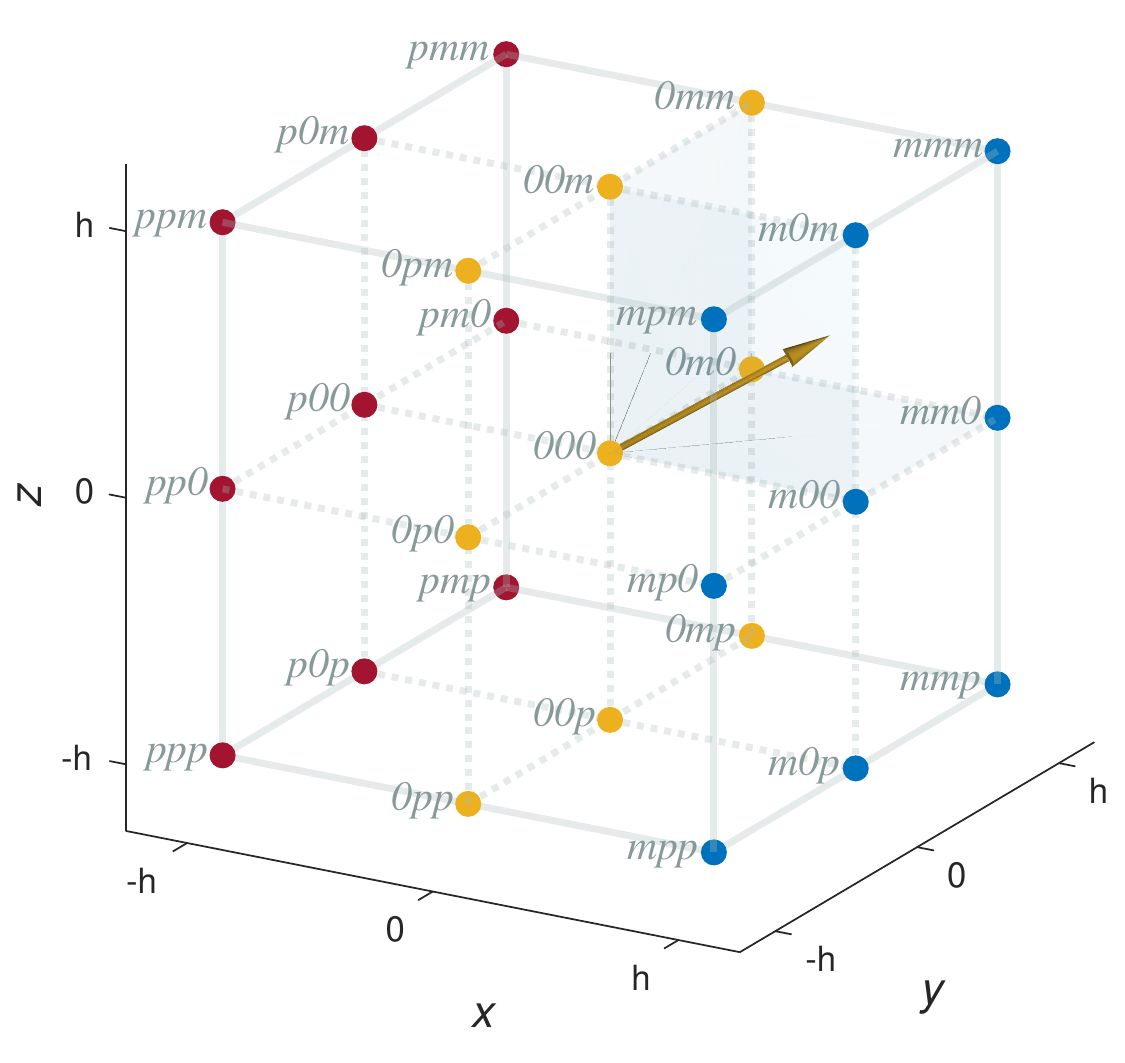}
		\caption{\footnotesize (Reoriented) $\mathcal{p}_{11}$ {\color{gray}($\alpha\hat{\vv{n}}_{000} = [0.75,\, 0.25,\, 0.5]^T$)}}
		\label{fig:AdditionalDataPackets.StdForm.StdForm}
	\end{subfigure}
	~
	\begin{subfigure}[b]{6.5cm}
		\includegraphics[width=\textwidth]{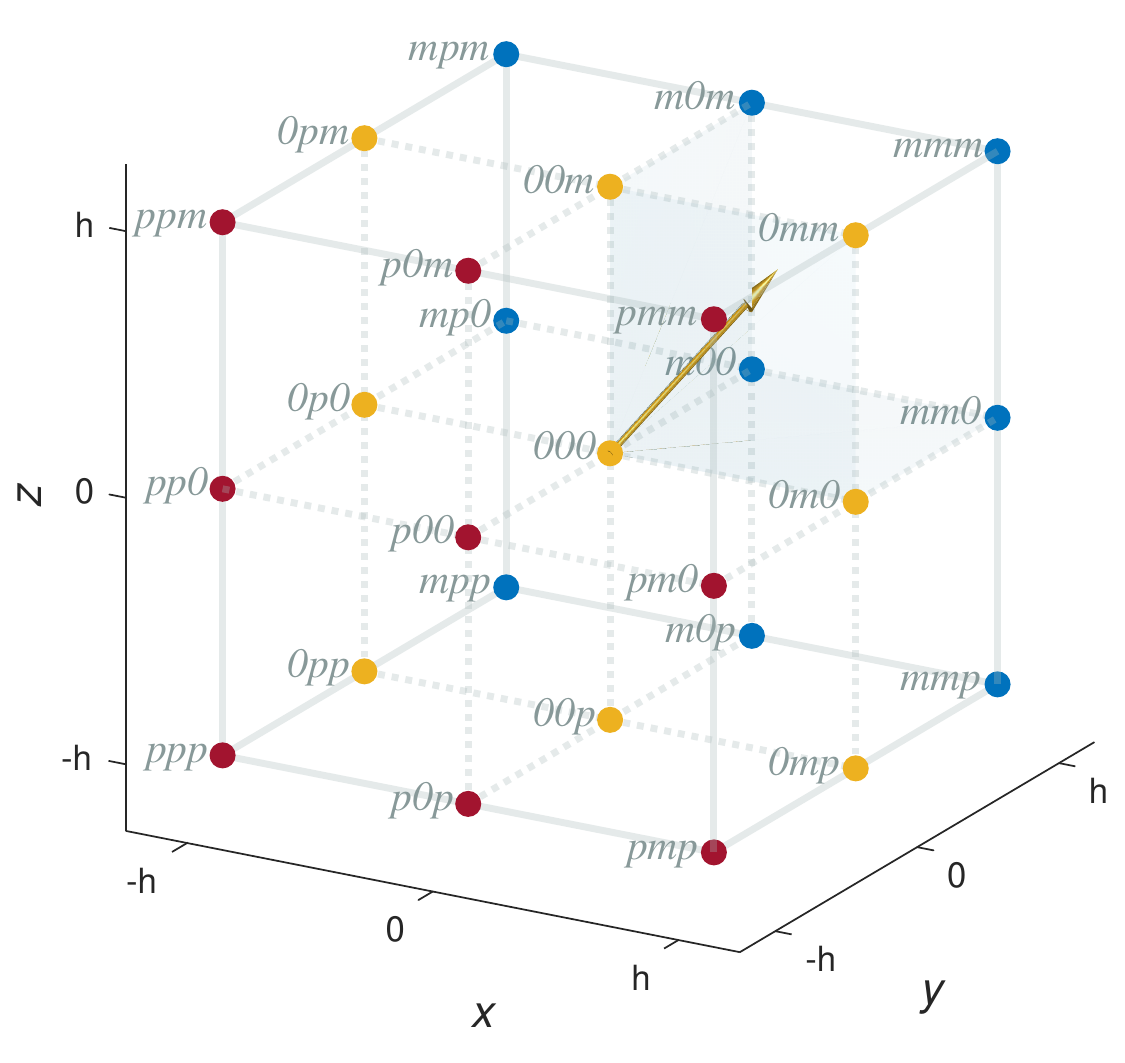}
		\caption{\footnotesize (Reflected) $\mathcal{p}_{12}$ {\color{gray}($\alpha\hat{\vv{n}}_{000} = [0.25,\, 0.75,\, 0.5]^T$)}}
		\label{fig:AdditionalDataPackets.StdForm.RefForm}
	\end{subfigure}
	\\
	\begin{subfigure}[b]{6.5cm}
		\includegraphics[width=\textwidth]{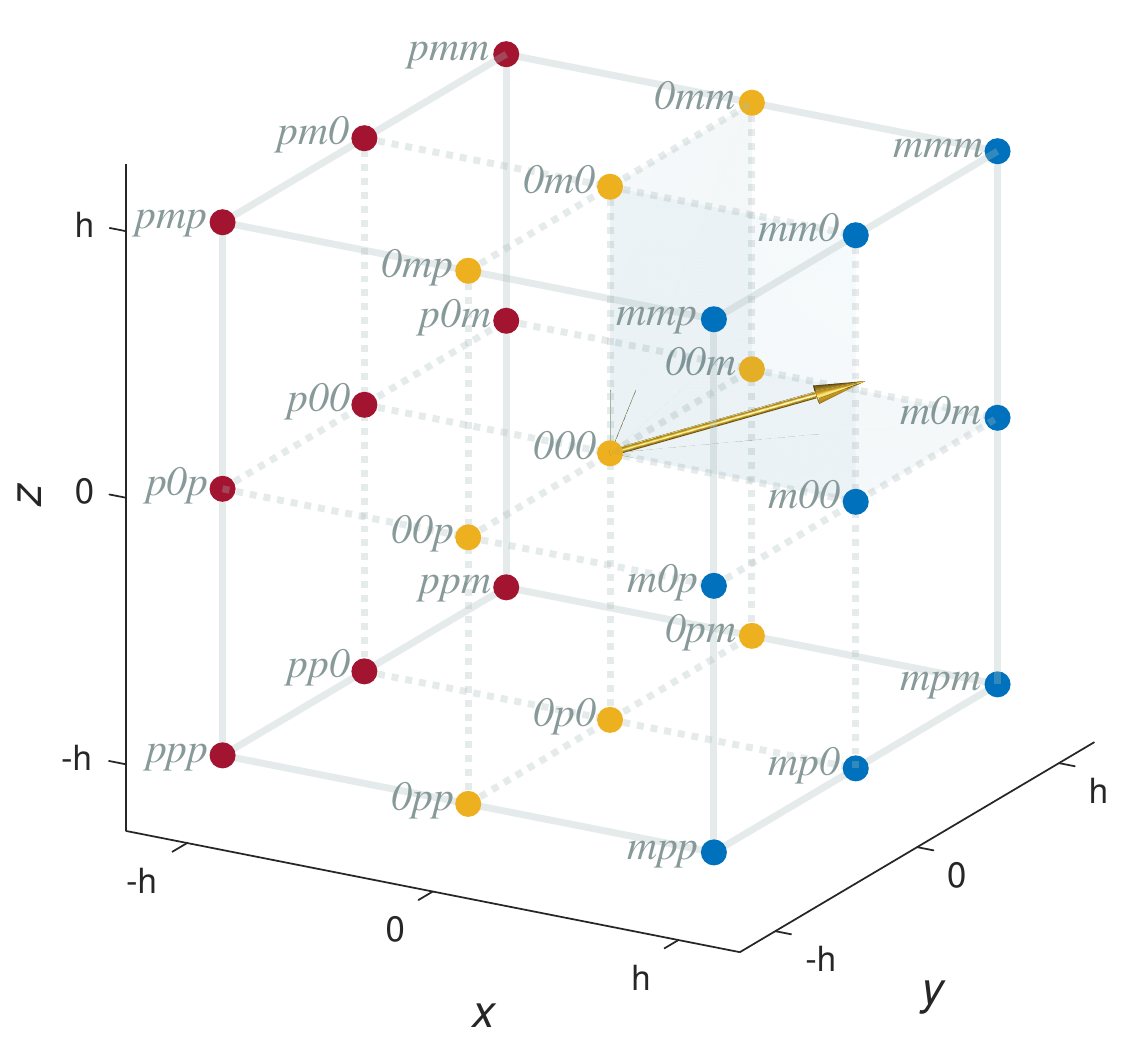}
		\caption{\footnotesize (Reoriented) $\mathcal{p}_{21}$ {\color{gray}($\alpha\hat{\vv{n}}_{000} = [0.75,\, 0.5,\, 0.25]^T$)}}
		\label{fig:AdditionalDataPackets.RefForm.StdForm}
	\end{subfigure}
	~
	\begin{subfigure}[b]{6.5cm}
		\includegraphics[width=\textwidth]{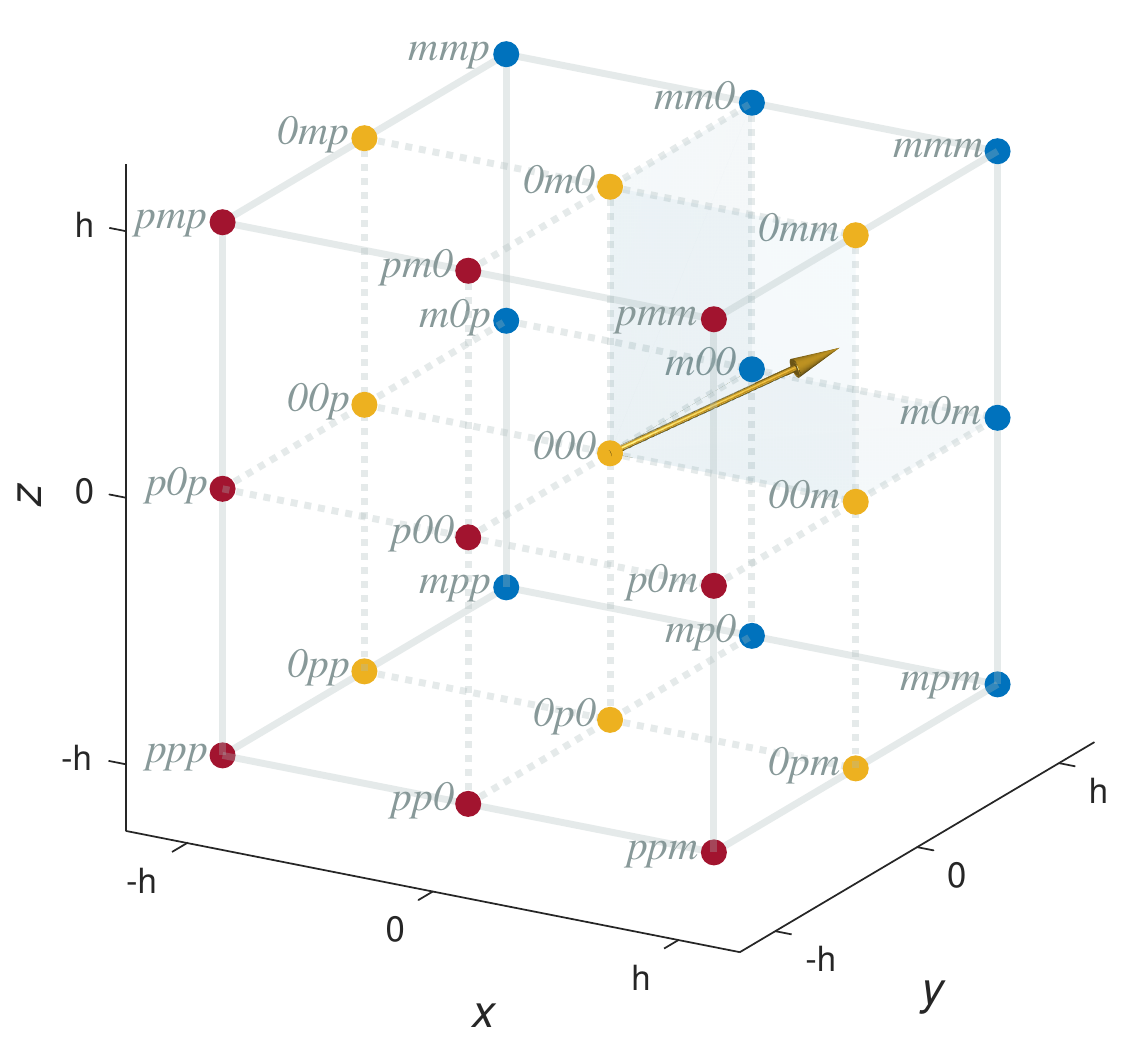}
		\caption{\footnotesize (Reflected) $\mathcal{p}_{22}$ {\color{gray}($\alpha\hat{\vv{n}}_{000} = [0.5,\, 0.75,\, 0.25]^T$)}}
		\label{fig:AdditionalDataPackets.RefForm.RefForm}
	\end{subfigure}
	\caption{Additional standard-formed data packets that complete the six permutations of $\hat{\vv{n}}_{000}$'s components at the center node.  $\mathcal{p}_{11}$ in (a) is extracted from \cref{fig:DataPackets.StdForm} by reflecting $\mathcal{p}_1$ about the plane $z - z_\mathcal{n} = 0$ followed by a $-\frac{\pi}{2}$ rotation about the $y$-axis.  To obtain $\mathcal{p}_{12}$ in (b), we simply reflect $\mathcal{p}_{11}$ about the plane $(x-x_\mathcal{n})-(y-y_\mathcal{n})=0$.  Replicating these steps on $\mathcal{p}_2$ from \cref{fig:DataPackets.RefForm} produces $\mathcal{p}_{21}$ and $\mathcal{p}_{22}$ in (c) and (d).  In all cases, we show the example coordinates for $\alpha\hat{\vv{n}}_{000}$ (with $\alpha > 0$), which appears as a golden arrow always pointing in the direction of the (highlighted) first local octant.  (Color online.)}
	\label{fig:AdditionalDataPackets}
\end{figure}

The second way of leveraging curvature rotation and reflection invariance regards symmetry enforcement.  Practitioners have already used this resource to estimate volume fractions and apertures \cite{Buhendwa;Bezgin;Adams;IRinLSwithML;2021} and improve two-dimensional curvature accuracy in level-set \cite{Larios;Gibou;KECNet2D;2022} and VOF \cite{Onder;etal;DLCurvatureVOF;2022} schemes.  The idea is to reflect and rotate a standard-formed data packet to extract additional stencil variations while keeping $\hat{\vv{n}}_\mathcal{n}$'s components nonnegative.  The benefit of such a process is twofold.  At the inference stage, we can evaluate our MLPs on these data packets and take the average of the corrected curvature predictions.  During training, this procedure facilitates data augmentation because it provides a geometric means for producing multiple samples per interface node.  \Crefrange{fig:DataPackets.StdForm}{fig:AdditionalDataPackets} depict $\mathcal{p}_0$'s six standard forms.  Each corresponds to a permutation of the reoriented unit normal-vector entries at the center grid point of \cref{fig:DataPackets.StdForm}.  To get these variations, we have combined a sequence of reflections about the planes $(x-x_\mathcal{n}) - (y-y_\mathcal{n}) = 0$ and $z-z_\mathcal{n} = 0$ and rotations about the Cartesian-coordinate axes, as shown in \Cref{alg:GenerateStdDataPackets}.

\begin{algorithm}[!t]
\SetAlgoLined

\KwIn{$\mathcal{p}_0$: feature data packet in its raw form; $\vv{x}_\mathcal{n}$: node $\mathcal{n}$'s location.}
\KwResult{$\mathcal{P}$: set of six data packets in standard form associated with node $\mathcal{n}$.}
\BlankLine

let $\mathcal{p}_1$ be the reoriented $\mathcal{p}_0$ so that $\mathcal{p}_1.\hat{\vv{n}}_{000}$ has all its components nonnegative\tcp*[r]{See \cref{fig:DataPackets.StdForm}}
let $\mathcal{p}_2$ be the reflected $\mathcal{p}_1$ about the plane $(x - x_\mathcal{n}) - (y - y_\mathcal{n}) = 0$ going through $\vv{x}_\mathcal{n} = [x_\mathcal{n},\, y_\mathcal{n},\, z_\mathcal{n}]^T$\tcp*[r]{See \cref{fig:DataPackets.RefForm}}
$\mathcal{P} \leftarrow \{\mathcal{p}_1,\, \mathcal{p}_2\}$\;
\BlankLine

\tcp{Produce two more packets for $\mathcal{p}_1$ and $\mathcal{p}_2$}
$\mathcal{Q} \leftarrow \varnothing$\;
\ForEach{\normalfont $\mathcal{q} \in \mathcal{P}$}{
	let $\mathcal{q}_z$ be the reflection of $\mathcal{q}$ about the plane $z - z_\mathcal{n} = 0$ going through $\vv{x}_\mathcal{n}$\;
	let $\mathcal{q}_1$ be the reoriented $\mathcal{q}_z$ so that $\mathcal{q}_1.\hat{\vv{n}}_{000}$ has all its components nonnegative\tcp*[r]{See \cref{fig:AdditionalDataPackets.StdForm.StdForm,fig:AdditionalDataPackets.RefForm.StdForm}}
	$\mathcal{Q} \leftarrow \mathcal{Q} \cup \{\mathcal{q}_1$\}\;
	\BlankLine
	
	let $\mathcal{q}_2$ be the reflected $\mathcal{q}_1$ about the plane $(x - x_\mathcal{n}) - (y - y_\mathcal{n}) = 0$\tcp*[r]{See \cref{fig:AdditionalDataPackets.StdForm.RefForm,fig:AdditionalDataPackets.RefForm.RefForm}}
	$\mathcal{Q} \leftarrow \mathcal{Q} \cup \{\mathcal{q}_2$\}\;
}
\BlankLine

$\mathcal{P} \leftarrow \mathcal{P} \cup \mathcal{Q}$\;
\Return $\mathcal{P}$\;

\caption{$\mathcal{P} \leftarrow$ {\tt GenerateStdDataPackets(}$\mathcal{p}_0$, $\vv{x}_\mathcal{n}${\tt )}: Generate all standard variations of an input data packet.}
\label{alg:GenerateStdDataPackets}
\end{algorithm}


\begin{algorithm}[!t]
\SetAlgoLined
\SetKwFunction{numcurvatures}{NumCurvatures}
\SetKwFunction{interpolate}{Interpolate}
\SetKwFunction{collectfeatures}{CollectFeatures}
\SetKwFunction{preprocess}{Preprocess}
\SetKwFunction{sign}{Sign}
\SetKwFunction{generatestddatapackets}{GenerateStdDataPackets}

\KwIn{$\mathcal{n}$: node object; $\mathcal{F}_\kappa^{ns}(\cdot)$ and $\mathcal{F}_\kappa^{sd}(\cdot)$: error-correcting neural networks for non-saddle and saddle samples; $\vv{\phi}$: nodal level-set values; $\hat{N}$: nodal unit normal vectors; $h$: mesh size; $h\kappa_{\min}^{low}$: minimum $|h\kappa|$ to enable neural inference on non-saddle samples; $h\kappa_{\min}^{up}$: $|h\kappa|$'s upper bound for blending numerical with neurally corrected approximation on non-saddle samples; $h^2\kappa_{G,\min}^{ns}$: minimum dimensionless Gaussian curvature to classify a sample as a non-saddle one.}
\KwResult{$h\kappa^\star$: dimensionless mean curvature at $\mathcal{n}$'s nearest location on the interface.}
\BlankLine

$(K,\, K_G) \leftarrow$ \numcurvatures{\normalfont $\mathcal{n}.\texttt{stencil}$, $\vv{\phi}$, $\hat{N}$}\tcp*[r]{Use $\mathcal{n}$'s stencil and \cref{eq:MeanCurvature.3d,eq:GaussianCurvature.Ext}}
$\vv{x}_\mathcal{n}^\Gamma \leftarrow \mathcal{n}.\vv{x} - \vv{\phi}[\mathcal{n}]\hat{N}[\mathcal{n}]$\tcp*[r]{to estimate $\kappa$ and $\kappa_G$.  See \cref{eq:NormalProjection}}
$h\kappa \leftarrow h \cdot$\interpolate{$\mathcal{G}$, $K$, $\vv{x}_\mathcal{n}^\Gamma$}; $\quad$ 
$h^2\kappa_G \leftarrow h^2 \cdot$\interpolate{$\mathcal{G}$, $K_G$, $\vv{x}_\mathcal{n}^\Gamma$}\tcp*[r]{See Algorithm \href{https://www.sciencedirect.com/science/article/pii/S002199911630242X\#fg0050}{2} in \cite{Mirzadeh;etal:16:Parallel-level-set}}
\BlankLine
			
$h\kappa^\star \leftarrow h\kappa$\tcp*[r]{Enable neural correction for $h\kappa$ only if needed}
\eIf{$h^2\kappa_G \geqslant h^2\kappa_{G,\min}^{ns}$}{
	\If{$|h\kappa| < h\kappa_{\min}^{low}$}{
		\Return $h\kappa^\star$\tcp*[r]{Do nothing for non-saddle samples with $h\kappa \approx 0$}
	}
	$\texttt{IsNonSaddle} \leftarrow \texttt{True}$\;
	$\mathcal{F}_\kappa(\cdot) \leftarrow \mathcal{F}_\kappa^{ns}(\cdot)$\tcp*[r]{Use model for non-saddle samples}
}{
	$\texttt{IsNonSaddle} \leftarrow \texttt{False}$\;
	$\mathcal{F}_\kappa(\cdot) \leftarrow \mathcal{F}_\kappa^{sd}(\cdot)$\tcp*[r]{Use model for saddle samples}
}
\BlankLine

$\mathcal{p} \leftarrow$ \collectfeatures{\normalfont $\mathcal{n}.\texttt{stencil}$, $\vv{\phi}$, $\hat{N}$}\tcp*[r]{Populate $\mathcal{p}$ (see \cref{eq:DataPacket})}
$\mathcal{p}.h\kappa \leftarrow h\kappa; \quad \mathcal{p}.h^2\kappa_G \leftarrow h^2\kappa_G$\;
\lIf{\normalfont $\texttt{IsNonSaddle}$}{transform $\mathcal{p}$ so that $\mathcal{p}.h\kappa$ becomes negative}
\BlankLine

$\mathcal{P} \leftarrow$ \generatestddatapackets{$\mathcal{p}$, $\mathcal{n}.\vv{x}$}\tcp*[r]{See \Cref{alg:GenerateStdDataPackets}}
$h\kappa_\mathcal{F} \leftarrow 0$\;
\ForEach{standard-formed data packet $\mathcal{q} \in \mathcal{P}$}{
	$h\kappa_\mathcal{F} \leftarrow h\kappa_\mathcal{F} + \frac{1}{6}\mathcal{F}_\kappa([$\preprocess{$\mathcal{q}$, $h$}, $\mathcal{q}.h\kappa])$\tcp*[r]{Average six predictions (see also \Cref{alg:Preprocess})}
}
\BlankLine

\eIf{\normalfont $\texttt{IsNonSaddle}$}{\label{alg:MLCurvature.BlendStart}
	\If{$|h\kappa| \leqslant h\kappa_{\min}^{up}$}{
		$\lambda \leftarrow (h\kappa_{\min}^{up} - |\mathcal{p}.h\kappa|)/(h\kappa_{\min}^{up} - h\kappa_{\min}^{low})$\tcp*[r]{Blend neural/numerical estimations if $|h\kappa|$ nears zero}
		$h\kappa_\mathcal{F} = (1 - \lambda)h\kappa_\mathcal{F} + \lambda(\mathcal{p}.h\kappa)$\;
	}
	$h\kappa^\star \leftarrow$ \sign{$h\kappa$}$\cdot |h\kappa_\mathcal{F}|$\tcp*[r]{Restore sign for the non-saddle-sample prediction}
}{
	$h\kappa^\star \leftarrow h\kappa_\mathcal{F}$\tcp*[r]{Saddle samples require no linear blending with $h\kappa$}
}\label{alg:MLCurvature.BlendEnd}
\BlankLine

\Return $h\kappa^\star$\;

\caption{$h\kappa^\star \leftarrow$ {\tt MLCurvature(}$\mathcal{n}$, $\mathcal{F}_\kappa^{ns}(\cdot)$, $\mathcal{F}_\kappa^{sd}(\cdot)$, $\vv{\phi}$, $\hat{N}$, $h$, $h\kappa_{\min}^{low}$, $h\kappa_{\min}^{up}$, $h^2\kappa_{G,\min}^{ns}${\tt )}: Compute the dimensionless mean curvature for an interface node $\mathcal{n}$ using standard numerical schemes with error correction provided by a neural network.}
\label{alg:MLCurvature}
\end{algorithm}

We now introduce our hybrid mean-curvature solver, which yields $h\kappa^\star$ for an interface node $\mathcal{n}$ with a complete, $h$-uniform stencil.  The listing, shown in \Cref{alg:MLCurvature}, integrates traditional numerical schemes with the machine learning ingredients we got acquainted with above.  This routine assumes that we have already fitted $\mathcal{F}_\kappa^{ns}(\cdot)$ and $\mathcal{F}_\kappa^{sd}(\cdot)$ to their respective training sets, as described in \Cref{subsec:Training}.  In the same manner, it expects precomputed nodal level-set values $\vv{\phi}$ and unit normal vectors $\hat{N}$,\footnote{We distinguish $m$-element vectors from 3-by-$m$ matrices by using bold lowercase symbols (e.g., $\vv{\phi}$) and normal-weight uppercase letters (e.g., $\hat{N}$).  Here, $m$ denotes the number of grid points in $\mathcal{G}$.  We have realized these abstractions using ghosted, parallel {\tt PETSc} \cite{Balay;Brown;Buschelman;etal:12:PETSc-Web-page} vectors in {\tt p4est} \cite{Burstedde;Wilcox;Ghattas:11:p4est:-Scalable-Algo}.} evaluated at least on a uniform band of half-width $3h$ around $\Gamma$ (see \cref{eq:UniformBandCriterion}).  Our hybrid solver also requires $\vv{\phi}$ and $\hat{N}$ to be extracted from a reinitialized level-set field that resembles a signed distance function.  Other formal parameters include the mesh size $h$ and the $h\kappa_{\min}^{low}$ and $h\kappa_{\min}^{up}$ bounds for blending non-saddle averaged $h\kappa_\mathcal{F}$ predictions with their numerical approximations near zero.  Similarly, we need the dimensionless Gaussian curvature lower bound $h^2\kappa_{G,\min}^{ns}$ to discriminate non-saddle from saddle data packets.

\Cref{alg:MLCurvature} is straightforward.  In the beginning, we compute $\mathcal{n}$'s numerical mean and Gaussian curvatures using finite-difference schemes to discretize \cref{eq:GaussianCurvature.Ext,eq:MeanCurvature.3d} \cite{Chene;Min;Gibou:08:Second-order-accurat}.  Then, we resort to linear interpolation to estimate $h\kappa$ and $h^2\kappa_G$ at $\Gamma$ \cite{Mirzadeh;etal:16:Parallel-level-set}.  $h\kappa$ represents the \textit{baseline} we want to improve, and the next course of action depends on whether $\mathcal{n}$ belongs to a non-saddle or saddle region.  We resolve this classification subproblem by verifying if $h^2\kappa_G$ exceeds $h^2\kappa_{G,\min}^{ns}$, which we have set throughout this study to $\eten{-7}{-6}$ (see \ref{item:OBNonSaddle} and \ref{item:OBSaddle}).  This threshold has resulted from examining various data sets, like those in \cref{fig:PrepAnalyses} and \Cref{subsubsec:HypParaboloidalInterfaceDataSetConstruction}.  In such analyses, we have observed that $|\bar{\varepsilon}|$ remains bounded by $\mathcal{O}(10^{-4})$, even for interface grid points where $h^2\kappa_{G,\min}^{ns} \leqslant h^2\kappa_G < 0$.  This finding has allowed us to slide the non-saddle/saddle decision boundary and extend our symmetrized approach to improve accuracy near $h^2\kappa_G \approx 0$.

If $\mathcal{n}$ belongs to a non-saddle region, we can terminate \Cref{alg:MLCurvature} prematurely whenever $|h\kappa|$ drops below $h\kappa_{\min}^{low}$.  This threshold accounts for the ability of traditional numerical schemes to perform well along flat interface sectors.  Here, $h\kappa_{\min}^{low}$ is also equivalent to the learning hyperparameter $h\kappa_{\min} = 0.004$, which we have chosen in analogy to the two-dimensional curvature problem \cite{Larios;Gibou;KECNet2D;2022}.  Next, we select $\mathcal{F}_\kappa^{ns}(\cdot)$ or $\mathcal{F}_\kappa^{sd}(\cdot)$, depending on the pattern type.  For succinctness, {\tt MLCurvature()} references to the appropriate MLP through the $\mathcal{F}_\kappa(\cdot)$ alias, which processes a transformed version of $\mathcal{n}$'s data packet $\mathcal{p}$.  To assemble $\mathcal{p}$, we employ the {\tt CollectFeatures()} module, which reads in and structures field data, as provided in \cref{eq:DataPacket}.  Then, we generate its six standard forms (see \crefrange{fig:DataPackets.StdForm}{fig:AdditionalDataPackets}) with \Cref{alg:GenerateStdDataPackets} while taking special care of the exclusive negative-mean-curvature normalization for non-saddle stencils.

The last portion of \Cref{alg:MLCurvature} carries out inference and symmetry enforcement.  In the former task, we evaluate $\mathcal{F}_\kappa(\cdot)$ on $\mathcal{p}$'s preprocessed standard forms and average the corresponding corrected mean curvatures into $h\kappa_\mathcal{F}$.  Afterward, {\tt MLCurvature()} finishes as soon as we find $h\kappa_\mathcal{F}$ for saddle stencils.  Non-saddle regions, on the other hand, require some post-processing before outputting $h\kappa^\star$.  To improve non-saddle accuracy and prevent sharp transitions near zero, we blend $h\kappa$ and $h\kappa_\mathcal{F}$ linearly for $-h\kappa \in [h\kappa_{\min}^{low},\, h\kappa_{\min}^{up}]$, where $h\kappa_{\min}^{up} = 0.007$.  Finally, we restore the expected mean-curvature sign from $h\kappa$ and exit \Cref{alg:MLCurvature} by returning $h\kappa^\star$ to the calling routine.

The following subsections describe our methodology to assemble $\mathcal{D}^{ns}$ and $\mathcal{D}^{sd}$---the training sets for $\mathcal{F}_\kappa^{ns}(\cdot)$ and $\mathcal{F}_\kappa^{sd}(\cdot)$.  Also, we provide implementation details to train, instantiate and deploy our MLPs to numerical applications.


\colorsubsection{Training}
\label{subsec:Training}

Although generating training data sets for our MLPs is an expensive endeavor, one must carry out this task just once.  The idea is to construct $\mathcal{D}^{ns}$ and $\mathcal{D}^{sd}$ by following randomized strategies that guarantee well-balanced $h\kappa^*$ distributions.

Our data-generating approach extends the algorithms described in \cite{Larios;Gibou;KECNet2D;2022, VOFCurvature3DML19} to three-dimensional level-set schemes.  In \cite{Larios;Gibou;KECNet2D;2022}, we developed a systematic yet scalable framework based on non-dimensionalizing the curvature-inference problem.  Once more, we have adopted that methodology to confine $\mathcal{F}_\kappa^{ns}(\cdot)$ and $\mathcal{F}_\kappa^{sd}(\cdot)$ to (absolute) dimensionless mean curvatures between $h\kappa_{\min}^*$ and $h\kappa_{\max}^*$,\footnote{It is recommended to enforce $0 < h\kappa_{\min}^* \leqslant h\kappa_{\min}^{low}$ to improve $\mathcal{F}_\kappa^{ns}(\cdot)$'s interpolative power near the upper end of the negative mean-curvature spectrum.  For us, $h\kappa_{\min}^*$ and $h\kappa_{\min}^{low}$ are equivalent.} where $h\kappa_{\min}^* = 0$ for saddle data packets (see \cref{fig:PrepAnalyses}).  Together, non-dimensionalization and preprocessing have made it possible to train a single pair of MLPs on one mesh size and port the optimized models across resolutions with marginal accuracy variations.

Next, we examine three kinds of training surfaces to build $\mathcal{D}^{ns}$ and $\mathcal{D}^{sd}$.  First, we consider \textit{spherical patches} to extract non-saddle learning tuples, similar to \cite{VOFCurvature3DML19}.  Likewise, in analogy to \cite{Larios;Gibou;KECNet2D;2022}, we use three-dimensional \textit{sinusoidal waves} to capture a wider diversity of patterns and enrich the contents of $\mathcal{D}^{ns}$.  As noticed above, sinusoids can produce saddle stencils, and here we incorporate a subset of those samples into $\mathcal{D}^{sd}$.  Then, we evaluate \textit{hyperbolic paraboloids} as the third class of training surfaces.  In particular, hyperbolic paraboloids are to $\mathcal{D}^{sd}$ as spheres are to $\mathcal{D}^{ns}$.  These saddle interfaces are necessary because it is more difficult and costly to determine the correct sinusoidal parameters to populate $\mathcal{D}^{sd}$ on the entire interval $[-h\kappa_{\max}^*,\, +h\kappa_{\max}^*]$ for $h\kappa_{\max}^* \gg 0$.


\colorsubsubsection{Spherical-interface data-set construction}
\label{subsubsec:SphericalInterfaceDataSetConstruction}


\begin{algorithm}[!t]
\SetAlgoLined
\SetKwFunction{randlinspace}{RandLinspace}
\SetKwFunction{sphericallevelset}{SphericalLevelSet}
\SetKwFunction{generategrid}{GenerateGrid}
\SetKwFunction{evaluate}{Evaluate}
\SetKwFunction{reinitialize}{Reinitialize}
\SetKwFunction{addnoise}{AddNoise}
\SetKwFunction{computenormals}{ComputeNormals}
\SetKwFunction{numcurvatures}{NumCurvatures}
\SetKwFunction{getnodesnexttogamma}{GetNodesNextToGamma}
\SetKwFunction{collectfeatures}{CollectFeatures}
\SetKwFunction{randomsamples}{RandomSamples}
\SetKwFunction{generatestddatapackets}{GenerateStdDataPackets}

\KwIn{$\eta$: unit-cube octree's maximum level of refinement; $h\kappa_{\min}^*$ and $h\kappa_{\max}^*$: minimum and maximum $|h\kappa^*|$; {\tt NSph}: number of radii; {\tt NSamPerSph}: number of samples per radius; $\nu$: number of redistancing steps; $\epsilon_{rnd}$: amount of uniform noise to perturb $\phi$.}
\KwResult{$\mathcal{D}_{sp}$: spherical-interface data set.}
\BlankLine

$h \leftarrow 2^{-\eta}; \quad \kappa_{\min}^* \leftarrow h\kappa_{\min}^*/h; \quad \kappa_{\max}^* \leftarrow h\kappa_{\max}^*/h$\tcp*[r]{Mesh size and curvature bounds}
\BlankLine

$\mathcal{D}_{sp} \leftarrow \varnothing$\;
$\texttt{TgtK} \leftarrow$ \randlinspace{\normalfont $\kappa_{\max}^*$, $\kappa_{\min}^*$, $\texttt{NSph}$}\tcp*[r]{Uniformly distributed $\kappa^* \in [\kappa_{\min}^*, \kappa_{\max}^*]$}
\ForEach{\normalfont $\kappa^* \in \texttt{TgtK}$}{
	$r_{sp} \leftarrow 1/\kappa^*$\tcp*[r]{Radius and random center $\vv{x}_{sp}=[x_{sp},\, y_{sp},\, z_{sp}]^T$}
	$\vv{x}_{sp} \sim \vv{\mathcal{U}}(-h/2, +h/2) \in \mathbb{R}^3$\;
	$\phi_{sp}(\cdot) \leftarrow$ \sphericallevelset{$\vv{x}_{sp}$, $r_{sp}$}\tcp*[r]{Level-set function from \cref{eq:SphericalLevelSetFunction}}
	\BlankLine
	
	$\vartheta \sim \mathcal{U}(0, 2\pi)$\tcp*[r]{Set up $\Omega$ around some point $\vv{p} \in \mathbb{R}^3$ on the sphere by using}\label{alg:GenerateSphericalDataSet.PointOnSphereStart}
	$\varphi \sim \arccos(2\cdot\mathcal{U}(0, 1) - 1)$\tcp*[r]{these random azimuthal and polar angles (see also \cite{VOFCurvature3DML19})}
	$\vv{p}_{sp} \leftarrow \left[x_{sp} + r_{sp}\cos\vartheta\sin\varphi,\quad y_{sp} + r_{sp}\sin\vartheta\sin\varphi,\quad z_{sp} + r_{sp}\cos\varphi\right]^T$\;\label{alg:GenerateSphericalDataSet.PointOnSphereEnd}
	let $\mathcal{B}_\Omega$ be the smallest box containing a sphere of radius $16h$ centered at $\roundh{\vv{p}_{sp}}$\;
	$\mathcal{G} \leftarrow$ \generategrid{\normalfont $\phi_{sp}(\cdot)$, $\eta$, $\mathcal{B}_\Omega$, $3h$}\tcp*[r]{Discretize $\Omega$ adaptively}
	\BlankLine
	
	$\vv{\phi} \leftarrow$ \evaluate{$\mathcal{G}$, $\phi_{sp}(\cdot)$}; \quad 
	$\vv{\phi} \leftarrow$ \addnoise{$\vv{\phi}$, $\epsilon_{rnd}\cdot\mathcal{U}(-h, +h)$}\tcp*[r]{Find noisy nodal level-set values}
	$\vv{\phi} \leftarrow$ \reinitialize{$\vv{\phi}$, $\nu$}\tcp*[r]{Solve \cref{eq:Reinitialization} with $\nu$ steps}

	$\hat{N} \leftarrow$ \computenormals{$\mathcal{G}$, $\vv{\phi}$}; \quad
	$(\vv{\kappa},\, \vv{\kappa}_G) \leftarrow$ \numcurvatures{$\mathcal{G}$, $\vv{\phi}$, $\hat{N}$}\tcp*[r]{See \cref{eq:NormalAndMeanCurvature,eq:MeanCurvature.3d,eq:GaussianCurvature.Ext}}
	\BlankLine
	
	$\mathcal{N} \leftarrow$ \getnodesnexttogamma{$\mathcal{G}$, $\vv{\phi}$}\tcp*[r]{Begin collecting samples}
	$\mathcal{S} \leftarrow \varnothing$\;
	\ForEach{node $\mathcal{n} \in \mathcal{N}$ with a complete, $h$-uniform stencil}{
			
		$\mathcal{p} \leftarrow$ \collectfeatures{\normalfont $\mathcal{n}.\texttt{stencil}$, $\vv{\phi}$, $\hat{N}$}\tcp*[r]{Populate $\mathcal{p}$ (see \cref{eq:DataPacket})}
		$\vv{x}_\mathcal{n}^\Gamma \leftarrow \mathcal{n}.\vv{x} - \vv{\phi}[\mathcal{n}]\hat{N}[\mathcal{n}]$\tcp*[r]{See \cref{eq:NormalProjection}}
		$\mathcal{p}.h\kappa \leftarrow h \cdot$\interpolate{$\mathcal{G}$, $\vv{\kappa}$, $\vv{x}_\mathcal{n}^\Gamma$}; $\quad$ 
		$\mathcal{p}.h^2\kappa_G \leftarrow h^2 \cdot$\interpolate{$\mathcal{G}$, $\vv{\kappa}_G$, $\vv{x}_\mathcal{n}^\Gamma$}\tcp*[r]{See Algorithm \href{https://www.sciencedirect.com/science/article/pii/S002199911630242X\#fg0050}{2} in \cite{Mirzadeh;etal:16:Parallel-level-set}}
		transform $\mathcal{p}$ so that $\mathcal{p}.h\kappa$ becomes negative\;
		\BlankLine
		
		$\mathcal{P} \leftarrow$ \generatestddatapackets{$\mathcal{p}$, $\mathcal{n}.\vv{x}$}\tcp*[r]{See \Cref{alg:GenerateStdDataPackets}}
		\ForEach{standard-formed data packet $\mathcal{q} \in \mathcal{P}$}{
			let $\vv{\xi}$ be the tuple $\left(\mathcal{q},\, -h\kappa^*\right)$ with target output $-h\kappa^*$\;
			$\mathcal{S} \leftarrow \mathcal{S} \cup \{\vv{\xi}\}$\;
		}
	}
	\BlankLine
	
	extract {\tt NSamPerSph} tuples from $\mathcal{S}$ randomly and add them to $\mathcal{D}_{sp}$\;
}
\BlankLine

\Return $\mathcal{D}_{sp}$\;

\caption{$\mathcal{D}_{sp} \leftarrow$ {\tt GenerateSphericalDataSet(}$\eta$, $h\kappa_{\min}^*$, $h\kappa_{\max}^*$, {\tt NSph}, {\tt NSamPerSph}, $\nu$, $\epsilon_{rnd}${\tt )}: Generate a randomized data set with samples from spherical-interface level-set functions.}
\label{alg:GenerateSphericalDataSet}
\end{algorithm}

Spheres are our first kind of elementary learning surface.  We have chosen them because it is trivial to calculate their exact mean curvature as $\kappa = 1/r_{sp}$, where $r_{sp}$ is their nonzero radius.  Also, it is easy to compute normal distances to spheres, and it is straightforward to integrate them into level-set applications as splitting criteria (see \crefrange{eq:RefinementCriterion}{eq:UniformBandCriterion}).  Formally, the level-set field

\begin{equation}
\phi_{sp}(\vv{x}) = \|\vv{x} - \vv{x}_{sp}\| - r_{sp}
\label{eq:SphericalLevelSetFunction}
\end{equation}
defines a signed distance function with a spherical interface centered at $\vv{x}_{sp} \in \mathbb{R}^3$, where $0 < \kappa_G = 1/r_{sp}^2$ for all points on $\Gamma$.  The latter statement makes spheres the surface of choice to populate $\mathcal{D}^{ns}$ in our training pipeline.

We have adapted the procedure to collect spherical training samples from the strategy suggested by Patel \etal \cite{VOFCurvature3DML19}.  In their research, the authors proposed to generate thousands of spheres with uniformly distributed random radii and extracted a single learning tuple from each $\Gamma$ at a random surface location.  We have modified those heuristics in two fundamental ways.  The first difference lies in how we space out $r_{sp}$ when defining $\phi_{sp}(\vv{x})$.  In this regard, we draw radii from randomly distributed mean curvatures instead of retrieving $\kappa^*$ from uniformly distributed radii.  Also, we produce more than one learning sample from each $r_{sp}$ while enforcing our problem-simplifying negative-mean-curvature normalization.  \Cref{alg:GenerateSphericalDataSet} combines these steps into a routine to build $\mathcal{D}_{sp}$---the set of spherical-interface learning pairs.  Ultimately, $\mathcal{D}^{ns}$ will take in $\mathcal{D}_{sp}$ alongside complementary data from the sinusoidal geometries discussed in \Cref{subsubsec:SinusoidalInterfaceDataSetConstruction}.

The list of formal parameters for \Cref{alg:GenerateSphericalDataSet} includes the maximum level of refinement, $\eta$, and the nonnegative target dimensionless mean-curvature bounds, $h\kappa_{\min}^*$ and $h\kappa_{\max}^*$.  In our case, we have chosen $h\kappa_{\min}^* = 0.004$ and $h\kappa_{\max}^* = 2/3$ across grid resolutions to non-dimensionalize the non-saddle portion of the problem.  Similarly, we have selected $2/3$ as the maximal $h\kappa^*$ because it is equivalent to the dimensionless mean curvature associated with the smallest resolvable sphere of radius $1.5h$.  In the same manner, we provide \Cref{alg:GenerateSphericalDataSet} with the number of distinct spheres to evaluate and how many samples we desire from each patch.  Here, we have chosen $\texttt{NSph} = \eten{2}{5}$ and $\texttt{NSamPerSph} = 10$ to populate $\mathcal{D}_{sp}$ with two million learning pairs.  At last, we supply $\nu$ and $\epsilon_{rnd}$: the number of redistancing steps and the amount of random uniform noise.  Later, we will employ these to alter the otherwise exact signed distance function in \cref{eq:SphericalLevelSetFunction}.

First, \Cref{alg:GenerateSphericalDataSet} solves for $h$ and the target curvature bounds.  Then, it uses these to populate the array $\texttt{TgtK}$ with $\texttt{NSph}$ decreasing uniform random $\kappa^*$ steps between $\kappa_{\max}^*$ and $\kappa_{\min}^*$.  

For each $\kappa^* \in \texttt{TgtK}$, we define its radius and use it to spawn a level-set field, $\phi_{sp}(\cdot)$, whose spherical interface centers randomly at $\vv{x}_{sp} \in (-h/2,\, +h/2)^3$.  Then, we find a couple of random (spherical-coordinated) angles, $\vartheta$ and $\varphi$, to locate a point $\vv{p}_{sp}$ on $\Gamma$.  The computation of $\vartheta$ and $\varphi$ replicates the techniques suggested by Patel \etal \cite{VOFCurvature3DML19}, who showed that the stochastic nature of these azimuthal and polar angles guarantees a uniform sampling coverage of the spherical surface.  $\vv{p}_{sp}$ allows us to create a $32h$-side-length bounding box $\mathcal{B}_\Omega$ centered at $\roundh{\vv{p}_{sp}}$, where $\roundh{\cdot}$ rounds each component to the nearest multiple of $h$.  Afterward, we use $\mathcal{B}_\Omega$ and $\phi_{sp}(\cdot)$ to discretize $\Omega$ with an adaptive Cartesian grid $\mathcal{G}$ that ensures a uniform band of half-width $3h$ around $\Gamma$.  Such a small bounding box helps cut down processing costs in $\mathbb{R}^3$, especially as $r_{sp} \to 1/\kappa_{\min}^*$.  Later, we will reap the benefits of minimal bounding boxes as we transition to interfaces with no analytical distance-to-point formulae.

The following statements in \Cref{alg:GenerateSphericalDataSet} evaluate \cref{eq:SphericalLevelSetFunction} on $\mathcal{G}$ and gather its nodal level-set values into the vector $\vv{\phi}$.  The latter contains exact signed distances, which we perturb in two ways to favor generalization.  First, we alter $\vv{\phi}$ by adding uniform noise, as in $\phi \leftarrow \phi + \epsilon_{rnd}\cdot\mathcal{U}(-h, +h)$, where $\epsilon_{rnd} = \eten{1}{-4}$.  Then, we reinitialize $\vv{\phi}$ by solving \cref{eq:Reinitialization} with $\nu = 10$ iterations, as in \cite{Larios;Gibou;HybridCurvature;2021, Larios;Gibou;KECNet2D;2022}.  Shortly after redistancing, we employ $\vv{\phi}$ to estimate nodal unit normal vectors, $\hat{N}$, and baseline curvatures, $\vv{\kappa}$ and $\vv{\kappa}_G$, to prepare data collection.

A loop traversing the set $\mathcal{N}$ of extracted interface nodes with regular, $h$-uniform stencils orchestrates the data collection subtask.  The traversal pattern is typical not only for data-generative routines but also for testing and deploying \Cref{alg:MLCurvature}.  Essentially, for each node $\mathcal{n} \in \mathcal{N}$, we start by structuring its features into a data packet $\mathcal{p}$ containing the (linearly) interpolated, non-dimensionalized curvatures at $\Gamma$ (see \cref{eq:DataPacket}).  Then, we normalize $\mathcal{p}$ so that $h\kappa < 0$ whenever the stencil belongs to a convex region and $\mathcal{p}.h^2\kappa_G \geqslant h^2\kappa_{G,\min}^{ns}$.  Furthermore, as remarked in \Cref{subsec:HybridMeanCurvatureSolverAndECNets}, any interface grid point can give rise to six standard-formed data packets through \Cref{alg:GenerateStdDataPackets} (see \crefrange{fig:DataPackets.StdForm}{fig:AdditionalDataPackets}).  Hence, one must generate a tuple $\vv{\xi}$ for every variation $\mathcal{q} \in \mathcal{P}$, having all different input patterns map to the same $h\kappa^*$ output.  In \Cref{alg:GenerateSphericalDataSet}, these tuples are accumulated in the auxiliary set $\mathcal{S}$.  And, eventually, we pull $\texttt{NSamPerSph}$ random samples from it and add them to $\mathcal{D}_{sp}$.  \Cref{alg:GenerateSphericalDataSet} thus completes after exhausting the entries in $\texttt{TgtK}$.  Upon exiting, the calling routine receives $\mathcal{D}_{sp}$ with $\texttt{NSph}\cdot\texttt{NSamPerSph}$ learning samples that exhibit a well-balanced frequency distribution on the $h\kappa^*$-axis.


\FloatBarrier
\colorsubsubsection{Sinusoidal-interface data-set construction}
\label{subsubsec:SinusoidalInterfaceDataSetConstruction}

\begin{figure}[!t]
	\centering
	\includegraphics[width=7.5cm]{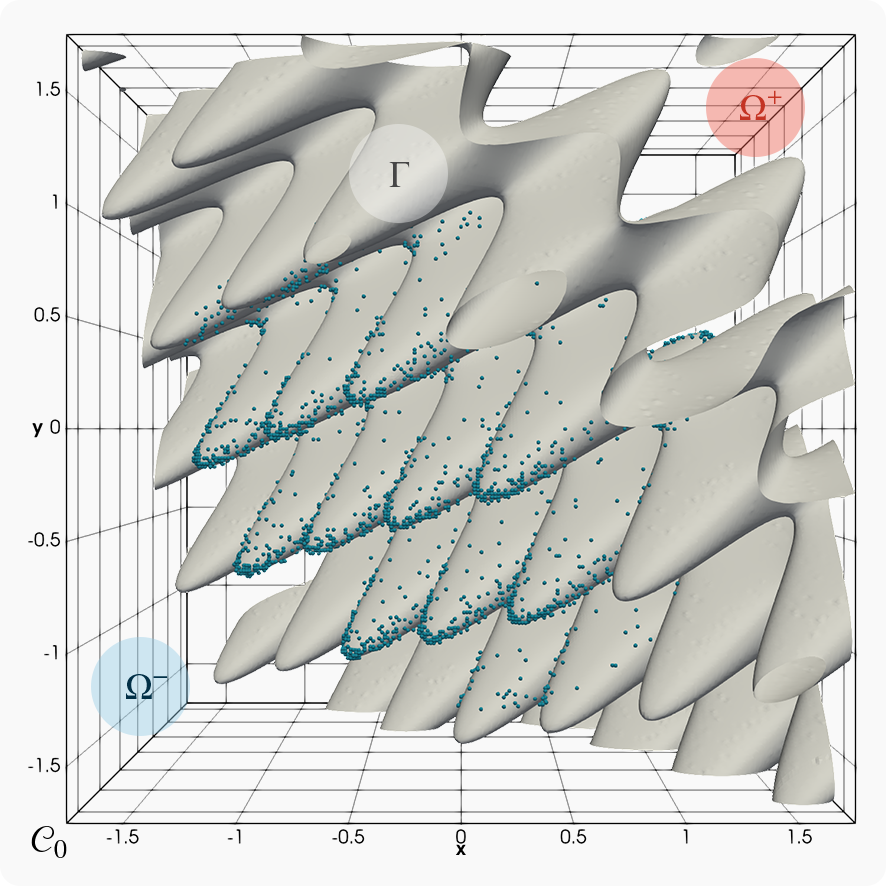}
	\caption{An affine-transformed sinusoidal wave $\vv{\qsn(\cdot)}$ embedded in $\mathcal{C}_0$, where $\vv{x}_{sn} = h~[-1/3,\, 1/4,\, -1/5]^T$, $\theta_{sn} = 11\pi/36$, $\hat{\vv{e}}_{sn} = \sqrt{2}~[1/2,\, -1/2,\, 0]^T$, and $h = 1/64$.  The small blue dots represent randomly sampled interface grid points with special emphasis along steep $\Gamma$ regions.  (Color online.)}
	\label{fig:SinusoidalExample}
\end{figure}

Three-dimensional sinusoidal waves are our second type of elementary surface.  They are the analog to the planar sinusoids first considered in \cite{Larios;Gibou;HybridCurvature;2021}.  Unlike spherical patches \cite{VOFCurvature3DML19}, sinusoidal waves are critical for training $\mathcal{F}_\kappa^{ns}(\cdot)$ and $\mathcal{F}_\kappa^{sd}(\cdot)$ because we can use them to extract non-saddle and saddle samples, as seen in \cref{fig:PrepAnalyses}.  

To better organize the generative process, we have chosen two separate data sets, $\mathcal{D}_{sn}^{ns}$ and $\mathcal{D}_{sn}^{sd}$, which prevent mixing non-saddle and saddle patterns.  Their learning tuples originate from level-set functions characterized by sinusoidal interfaces.  These waves embedded in $\mathbb{R}^3$ are parametrized as Monge patches $\vv{q}(u,v) : \mathbb{R}^2 \mapsto \mathbb{R}^3$ \cite{ModernDifferentialGeometry;2006} of the form

\begin{equation}
\vv{\qsn}(u,v) \doteq (u, v, \qsn(u,v)) \quad \textrm{with} \quad \qsn(u,v) = A_{sn}\sin(\omega_1 u)\sin(\omega_2 v),
\label{eq:SinusoidalInterface}
\end{equation}
where $A_{sn}$ denotes the amplitude, and $\omega_1$ and $\omega_2$ are the frequencies in the $u$ and $v$ directions.  Expressing a sinusoidal surface in this manner allows us to calculate the exact mean and Gaussian curvatures for any $\vv{p} = [u,\, v]^T \in \mathbb{R}^2$ with

\begin{equation}
\kappa_{sn}(\vv{p}) = -\qsn(\vv{p})\frac{(1+\qsn_v^2(\vv{p}))\omega_1^2 + 2\qsn_{uv}^2(\vv{p}) + (1+\qsn_u^2(\vv{p}))\omega_2^2}{2(1 + \qsn_u^2(\vv{p}) + \qsn_v^2(\vv{p}))^{3/2}} \quad \textrm{and} \quad
\kappa_{G,sn}(\vv{p}) = \frac{\omega_1^2 \omega_2^2 \qsn^2(\vv{p}) - \qsn_{uv}^2(\vv{p})}{(1 + \qsn_u^2(\vv{p}) + \qsn_v^2(\vv{p}))^2},
\label{eq:SinusoidalCurvatures}
\end{equation}
where $\qsn_u(u,v) = A_{sn}\omega_1 \cos(\omega_1 u)\sin(\omega_2 v)$ and $\qsn_v(u,v) = A_{sn}\omega_2 \sin(\omega_1 u)\cos(\omega_2 v)$.  

Given a sinusoid in its \textit{canonical} frame (i.e., as provided in \cref{eq:SinusoidalInterface}), we can construct the level-set function

\begin{equation}
\phi_{sn}(\vv{x}) = \left\{\begin{array}{ll}
	-\dist{\vv{x}}{\vv{\qsn}(\cdot)} & \textrm{if } z > \qsn(x,y), \\
	0                                & \textrm{if } z = \qsn(x,y) \;(\textrm{i.e., } \vv{x} \in \Gamma), \\
	+\dist{\vv{x}}{\vv{\qsn}(\cdot)} & \textrm{if } z < \qsn(x,y), 
\end{array}\right.
\label{eq:SinusoidalLevelSetFunction}
\end{equation}
where $\dist{\cdot}{\cdot}$ yields the shortest distance between a point $\vv{x} \in \mathbb{R}^3$ and a surface.  This definition of $\phi_{sn}(\vv{x})$ partitions the computational domain so that $\Omega^-$ lies strictly above $\Gamma \equiv \vv{\qsn}(\cdot)$, and $\Omega^+$ is found right beneath it.

The procedure to populate $\mathcal{D}_{sn}^{ns}$ and $\mathcal{D}_{sn}^{sd}$ requires fiddling with $A_{sn}$, $\omega_1$, and $\omega_2$ to gather an assortment of learning tuples.  However, additional pattern variations are possible if one introduces affine transformations \cite{CGUsingOpenGL01} to perturb the canonical patch in \cref{eq:SinusoidalInterface}.  In this work, we have constrained ourselves to rotating and translating the local coordinate system, $\mathcal{C}_{sn}$, that holds $\vv{\qsn}(\cdot)$.  More specifically, if $T(\vv{x}_{sn}) \in \mathbb{R}^{4\times 4}$ is the translation induced by a shift vector $\vv{x}_{sn}$, and $R(\hat{\vv{e}}_{sn}, \theta_{sn}) \in \mathbb{R}^{4\times 4}$ is the rotation by an angle $\theta_{sn}$ about some unit axis $\hat{\vv{e}}_{sn}$, then

\begin{equation}
 _0M_{sn} = T(\vv{x}_{sn}) R(\hat{\vv{e}}_{sn}, \theta_{sn})
\label{eq:AffineTransform}
\end{equation}
maps points and vectors in \textit{homogeneous coordinates}\footnote{Points and vectors in homogenous coordinates are suffixed with 1 and 0, respectively.  Thus, $[x,\, y,\, z,\, 1]^T$ and $[x,\, y,\, z,\, 0]^T$ denote the point and vector versions of $\vv{x} = [x,\, y,\, z]^T \in \mathbb{R}^3$.  This is also the reason why three-dimensional affine transformations are $4$-by-$4$ matrices.  Refer to \cite{CGUsingOpenGL01} for more details.} from $\mathcal{C}_{sn}$ into their representation within the world frame $\mathcal{C}_0$.  With similar reasoning, we can derive the inverse transformation 

\begin{equation}
 _{sn}M_0 = R^T(\hat{\vv{e}}_{sn}, \theta_{sn}) T(-\vv{x}_{sn}),
\label{eq:InverseAffineTransform} 
\end{equation}
which maps points and vectors expressed in $\mathcal{C}_0$ coordinates in terms of the sine wave's canonical frame.  \Cref{fig:SinusoidalExample} shows a sinusoid and its sampled nodes under the effects of these rigid-body transformations.

\Cref{eq:InverseAffineTransform} is indispensable for keeping \cref{eq:SinusoidalLevelSetFunction}'s evaluation simple, especially when finding a query point's perpendicular projection to compute $\dist{\vv{x}}{\vv{\qsn}(\cdot)}$.  For instance, if $\vv{x}$ is an arbitrary nodal position provided in $\mathcal{C}_0$ (e.g., one of those in \cref{fig:Stencil} or \cref{fig:SinusoidalExample}), we can find its level-set value with ease as $\phi_{sn}(\cancel{\llbracket}_{sn}M_0\llbracket\vv{x}\rrbracket\cancel{\rrbracket})$.  Likewise, we can compute the target mean and Gaussian curvatures of an interface node $\mathcal{n}$ by estimating $\vv{x}_\mathcal{n}^\perp = [u^*,\, v^*,\, \qsn(u^*, v^*)]^T$ from $\cancel{\llbracket}_{sn}M_0\llbracket\vv{x}_\mathcal{n}\rrbracket\cancel{\rrbracket}$ first and then using $\vv{p}^* = [u^*,\, v^*]^T$ in \cref{eq:SinusoidalCurvatures}.  Here, the operators $\llbracket\cdot\rrbracket$ and $\cancel{\llbracket}\cdot\cancel{\rrbracket}$ append and remove homogeneous suffixes.


\begin{algorithm}[!t]
\SetAlgoLined
\SetKwFunction{randlinspace}{RandLinspace}
\SetKwFunction{sinusoidallevelset}{SinusoidalLevelSet}
\SetKwFunction{generategrid}{GenerateGrid}
\SetKwFunction{evaluate}{Evaluate}
\SetKwFunction{addnoise}{AddNoise}
\SetKwFunction{reinitialize}{Reinitialize}
\SetKwFunction{buildrandombasis}{BuildRandomBasis}
\SetKwFunction{collectsinusoidalsamples}{CollectSinusoidalSamples}

\KwIn{$\eta$: maximum level of refinement per unit-cube octree; $h\kappa_{\min}^*$ and {\tt minHKPr}: minimum $|h\kappa^*|$ and its easing probability (for non-saddle samples); $h\kappa_{\max}^*$, {\tt maxHKLowPr}, and {\tt maxHKUpPr}: maximum $|h\kappa^*|$ and its lower and upper bound easing probabilities (for non-saddle samples); $h^2\tilde{\kappa}_{G,\min}$ and {\tt minIH2KGPr}: minimum $|h^2\kappa_G|$ and its easing probability (for saddle samples); $h^2\tilde{\kappa}_{G,\max}^{low}$ and {\tt maxIH2KGPr}: lower bound for maximum $|h^2\kappa_G|$ and its easing probability (for saddle samples); {\tt NA}: number of amplitudes; {\tt NT}: number of discrete rotation angles; {\tt Nhk}: number of steps to vary the expected maximum curvature at the crests; $\nu$: number of iterations for level-set reinitialization; $\epsilon_{rnd}$: amount of uniform noise to perturb $\phi$; $h^2\kappa_{G,\min}^{ns}$: minimum $h^2\kappa_G$ to classify a sample as a non-saddle one.}
\KwResult{$\mathcal{D}_{sn}^{ns}$ and $\mathcal{D}_{sn}^{sd}$: non-saddle and saddle sinusoidal-interface samples.}
\BlankLine

$h \leftarrow 2^{-\eta}; \quad
\kappa_{\min}^* \leftarrow h\kappa_{\min}^*/h; \quad 
\kappa_{\max}^* \leftarrow h\kappa_{\max}^*/h$\tcp*[r]{Mesh size and mean-curvature bounds}
$A_{\min} \leftarrow 5/\kappa_{\max}^*; \quad A_{\max} \leftarrow 1/(2\kappa_{\min}^*)$\tcp*[r]{Amplitude bounds}
$h\kappa_{\max}^{low} \leftarrow \frac{1}{2}h\kappa_{\max}^*; \quad h\kappa_{\max}^{up} \leftarrow h\kappa_{\max}^*$\tcp*[r]{Bounds for maximum curvature at the crests}
\BlankLine

$\mathcal{D}_{sn}^{ns} \leftarrow \varnothing; \quad 
\mathcal{D}_{sn}^{sd} \leftarrow \varnothing$\tcp*[r]{Sets for non-saddle and saddle samples}
$\texttt{hkMax} \leftarrow$ \randlinspace{\normalfont $h\kappa_{\max}^{low}$, $h\kappa_{\max}^{up}$, $\texttt{Nhk}$}\tcp*[r]{Array of $|h\kappa^*|$ at the crests}
\ForEach{amplitude $A_{sn} \in$ \randlinspace{\normalfont $A_{\min}$, $A_{\max}$, $\texttt{NA}$}}{
	\BlankLine
	
	$\mathcal{B}_A^{ns} \leftarrow \varnothing; \quad 
	\mathcal{B}_A^{sd} \leftarrow \varnothing$\tcp*[r]{Sample buffers for current $A_{sn}$}
	\For{\normalfont $s \leftarrow 0, ..., \texttt{Nhk}-1$}{\label{alg:GenerateSinusoidalDataSets.sLoop}
		$\kappa_s \leftarrow \texttt{hkMax}[s]/h; \quad \omega_1 \leftarrow \sqrt{\kappa_s / A_{sn}}$\;
		\BlankLine
		
		\For{\normalfont $t \leftarrow s, ..., \texttt{Nhk}-1$}{\label{alg:GenerateSinusoidalDataSets.tLoop}
			$\kappa_t \leftarrow \texttt{hkMax}[t]/h$\tcp*[r]{Go from $\kappa_s$ to $\kappa_t$ while transitioning}
			$\omega_2 \leftarrow \sqrt{2\kappa_t / A_{sn} - \omega_1^2}$\tcp*[r]{from circular to elliptical isocontours}\label{alg:GenerateSinusoidalDataSets.w2Definition}
			
			$\mathcal{E} \leftarrow$ \buildrandombasis{}\tcp*[r]{A random orthonormal basis for $\mathbb{R}^3$}
			\ForEach{unit vector $\hat{\vv{e}}_{sn} \in \mathcal{E}$}{
				\BlankLine
				
				\ForEach{angle $\theta_{sn} \in$ \randlinspace{\normalfont $-\pi/2$, $+\pi/2$, $\texttt{NT}$}, excluding right endpoint}{
					$\vv{x}_{sn} \sim \vv{\mathcal{U}}(-h/2, +h/2)$\tcp*[r]{Random shift from the origin}
					$r_{sam} \leftarrow 6h + \min(1.5A_{\max},\, \max( A_{sn},\, 2\cdot 2\pi\cdot \max(1/\omega_1,\, 1/\omega_2)))$\tcp*[r]{Sampling radius}
					let $\mathcal{B}_\Omega$ be the smallest cube enclosing a sphere of radius $r_{sam}$\;
					$\phi_{sn}(\cdot) \leftarrow$ \sinusoidallevelset{\normalfont $A_{sn}$, $\omega_1$, $\omega_2$, $\vv{x}_{sn}$, $\hat{\vv{e}}_{sn}$, $\theta_{sn}$}\tcp*[r]{See \cref{eq:SinusoidalLevelSetFunction}.  Triangulate $\Gamma$}
					$\mathcal{G} \leftarrow$ \generategrid{\normalfont $\phi_{sn}(\cdot)$, $\eta$, $\mathcal{B}_\Omega$, $3h$}\tcp*[r]{in \cref{eq:SinusoidalInterface} to discretize $\Omega$}
					\BlankLine
	
					$\vv{\phi} \leftarrow$ \evaluate{\normalfont $\mathcal{G}$, $\phi_{sn}(\cdot)$}; \quad 
					$\vv{\phi} \leftarrow$ \addnoise{\normalfont $\vv{\phi}$, $\epsilon_{rnd}\cdot\mathcal{U}(-h, +h)$}\tcp*[r]{Compute level-set values}
					$\vv{\phi} \leftarrow$ \reinitialize{\normalfont $\vv{\phi}$, $\nu$}\tcp*[r]{and solve \cref{eq:Reinitialization} with $\nu$ iterations}
					\BlankLine
					
					$(\mathcal{S}^{ns},\, \mathcal{S}^{sd}) \leftarrow$ \collectsinusoidalsamples{\normalfont $\mathcal{G}$, $h$, $\vv{\phi}$, $\phi_{sn}(\cdot)$, $h\kappa_{\min}^*$, {\tt minHKPr}, $h\kappa_{\max}^{low}$, {\tt maxHKLowPr}, $h\kappa_{\max}^{up}$, {\tt maxHKUpPr}, $h^2\tilde{\kappa}_{G,\min}$, {\tt minIH2KGPr}, $h^2\tilde{\kappa}_{G,\max}^{low}$, {\tt maxIH2KGPr}, $h^2\kappa_{G,\min}^{ns}$, $r_{sam}$}\tcp*[r]{See \Cref{alg:CollectSinusoidalSamples}}\label{alg:GenerateSinusoidalDataSets.CallingCollectSamples}
					$\mathcal{B}_A^{ns} \leftarrow \mathcal{B}_A^{ns} \cup \mathcal{S}^{ns}; \quad
					\mathcal{B}_A^{sd} \leftarrow \mathcal{B}_A^{sd} \cup \mathcal{S}^{sd}$\;
				}
			}
		}
	}
	
	use histogram-based subsampling to balance $\mathcal{B}_A^{ns}$ and $\mathcal{B}_A^{sd}$\;
	$\mathcal{D}_{sn}^{ns} \leftarrow \mathcal{D}_{sn}^{ns} \cup \mathcal{B}_A^{ns}; \quad
	\mathcal{D}_{sn}^{sd} \leftarrow \mathcal{D}_{sn}^{sd} \cup \mathcal{B}_A^{sd}$\;
}

use histogram-based subsampling to balance $\mathcal{D}_{sn}^{ns}$ and $\mathcal{D}_{sn}^{sd}$\;
\Return $(\mathcal{D}_{sn}^{ns},\, \mathcal{D}_{sn}^{sd})$\;

\caption{$(\mathcal{D}_{sn}^{ns},\, \mathcal{D}_{sn}^{sd}) \leftarrow$ {\tt GenerateSinusoidalDataSets(}$\eta$, $h\kappa_{\min}^*$, {\tt minHKPr}, $h\kappa_{\max}^*$, {\tt maxHKLowPr}, {\tt maxHKUpPr}, $h^2\tilde{\kappa}_{G,\min}$, {\tt minIH2KGPr}, $h^2\tilde{\kappa}_{G,\max}^{low}$, {\tt maxIH2KGPr}, {\tt NA}, {\tt NT}, {\tt Nhk}, $\nu$, $\epsilon_{rnd}$, $h^2\kappa_{G,\min}^{ns}${\tt )}: Generate randomized data sets with non-saddle and saddle samples from sinusoidal-interface level-set functions.}
\label{alg:GenerateSinusoidalDataSets}
\end{algorithm}


\begin{algorithm}[!t]
\SetAlgoLined
\SetKwFunction{computenormals}{ComputeNormals}
\SetKwFunction{numcurvatures}{NumCurvatures}
\SetKwFunction{getnodesnexttogamma}{GetNodesNextToGamma}
\SetKwFunction{ease}{Ease}
\SetKwFunction{collectfeatures}{CollectFeatures}
\SetKwFunction{generatestddatapackets}{GenerateStdDataPackets}

\KwIn{$\mathcal{G}$: grid; $h$: mesh size; $\vv{\phi}$: nodal level-set values; $\phi_{sn}(\cdot)$: sinusoidal-interface level-set function object; $h\kappa_{\min}^*$ and {\tt minHKPr}: minimum $|h\kappa^*|$ and its easing probability;  $h\kappa_{\max}^{low}$ and {\tt maxHKLowPr}: lower bound for maximum $|h\kappa^*|$ and its easing probability; $h\kappa_{\max}^{up}$ and {\tt maxHKUpPr}: upper bound for maximum $|h\kappa^*|$ and its easing probability; $h^2\tilde{\kappa}_{G,\min}$ and {\tt minIH2KGPr}: minimum $|h^2\kappa_G|$ and its easing probability; $h^2\tilde{\kappa}_{G,\max}^{low}$ and {\tt maxIH2KGPr}: lower bound for maximum $|h^2\kappa_G|$ and its easing probability; $h^2\kappa_{G,\min}^{ns}$: minimum $h^2\kappa_G$ to classify a sample as a non-saddle one; $r_{sam}$: sampling radius.}
\KwResult{$\mathcal{S}^{ns}$ and $\mathcal{S}^{sd}$: non-saddle and saddle samples.}
\BlankLine

$\hat{N} \leftarrow$ \computenormals{$\mathcal{G}$, $\vv{\phi}$}$; \quad
(\vv{\kappa},\, \vv{\kappa}_G) \leftarrow$ \numcurvatures{$\mathcal{G}$, $\vv{\phi}$, $\hat{N}$}\tcp*[r]{See \cref{eq:NormalAndMeanCurvature,eq:MeanCurvature.3d,eq:GaussianCurvature.Ext}}
\BlankLine

$\mathcal{S}^{ns} \leftarrow \varnothing; \quad 
\mathcal{S}^{sd} \leftarrow \varnothing$\;			
$\mathcal{N} \leftarrow$ \getnodesnexttogamma{$\mathcal{G}$, $\vv{\phi}$}\;
let $\vv{x}_{sn}$ be $\phi_{sn}(\cdot)$'s offset in terms of the world coordinate system\;
\ForEach{node $\mathcal{n} \in \mathcal{N}$ with a complete, $h$-uniform stencil}{\label{alg:CollectSinusoidalSamples.TraverseLoopStart}
	\BlankLine
	
	\lIf{$\|\mathcal{n}.\vv{x} - \vv{x}_{sn}\| > r_{sam}$ {\normalfont\textbf{or}} $\mathcal{n}.\vv{x}$ is less than $4h$ away from $\mathcal{G}$'s walls}{skip node $\mathcal{n}$}
	\BlankLine
	
	compute the target $h\kappa^*$ at the nearest point $\vv{x}_\mathcal{n}^\perp$ to $\mathcal{n}.\vv{x}$ on $\Gamma$\tcp*[r]{See \cref{eq:SinusoidalCurvatures}}
	$\vv{x}_\mathcal{n}^\Gamma \leftarrow \mathcal{n}.\vv{x} - \vv{\phi}[\mathcal{n}]\hat{N}[\mathcal{n}]$\tcp*[r]{See \cref{eq:NormalProjection}}
	$h\kappa \leftarrow h \cdot$\interpolate{$\mathcal{G}$, $\vv{\kappa}$, $\vv{x}_\mathcal{n}^\Gamma$}$; \quad
	h^2\kappa_G \leftarrow h^2 \cdot$\interpolate{$\mathcal{G}$, $\vv{\kappa}_G$, $\vv{x}_\mathcal{n}^\Gamma$}\tcp*[r]{See Algorithm \href{https://www.sciencedirect.com/science/article/pii/S002199911630242X\#fg0050}{2} in \cite{Mirzadeh;etal:16:Parallel-level-set}}
	\BlankLine
	
	\tcp{Different probabilistic filtering (see \cref{eq:Ease}) for non-saddle and saddle regions}
	\eIf{$h^2\kappa_G \geqslant h^2\kappa_{G,\min}^{ns}$}{
		\lIf{$|h\kappa^*| < h\kappa_{\min}^*$}{skip node $\mathcal{n}$}
		\eIf{$|h\kappa^*| \leqslant h\kappa_{\max}^{low}$}{
			\lIf{$\mathcal{U}(0, 1) >$ \ease{\normalfont $|h\kappa^*|$; $h\kappa_{\min}^*$, {\tt minHKPr}, $h\kappa_{\max}^{low}$, {\tt maxHKLowPr}}}{skip node $\mathcal{n}$}
		}{
			\lIf{$\mathcal{U}(0, 1) >$ \ease{\normalfont $|h\kappa^*|$; $h\kappa_{\max}^{low}$, {\tt maxHKLowPr}, $h\kappa_{\max}^{up}$, {\tt maxHKUpPr}}}{skip node $\mathcal{n}$}
		}
		$\texttt{IsNonSaddle} \leftarrow \texttt{True}$\;
		$h\kappa^* \leftarrow -|h\kappa|$\tcp*[r]{Anticipate negative-mean-curvature normalization}
	}{
		\lIf{$|h^2\kappa_G| < h^2\tilde{\kappa}_{G,\min}$}{skip node $\mathcal{n}$}
		\lIf{$\mathcal{U}(0, 1) >$ \ease{\normalfont $|h^2\kappa_G|$; $h^2\tilde{\kappa}_{G,\min}$, {\tt minIH2KGPr}, $h^2\tilde{\kappa}_{G,\max}^{low}$, {\tt maxIH2KGPr}}}{skip node $\mathcal{n}$}
		$\texttt{IsNonSaddle} \leftarrow \texttt{False}$\;
	}
	\BlankLine
	
	$\mathcal{p} \leftarrow$ \collectfeatures{\normalfont $\mathcal{n}.\texttt{stencil}$, $\vv{\phi}$, $\hat{N}$}\tcp*[r]{Populate $\mathcal{p}$ (see \cref{eq:DataPacket})}
	$\mathcal{p}.h\kappa \leftarrow h\kappa; \quad \mathcal{p}.h^2\kappa_G \leftarrow h^2\kappa_G$\;
	\lIf{\normalfont {\tt IsNonSaddle}}{transform $\mathcal{p}$ so that $\mathcal{p}.h\kappa$ becomes negative}
	\BlankLine
	
	$\mathcal{P} \leftarrow$ \generatestddatapackets{$\mathcal{p}$, $\mathcal{n}.\vv{x}$}\tcp*[r]{See \Cref{alg:GenerateStdDataPackets}}
	\ForEach{standard-formed data packet $q \in \mathcal{P}$}{
		let $\vv{\xi}$ be the learning tuple $\left(\mathcal{q},\, h\kappa^*\right)$ with inputs $\mathcal{q}$ and expected output $h\kappa^*$\;
		\eIf{\normalfont {\tt IsNonSaddle}}{
			$\mathcal{S}^{ns} \leftarrow \mathcal{S}^{ns} \cup \{\vv{\xi}\}$\;
		}{
			$\mathcal{S}^{sd} \leftarrow \mathcal{S}^{sd} \cup \{\vv{\xi}\}$\;
		}
	}
}\label{alg:CollectSinusoidalSamples.TraverseLoopEnd}
\BlankLine

\Return $(\mathcal{S}^{ns},\, \mathcal{S}^{sd})$\;

\caption{$(\mathcal{S}^{ns},\, \mathcal{S}^{sd}) \leftarrow$ {\tt CollectSinusoidalSamples(}$\mathcal{G}$, $h$, $\vv{\phi}$, $\phi_{sn}(\cdot)$, $h\kappa_{\min}^*$, {\tt minHKPr}, $h\kappa_{\max}^{low}$, {\tt maxHKLowPr}, $h\kappa_{\max}^{up}$, {\tt maxHKUpPr}, $h^2\tilde{\kappa}_{G,\min}$, {\tt minIH2KGPr}, $h^2\tilde{\kappa}_{G,\max}^{low}$, {\tt maxIH2KGPr}, $h^2\kappa_{G,\min}^{ns}$, $r_{sam}${\tt )}: Collect non-saddle and saddle samples along a sinusoidal surface.}
\label{alg:CollectSinusoidalSamples}
\end{algorithm}

\Cref{alg:GenerateSinusoidalDataSets} blends the above numerical ingredients within a combinatorial procedure that assembles $\mathcal{D}_{sn}^{ns}$ and $\mathcal{D}_{ns}^{sd}$.  As in \Cref{alg:GenerateSphericalDataSet}, we must provide $h\kappa_{\min}^* = 0.004$ and $h\kappa_{\max}^* = 2/3$, which we observe to both parametrize sinusoidal geometries and enforce subsampling protocols that contain flat-region over-representation.  \Cref{alg:GenerateSinusoidalDataSets} also expects the related probabilities $\texttt{minHKPr}$, $\texttt{maxHKLowPr}$, and $\texttt{maxHKUpPr}$.  They help us construct nonuniform probability distributions for non-saddle patterns where the odds of selecting learning tuples with $|h\kappa^*| \approx h\kappa_{\min}^*$ are far smaller than those of $|h\kappa^*| \approx h\kappa_{\max}^*$.  Analogously, the routine requires a couple of subsampling Gaussian curvature bounds, $h^2\tilde{\kappa}_{G,\min}$ and $h^2\tilde{\kappa}_{G,\max}^{low}$, alongside their respective probabilities, $\texttt{minIH2KGPr}$ and $\texttt{maxIH2KGPr}$.  Unlike $h\kappa_{\min}^*$ and $h\kappa_{\max}^*$, this last group of input arguments does not influence \cref{eq:SinusoidalInterface}'s parametrization directly; however, they act as filters that balance saddle-pattern variations in $\mathcal{D}_{sn}^{sd}$ (e.g., see \Cref{alg:CollectSinusoidalSamples}).  

Other formal parameters in \Cref{alg:GenerateSinusoidalDataSets} include the number of iterations to reinitialize \cref{eq:SinusoidalLevelSetFunction}, the amount of random noise to perturb $\phi_{sn}(\vv{x})$, and the Gaussian curvature lower bound to distinguish non-saddle stencils.  In addition, we must supply the number of amplitude steps ($\texttt{NA}$), how many rotations ($\texttt{NT}$) we need per random unit axis, and the number of steps ($\texttt{Nhk}$) we would like to have for the maximum achievable $h\kappa^*$ between $\frac{1}{2}h\kappa_{\max}^*$ and $h\kappa_{\max}^*$.  Preliminary analyses have resolved that $\texttt{NA} = 13$, $\texttt{NT} = 10$, and $\texttt{Nhk} = 7$ can produce reasonably sized data sets for any grid resolution $\eta$.

As in the case of the spherical-interface sampling routine, the first few statements in \Cref{alg:GenerateSinusoidalDataSets} begin the generative procedure by finding the mesh size and the minimum and maximum $\kappa^*$ values.  This time, however, $\kappa_{\min}^*$ and $\kappa_{\max}^*$ determine the amplitude bounds, $A_{\min}$ and $A_{\max}$, where the latter is equivalent to half the largest radius from \Cref{alg:GenerateSphericalDataSet}.  Also, here, we set the interval $[h\kappa_{\max}^{low}, h\kappa_{\max}^{up}]$, which helps populate the array $\texttt{hkMax}$ with $\texttt{Nhk}$ randomly distributed target mean curvatures for the crests and troughs.  Then, we enter a series of nested loops that employ $\texttt{NA}$ and $\texttt{NT}$ to vary the surface and the canonical frame affine-transformation parameters in a combinatorial fashion.

The outermost iterations in \Cref{alg:GenerateSinusoidalDataSets} process $\texttt{NA}$ uniformly distributed random sinusoidal amplitudes.  For each $A_{sn} \in [A_{\min}, A_{\max}]$, we allocate the buffers $\mathcal{B}_A^{ns}$ and $\mathcal{B}_A^{sd}$ to accumulate non-saddle and saddle learning tuples.  Their role is to organize data more effectively in anticipation of histogram-based balancing and keep memory consumption under control.   Likewise, we must solve for $\omega_1$ and $\omega_2$ in \cref{eq:SinusoidalInterface} for every $A_{sn}$, so that $|\max_{\vv{p}\in \mathbb{R}^2}(\kappa_{sn}(\vv{p}))|$ matches the expected $\kappa^*$ at the peaks.  In general, this problem is under-determined, but we approach it by first taking $\omega_1 = \omega_2$ with $\kappa_s$ being the corresponding maximal mean curvature.  Then, we can have $\omega_2 \neq \omega_1$ if we find the frequency $\omega_2$ that yields a maximal $\kappa_t$, such that $\kappa_{\max}^{low} \leqslant \kappa_s < \kappa_t \leqslant \kappa_{\max}^{up}$.  Intuitively, these steps vary the wave geometry for a given $A_{sn}$ by transitioning from circular to elliptical $uv$ profiles every time we pull $\kappa_s$ from $\texttt{hkMax}/h$.  \Crefrange{alg:GenerateSinusoidalDataSets.sLoop}{alg:GenerateSinusoidalDataSets.w2Definition} in \Cref{alg:GenerateSinusoidalDataSets} describe these actions more clearly with a couple of nested loops that build sequences of $(\omega_1, \omega_2)$ pairs.

The two innermost iterative blocks in \Cref{alg:GenerateSinusoidalDataSets} incorporate affine transformations to alter $\mathcal{C}_{sn}$.  To assemble these, we first construct a set $\mathcal{E}$ of orthonormal random basis vectors.  After that, we take $\texttt{NT}-1$ uniformly distributed random $\theta_{sn}$ angles in the range of $[-\pi/2, +\pi/2)$ to define multiple $R(\hat{\vv{e}}_{sn}, \theta_{sn})$ matrices for each $\hat{\vv{e}}_{sn} \in \mathcal{E}$.  As in \Cref{alg:GenerateSphericalDataSet}, we also compute a random shift $\vv{x}_{sn}$ whose components are never farther than half a (smallest) cell away from $\mathcal{C}_0$'s origin.  $\vv{x}_{sn}$ thus induces the translation matrix $T(\vv{x}_{sn})$ that completes \cref{eq:AffineTransform,eq:InverseAffineTransform}.  In practice, $\vv{x}_{sn}$, $\hat{\vv{e}}_{sn}$, and $\theta_{sn}$ are all absorbed alongside the amplitude and frequencies into the $\texttt{SinusoidalLevelSet}$ class instantiated by $\phi_{sn}(\cdot)$.  This object shall carry out the change of coordinates and level-set evaluation transparently.

Realizing $\phi_{sn}(\cdot)$ and discretizing $\Omega$ through $\mathcal{G}$ in \Cref{alg:GenerateSinusoidalDataSets} is more complex than in \Cref{alg:GenerateSphericalDataSet}.  The problem lies in the lack of a closed form to calculate $\dist{\vv{x}}{\vv{\qsn}(\cdot)}$ in \cref{eq:SinusoidalLevelSetFunction}, even if we express $\vv{x}$ in terms of the sinusoidal Monge patch coordinate system.  Certainly, we can estimate $\dist{\vv{x}}{\vv{\qsn}(\cdot)}$ via numerical optimization.  But doing so for every node in $\mathcal{G}$ is prohibitive since we must repeat the process thousands of times---for all parameter combinations.  Instead, we construct and partition $\mathcal{G}$ based on rough distance estimations to $\Gamma$.  Then, we recompute $\dist{\vv{x}}{\vv{\qsn}(\cdot)}$ via numerical optimization, but only within a small neighborhood about $\vv{\qsn}(\cdot)$.  To prepare $\mathcal{G}$, we begin by establishing a limiting sphere.  Its radius, $r_{sam}$, ought to be no larger than $1.5A_{\max}$ and at least $A_{sn}$ or twice the (longer) sinusoid period, plus some slack of $6h$.  This sampling sphere restricts the workspace with a circumscribing box $\mathcal{B}_\Omega$ that {\tt p4est} uses for spawning a macromesh with a uniform band of half-width $3h$ around $\Gamma$.  

Behind scenes, the {\tt GenerateGrid()} subroutine in \Cref{alg:GenerateSinusoidalDataSets} estimates $\dist{\vv{x}}{\vv{\qsn}(\cdot)}$ linearly by computing distances to a triangulated surface structured into a fast-querying \textit{balltree} \cite{Liu;etal;BallTreeInvestigation;2004, Omohundro;FiveBalltreeAlgorithms;1989}.  At the same time, this function memoizes such approximations to shortcut the recursive computations during the octree subdivision.  The cache constructed this way accelerates $\vv{\phi}$'s populating task, too, especially for nodes far away from $\Gamma$.  For grid points near the interface, however, we prefer to evaluate $\phi_{sn}(\cdot)$ by solving the shortest-distance problem via thrust-region-based unconstrained minimization with {\tt dlib} \cite{dlib;2009}.  With this numerical tool, we compute the perpendicular projections $\vv{x}_\mathcal{n}^\perp$, which we exploit later to retrieve target mean curvatures when collecting learning tuples.

After gathering the level-set values into $\vv{\phi}$, we continue to perturb the otherwise ``exact'' distances near $\Gamma$ by adding uniform random noise in the scaled range of $\epsilon_{rnd}h\cdot[-1, +1]$, where $\epsilon_{rnd} = \eten{1}{-4}$.  Likewise, we reinitialize the level-set field with $\nu = 10$ iterations before sampling the surface through the specialized {\tt CollectSinusoidalSamples()} method.

\Cref{alg:CollectSinusoidalSamples} describes the steps in {\tt CollectSinusoidalSamples()} for extracting learning tuples along $\vv{\qsn}(\cdot)$.  Unlike \Cref{alg:GenerateSphericalDataSet}, it tackles the sampling problem with nonuniform probability distributions to keep flat-region overrepresentation at bay.  In other words, it is clear from \cref{fig:SinusoidalExample} that the ratio between crest-/trough-stencils and near-planar stencils is disproportionally skewed in favor of the latter.  Thus, it is paramount to provide steep-mean-curvature patterns with higher selection probabilities not only to balance the $h\kappa^*$ histogram but also to save memory and disk space.  We have approached this requirement in \Cref{alg:CollectSinusoidalSamples} for non-saddle and saddle patterns separately, using $h\kappa^*$ and $h^2\kappa_G$ as their corresponding inputs to custom probability maps (defined below, in \cref{eq:Ease}).

The leading statements in \Cref{alg:CollectSinusoidalSamples} compute the normalized gradient and curvatures across $\mathcal{G}$ by operating on the input nodal level-set vector $\vv{\phi}$.  Then, the subroutine traverses the set $\mathcal{N}$ of interface nodes with complete, $h$-uniform stencils to decide which $\mathcal{n} \in \mathcal{N}$ should produce training samples to return in $\mathcal{S}^{ns}$ or $\mathcal{S}^{sd}$.  The opening criterion within the main loop (\crefrange{alg:CollectSinusoidalSamples.TraverseLoopStart}{alg:CollectSinusoidalSamples.TraverseLoopEnd}) excludes nodes too close to the walls or outside the sampling sphere of radius $r_{sam}$.  For the remaining candidate grid points, subsequent tests probe the interpolated $h^2\kappa_G$ and reference $h\kappa^*$ at $\Gamma$ to keep or discard $\mathcal{n}$ depending on whether the stencil belongs to a saddle or non-saddle region.  These last criteria are stochastic.  For example, if $\mathcal{n}$ owns a non-saddle stencil (i.e., $h^2\kappa_G \geqslant h^2\kappa_{G,\min}^{ns} = \eten{-7}{-6}$), we must use $|h\kappa^*|$ to compute $\mathcal{n}$'s likelihood of making it into $\mathcal{S}^{ns}$.  As noted above, we do not take in non-saddle samples whose $|h\kappa^*|$ values fall below $h\kappa_{\min}^* = 0.004$.  For steep interface sectors, however, we have adopted a conditioned filtering strategy assisted by easing functions.  Easing-in and -off maps are commonly employed to interpolate and control motion along curves in computer animation. \cite{ComputerAnimation08}.  Here, they furnish us with nonlinear, smooth transitions expressed by

\begin{equation}
\texttt{Ease}(t; a_e, A_e, b_e, B_e) = \left\{\begin{array}{ll}
	A_e & \textrm{if } t < a_e, \\
	A_e + \left(1 + \frac{1}{2}\left(B_e - A_e\right)\sin\left(\pi\frac{t - a_e}{b_e - a_e} - \frac{\pi}{2}\right) \right) & \textrm{if } a_e \leqslant t \leqslant b_e, \\
	B_e & \textrm{if } t > b_e, 
\end{array}\right.
\label{eq:Ease}
\end{equation}
where $a_e < b_e$ represents a finite support interval, and $A_e < B_e$ denotes the probability range.  In particular, if $|h\kappa^*|$ does not exceed $h\kappa_{\max}^{low} \equiv \frac{1}{2}h\kappa_{\max}^*$, the probability of keeping $\mathcal{n}$ results from evaluating \cref{eq:Ease} with $a_e = h\kappa_{\min}^*$, $b_e = h\kappa_{\max}^{low}$, $A_e = \texttt{minHKPr} = 0.0025$, and $B_e = \texttt{maxHKLowPr} = 0.2$.  Similarly, we can replicate this probabilistic selection process when $h\kappa_{\max}^{low} < |h\kappa^*| \leqslant h\kappa_{\max}^{up}$ by ``left-shifting'' the limits in \cref{eq:Ease} and introducing $b_e = h\kappa_{\max}^{up} \equiv h\kappa_{\max}^*$ and $B_e = \texttt{maxHKUpPr} = 0.6$ through the right end.  Intuitively, these probability arguments should offer a threefold higher chance of selecting non-saddle stencils whose $|h\kappa^*|$ values lie in the upper portion of $[h\kappa_{\min}^*, h\kappa_{\max}^*]$.  Nevertheless, further measures are still necessary if one wishes to increase non-saddle patterns' visibility around sinusoidal peaks and valleys.

The saddle-pattern selection policy in \Cref{alg:CollectSinusoidalSamples} recreates the non-saddle methodology above by replacing $h^2\kappa_G$ for $h\kappa^*$ as the definitive sampling factor.  The rationale behind this choice is the lack of a simplified relation between $|h\kappa^*|$ and $|\bar{\varepsilon}|$, as shown in \cref{fig:PrepAnalyses}.  As long as $|h^2\kappa_G| > 0$, we keep $\mathcal{n}$ with a probability that results from evaluating $|h^2\kappa_G|$ in \cref{eq:Ease}, where the support interval $[h^2\tilde{\kappa}_{G,\min}, h^2\tilde{\kappa}_{G,\max}^{low}] = [0, 0.05]$ maps to the range of $[\texttt{minIH2KGPr}, \texttt{maxIH2KGPr}] = [0.0025, 0.075]$.  Although these arguments (i.e., the chosen values for $a_e = h^2\tilde{\kappa}_{G,\min}$, $b_e = h^2\tilde{\kappa}_{G,\max}^{low}$, $A_e = \texttt{minIH2KGPr}$, and $B_e = \texttt{maxIH2KGPr}$ in \cref{eq:Ease}) should remove the excess of small-$|\bar{\varepsilon}|$ saddle data when $|h^2\kappa_G| \approx 0$ (see \cref{fig:PrepAnalyses.hk_h2kg_plane}), some post-processing is again necessary to attain a uniform distribution of categorized $h\kappa^*$ values.  Such a post-processing step takes place in \Cref{alg:GenerateSinusoidalDataSets}, as explained below.

The concluding section of \Cref{alg:CollectSinusoidalSamples} puts together six standardized data packets into $\mathcal{P}$ using \Cref{alg:GenerateStdDataPackets} for interface nodes that got through the selection filters successfully.  Then, it constructs the learning tuples $(\mathcal{q},\, h\kappa^*)$ for each $\mathcal{q} \in \mathcal{P}$ and adds them to $\mathcal{S}^{ns}$ or $\mathcal{S}^{sd}$, applying non-saddle negative-mean-curvature normalization when appropriate.  Finally, the subroutine hands over $\mathcal{S}^{ns}$ and $\mathcal{S}^{sd}$ back to \Cref{alg:GenerateSinusoidalDataSets}.  On \cref{alg:GenerateSinusoidalDataSets.CallingCollectSamples}, the latter dumps their contents into their respective buffers for the ongoing amplitude iteration before moving on to the next random affine transformation for the current geometry.

After completing each $A_{sn}$-iteration in \Cref{alg:GenerateSinusoidalDataSets}, we perform histogram-based subsampling on $\mathcal{B}_A^{ns}$ and $\mathcal{B}_A^{sd}$ to reduce their sizes and balance the target mean-curvature distribution.  This task first buckets learning pairs according to their $|h\kappa^*|$ values into a histogram $\mathcal{H}$.  This histogram has up to $100$ equally spaced bins for non-saddle patterns and up to $50$ for saddle ones.  Then, we randomly drop samples from any bucket $b_\mathcal{H} \in \mathcal{H}$ as needed until 

\begin{equation}
|b_\mathcal{H}| \leqslant \min\left(\tfrac{1}{3}m_\mathcal{H}^+,\, \tfrac{3}{2}\min_{b_\mathcal{H} \in \mathcal{H}^+}{|b_\mathcal{H}|}\right)
\label{eq:HistrogramBasedSubsampling}
\end{equation}
holds for all bins.  Here, $m_\mathcal{H}^+ = \textrm{median}(\mathcal{H}^+)$, and $\mathcal{H}^+ \doteq \{b_\mathcal{H}\, :\, b_\mathcal{H} \in \mathcal{H}\textrm{ and }|b_\mathcal{H}| > 0\}$.

Lastly, \Cref{alg:GenerateSinusoidalDataSets} reapplies histogram-based subsampling on the consolidated $\mathcal{D}_{sn}^{ns}$ and $\mathcal{D}_{sn}^{sd}$ and returns their final versions to the caller.  $\mathcal{D}_{sn}^{ns}$ and $\mathcal{D}_{sp}$ from \Cref{alg:GenerateSphericalDataSet} should then be merged into $\mathcal{D}^{ns}$ to optimize $\mathcal{F}_\kappa^{ns}(\cdot)$ by taking all or just a (uniform random) fraction of their contents.


\colorsubsubsection{Hyperbolic-paraboloidal-interface data-set construction}
\label{subsubsec:HypParaboloidalInterfaceDataSetConstruction}

Hyperbolic paraboloids are the third class of training interfaces.  They complement the methodology depicted in \Cref{alg:GenerateSinusoidalDataSets,alg:CollectSinusoidalSamples}, which has been observed to populate $\mathcal{D}^{sd}$ only with learning tuples whose $|h\kappa^*| \lessapprox 0.13$.  Hyperbolic paraboloids are easier to handle than sinusoids.  Also, they are one of the simplest saddle surfaces in Euclidean space that we can manipulate to yield any desirable mean curvatures in $\mathcal{D}_{hp}$---the set of hyperbolic-paraboloidal samples.

To assemble $\mathcal{D}_{hp}$, we extract data from level-set functions characterized by hyperbolic paraboloidal interfaces.  The model that describes these surfaces in $\mathbb{R}^3$ is the Monge patch

\begin{equation}
\vv{\qhp}(u,v) \doteq (u, v, \qhp(u,v)) \quad \textrm{with} \quad \qhp(u,v) = a_{hp}u^2 - b_{hp}v^2,
\label{eq:HypParaboloidalInterface}
\end{equation}
where $a_{hp}$ and $b_{hp}$ are positive shape parameters.  Expressing the saddle surface in this way allows us to calculate the mean and Gaussian curvatures at any $\vv{p} = [u,\, v]^T \in \mathbb{R}^2$ with

\begin{equation}
\kappa_{hp}(u,v) = \frac{(1+4b_{hp}^2v^2)a_{hp} - (1+4a_{hp}^2u^2)b_{hp}}{(1 + 4a_{hp}^2u^2 + 4b_{hp}^2v^2)^{3/2}} \quad \textrm{and} \quad
\kappa_{G,hp}(u,v) = \frac{-\,4a_{hp}b_{hp}}{(1 + 4a_{hp}^2u^2 + 4b_{hp}^2v^2)^2},
\label{eq:HypParaboloidalCurvatures}
\end{equation}
where $\kappa_{G, hp}$ is always negative, as expected.

Formally, the patch definition in \cref{eq:HypParaboloidalInterface} gives rise to a level-set function

\begin{equation}
\phi_{hp}(\vv{x}) = \left\{\begin{array}{ll}
	-\dist{\vv{x}}{\vv{\qhp}(\cdot)} & \textrm{if } z > \qhp(x,y), \\
	0                                & \textrm{if } z = \qhp(x,y) \;(\textrm{i.e., } \vv{x} \in \Gamma), \\
	+\dist{\vv{x}}{\vv{\qhp}(\cdot)} & \textrm{if } z < \qhp(x,y),
\end{array}\right.
\label{eq:HypParaboloidalLevelSetFunction}
\end{equation}
where $\vv{x} \in \mathbb{R}^3$ is an arbitrary point expressed in terms of the local coordinate system $\mathcal{C}_{hp}$.  Like $\phi_{sn}(\cdot)$ in \cref{eq:SinusoidalLevelSetFunction}, $\phi_{hp}(\cdot)$ partitions the computational domain into $\Omega^-$ above $\Gamma \equiv \vv{\qhp}(\cdot)$ and $\Omega^+$ right beneath it.

By varying $a_{hp}$ and $b_{hp}$ systematically in \cref{eq:HypParaboloidalInterface}, we can retrieve distinct mean curvatures and stencil patterns.  However, further generalization is possible if we again incorporate random affine transformations to perturb $\vv{\qhp}(\cdot)$'s canonical frame.  As in \Cref{subsubsec:SinusoidalInterfaceDataSetConstruction}, we have restrained these transforms to rotations and translations (see \cref{eq:AffineTransform,eq:InverseAffineTransform}).  These should be good enough to enrich $\mathcal{D}_{hp}$ and simplify $\dist{\vv{x}}{\vv{\qhp}(\cdot)}$'s approximation in \cref{eq:HypParaboloidalLevelSetFunction}.


\begin{algorithm}[!t]
\SetAlgoLined
\SetKwFunction{randlinspace}{RandLinspace}
\SetKwFunction{hypparaboloidallevelset}{HypParaboloidalLevelSet}
\SetKwFunction{generategrid}{GenerateGrid}
\SetKwFunction{evaluate}{Evaluate}
\SetKwFunction{addnoise}{AddNoise}
\SetKwFunction{reinitialize}{Reinitialize}
\SetKwFunction{randomunitaxis}{RandomUnitAxis}
\SetKwFunction{collecthypparaboloidalsamples}{CollectHypParaboloidalSamples}
\SetKwFunction{round}{Round}

\KwIn{$\eta$: maximum level of refinement per unit-cube octree; {\tt NR} and $r_{hp,\max}$: minimum number of $a_{hp}:b_{hp}$ ratios and their upper bound; {\tt NT}: minimum number of affine transformations per ratio; {\tt Nhk}: number of steps to vary the expected maximum $|h\kappa^*|$; $h\kappa_{\max}^*$: maximum target $|h\kappa^*|$;  {\tt minIH2KGPr}: minimum $|h^2\kappa_G|$'s easing probability; $h^2\tilde{\kappa}_{G,\max}^{low}$ and {\tt maxIH2KGPr}: lower bound for maximum $|h^2\kappa_G|$ and its easing probability; {\tt hk0Pr}: easing probability for samples with zero mean curvature; {\tt hkNoErrorPr}: easing probability for samples with zero mean-curvature error; $h\kappa_{err}^{low}$: mean-curvature error lower bound to keep a candidate grid point; $\nu$: number of iterations for level-set reinitialization; $\epsilon_{rnd}$: amount of uniform noise to perturb $\phi$; $h^2\kappa_{G,\min}^{ns}$: minimum $h^2\kappa_G$ to classify a sample as a non-saddle one.}
\KwResult{$\mathcal{D}_{hp}$: hyperbolic-paraboloidal-interface samples.}
\BlankLine

$h \leftarrow 2^{-\eta}; \quad 
h\kappa_{\max}^{low} \leftarrow \frac{1}{5}h\kappa_{\max}^*; \quad 
h\kappa_{\max}^{up} \leftarrow h\kappa_{\max}^*$\tcp*[r]{Mesh size and bounds for desired maximum $h\kappa^*$}

$\mathcal{D}_{hp} \leftarrow \varnothing$\;
$\texttt{hkMax} \leftarrow$ \randlinspace{\normalfont $h\kappa_{\max}^{low}$, $h\kappa_{\max}^{up}$, $\texttt{Nhk}$}\tcp*[r]{Array of attainable maximum $|h\kappa^*|$ values}
	
\For{\normalfont $j \leftarrow 0, ..., \texttt{Nhk}-1$}{
	$h\kappa_{\max}^{attain} \leftarrow \texttt{hkMax}[j]$\;
	$\texttt{Nr} \leftarrow$ \round{\normalfont $\texttt{NR}\cdot(1 + 2j/(\texttt{Nhk} - 1))$}\tcp*[r]{Use more ratios as $h\kappa_{\max}^{attain} \to h\kappa_{\max}^{up}$}
	$\texttt{ratios} \leftarrow$ \randlinspace{\normalfont $1$, $r_{hp,\max}$, {\tt Nr}}\;
	
	\ForEach{$a_{hp}:b_{hp}$ ratio {\normalfont $r_{hp} \in \texttt{ratios}$}}{
		select randomly whether $r_{hp} \doteq a_{hp}/b_{hp}$ or $r_{hp} \doteq b_{hp}/a_{hp}$\;\label{alg:GenerateHypParaboloidalDataSet.ABSolutionStart}
		\eIf{$r_{hp} \doteq b_{hp}/a_{hp}$}{
			$\kappa_j \leftarrow -h\kappa_{\max}^{attain} / h; \quad v_j \leftarrow 0; \quad u_j \leftarrow 0$\tcp*[r]{Steepest $\kappa_{hp}$ is negative...}
			\eIf{$r_{hp} < 3$}{
				$a_{hp} \leftarrow \frac{1}{2}|\kappa_j|(3 / r_{hp})^{3/2}; \quad
				u_j \leftarrow \sqrt{(3 / r_{hp} - 1) / (2a_{hp})^2}$\tcp*[r]{...and occurs at $(\pm u_j, 0)$}
				\lIf{$2u_j < 1.5h$}{skip this $r_{hp}$ as the critical points are too close}
			}{
				$a_{hp} \leftarrow |\kappa_j| / (r_{hp} - 1)$\tcp*[r]{...or at $(0, 0)$}
			}
			$b_{hp} \leftarrow r_{hp}a_{hp}$\;
		}{
			carry out lines \textbf{12} to \textbf{18} with $\kappa_j \leftarrow h\kappa_{\max}^{attain}$ while inverting the roles of $a_{hp}$ and $b_{hp}$ (and $u_j$ and $v_j$)\;
		}\label{alg:GenerateHypParaboloidalDataSet.ABSolutionEnd}

		$\mathcal{B}_{r} \leftarrow \varnothing$\tcp*[r]{Sample buffer for current $r_{hp}$}
		$r_{sam} \leftarrow \max(u_j, v_j) + 16h$\tcp*[r]{Sampling radius}
		$\texttt{Nt} \leftarrow$ \round{\normalfont $\texttt{NT}\cdot(1 + 2j/(\texttt{Nhk} - 1))$}\tcp*[r]{Perform more rotations as $h\kappa_{\max}^{attain} \to h\kappa_{\max}^{up}$}
		\For{\normalfont $l \leftarrow 0, ..., \texttt{Nt}-1$}{
			$\theta_{hp} \sim 2\pi\cdot \mathcal{U}(0, 1); \quad
			\hat{\vv{e}}_{hp} \leftarrow$ \randomunitaxis{}\tcp*[r]{Random affine-transforming parameters}
			$\vv{x}_{hp} \sim \vv{\mathcal{U}}(-h/2, +h/2)$\;
			let $\mathcal{B}_\Omega$ be the smallest cube enclosing the affine-transformed cylinder of radius $r_{sam}$ and maximum height $72h$\;
			\BlankLine
			
			$\phi_{hp}(\cdot) \leftarrow$ \hypparaboloidallevelset{\normalfont $a_{hp}$, $b_{hp}$, $\vv{x}_{hp}$, $\hat{\vv{e}}_{hp}$, $\theta_{hp}$}\tcp*[r]{See \cref{eq:HypParaboloidalLevelSetFunction}.  Triangulate $\Gamma$}
			$\mathcal{G} \leftarrow$ \generategrid{\normalfont $\phi_{hp}(\cdot)$, $\eta$, $\mathcal{B}_\Omega$, $3h$}\tcp*[r]{in \cref{eq:HypParaboloidalInterface} to discretize $\Omega$}
			
			$\vv{\phi} \leftarrow$ \evaluate{\normalfont $\mathcal{G}$, $\phi_{hp}(\cdot)$}; \quad 
			$\vv{\phi} \leftarrow$ \addnoise{\normalfont $\vv{\phi}$, $\epsilon_{rnd}\cdot\mathcal{U}(-h, +h)$}\tcp*[r]{Compute level-set values}
			$\vv{\phi} \leftarrow$ \reinitialize{\normalfont $\vv{\phi}$, $\nu$}\tcp*[r]{and solve \cref{eq:Reinitialization} with $\nu$ iterations}
			\BlankLine
					
			$\mathcal{S} \leftarrow$ \collecthypparaboloidalsamples{\normalfont $\mathcal{G}$, $h$, $\vv{\phi}$, $\phi_{hp}(\cdot)$, $h\kappa_{\max}^{attain}$, {\tt minIH2KGPr}, $h^2\tilde{\kappa}_{G,\max}^{low}$, {\tt maxIH2KGPr}, {\tt hk0Pr}, {\tt hkNoErrorPr}, $h\kappa_{err}^{low}$, $h^2\kappa_{G,\min}^{ns}$, $r_{sam}$}\tcp*[r]{See \Cref{alg:CollectHypParaboloidalSamples}}\label{alg:GenerateHypParaboloidalDataSet.CallingCollectSamples}
			
			$\mathcal{B}_{r} \leftarrow \mathcal{B}_{r} \cup \mathcal{S}$\;
		}
		
		use histogram-based subsampling to balance $\mathcal{B}_r$\;
		$\mathcal{D}_{hp} \leftarrow \mathcal{D}_{hp} \cup \mathcal{B}_r$\;
	}
}

use histogram-based subsampling to balance $\mathcal{D}_{hp}$\;
\Return $\mathcal{D}_{hp}$\;

\caption{$\mathcal{D}_{hp} \leftarrow$ {\tt GenerateHypParaboloidalDataSet(}$\eta$, {\tt NR}, $r_{hp,\max}$, {\tt NT}, {\tt Nhk}, $h\kappa_{\max}^*$, {\tt minIH2KGPr}, $h^2\tilde{\kappa}_{G,\max}^{low}$, {\tt maxIH2KGPr}, {\tt hk0Pr}, {\tt hkNoErrorPr}, $h\kappa_{err}^{low}$, $\nu$, $\epsilon_{rnd}$, $h^2\kappa_{G,\min}^{ns}${\tt )}: Generate a randomized data set with saddle samples from hyperbolic-paraboloidal-interface level-set functions.}
\label{alg:GenerateHypParaboloidalDataSet}
\end{algorithm}


\begin{algorithm}[!t]
\SetAlgoLined
\SetKwFunction{computenormals}{ComputeNormals}
\SetKwFunction{numcurvatures}{NumCurvatures}
\SetKwFunction{getnodesnexttogamma}{GetNodesNextToGamma}
\SetKwFunction{ease}{Ease}
\SetKwFunction{collectfeatures}{CollectFeatures}
\SetKwFunction{generatestddatapackets}{GenerateStdDataPackets}

\KwIn{$\mathcal{G}$: grid; $h$: mesh size; $\vv{\phi}$: nodal level-set values; $\phi_{hp}(\cdot)$: hyperbolic-paraboloidal-interface level-set function object; $h\kappa_{\max}^{attain}$: maximum attainable $|h\kappa^*|$; {\tt minIH2KGPr}: minimum $|h^2\kappa_G|$'s easing probability; $h^2\tilde{\kappa}_{G,\max}^{low}$ and {\tt maxIH2KGPr}: lower bound for maximum $|h^2\kappa_G|$ and its easing probability; {\tt hk0Pr}: easing probability for samples with zero mean curvature; {\tt hkNoErrorPr}: easing probability for samples with zero mean-curvature error; $h\kappa_{err}^{low}$: mean-curvature error lower bound to keep a sample; $h^2\kappa_{G,\min}^{ns}$: minimum $h^2\kappa_G$ to classify a sample as a non-saddle one; $r_{sam}$: sampling radius.}
\KwResult{$\mathcal{S}$: saddle samples.}
\BlankLine

$\hat{N} \leftarrow$ \computenormals{$\mathcal{G}$, $\vv{\phi}$}$; \quad
(\vv{\kappa},\, \vv{\kappa}_G) \leftarrow$ \numcurvatures{$\mathcal{G}$, $\vv{\phi}$, $\hat{N}$}\tcp*[r]{See \cref{eq:NormalAndMeanCurvature,eq:MeanCurvature.3d,eq:GaussianCurvature.Ext}}
\BlankLine

$\mathcal{S} \leftarrow \varnothing$\;			
$\mathcal{N} \leftarrow$ \getnodesnexttogamma{$\mathcal{G}$, $\vv{\phi}$}\;
let $\vv{x}_{hp}$ be $\phi_{hp}(\cdot)$'s offset in terms of the world coordinate system\;
\ForEach{node $\mathcal{n} \in \mathcal{N}$ with a complete, $h$-uniform stencil}{
	\BlankLine
	
	\lIf{$\|\mathcal{n}.\vv{x} - \vv{x}_{hp}\| > r_{sam}$ {\normalfont\textbf{or}} $\mathcal{n}.\vv{x}$ is less than $4h$ away from $\mathcal{G}$'s walls}{skip node $\mathcal{n}$}
	\BlankLine
	
	compute the target $h\kappa^*$ at the nearest point $\vv{x}_\mathcal{n}^\perp$ to $\mathcal{n}.\vv{x}$ on $\Gamma$\tcp*[r]{See \cref{eq:SinusoidalCurvatures}}
	$\vv{x}_\mathcal{n}^\Gamma \leftarrow \mathcal{n}.\vv{x} - \vv{\phi}[\mathcal{n}]\hat{N}[\mathcal{n}]$\tcp*[r]{See \cref{eq:NormalProjection}}
	$h\kappa \leftarrow h \cdot$\interpolate{$\mathcal{G}$, $\vv{\kappa}$, $\vv{x}_\mathcal{n}^\Gamma$}$; \quad
	h^2\kappa_G \leftarrow h^2 \cdot$\interpolate{$\mathcal{G}$, $\vv{\kappa}_G$, $\vv{x}_\mathcal{n}^\Gamma$}\tcp*[r]{See Algorithm \href{https://www.sciencedirect.com/science/article/pii/S002199911630242X\#fg0050}{2} in \cite{Mirzadeh;etal:16:Parallel-level-set}}
	\BlankLine
	
	\tcp{Apply probabilistic filtering (see \cref{eq:Ease}) for saddle regions}
	\eIf{$h^2\kappa_G \geqslant h^2\kappa_{G,\min}^{ns}$}{\label{alg:CollectHypParaboloidalSamples.FiltersStart}
		skip node $\mathcal{n}$\tcp*[r]{Skip numerical non-saddle regions}
	}{
		\lIf{$\mathcal{U}(0, 1) >$ \ease{\normalfont $|h^2\kappa_G|$; $|h^2\kappa_{G,\min}^{ns}|$, {\tt minIH2KGPr}, $h^2\tilde{\kappa}_{G,\max}^{low}$, {\tt maxIH2KGPr}}}{skip node $\mathcal{n}$}
		
		\lIf{$\mathcal{U}(0, 1) >$ \ease{\normalfont $|h\kappa|$; $0$, {\tt hk0Pr}, $\frac{1}{2}h\kappa_{\max}^{attain}$, $1$}}{skip node $\mathcal{n}$}
		
		\lIf{$\mathcal{U}(0, 1) >$ \ease{\normalfont $|h\kappa^* - h\kappa|$; $0$, {\tt hkNoErrorPr}, $h\kappa_{err}^{low}$, $1$}}{skip node $\mathcal{n}$}
	}\label{alg:CollectHypParaboloidalSamples.FiltersEnd}
	\BlankLine
	
	$\mathcal{p} \leftarrow$ \collectfeatures{\normalfont $\mathcal{n}.\texttt{stencil}$, $\vv{\phi}$, $\hat{N}$}\tcp*[r]{Populate $\mathcal{p}$ (see \cref{eq:DataPacket})}
	$\mathcal{p}.h\kappa \leftarrow h\kappa; \quad \mathcal{p}.h^2\kappa_G \leftarrow h^2\kappa_G$\;
	\BlankLine
	
	$\mathcal{P} \leftarrow$ \generatestddatapackets{$\mathcal{p}$, $\mathcal{n}.\vv{x}$}\tcp*[r]{See \Cref{alg:GenerateStdDataPackets}}
	\ForEach{standard-formed data packet $q \in \mathcal{P}$}{
		let $\vv{\xi}$ be the learning tuple $\left(\mathcal{q},\, h\kappa^*\right)$ with inputs $\mathcal{q}$ and expected output $h\kappa^*$\;
		$\mathcal{S} \leftarrow \mathcal{S} \cup \{\vv{\xi}\}$\;
	}
}
\BlankLine

\Return $\mathcal{S}$\;

\caption{$\mathcal{S} \leftarrow$ {\tt CollectHypParaboloidalSamples(}$\mathcal{G}$, $h$, $\vv{\phi}$, $\phi_{hp}(\cdot)$, $h\kappa_{\max}^{attain}$, {\tt minIH2KGPr}, $h^2\tilde{\kappa}_{G,\max}^{low}$, {\tt maxIH2KGPr}, {\tt hk0Pr}, {\tt hkNoErrorPr}, $h\kappa_{err}^{low}$, $h^2\kappa_{G,\min}^{ns}$, $r_{sam}${\tt )}: Collect saddle samples along a hyperbolic paraboloidal surface.}
\label{alg:CollectHypParaboloidalSamples}
\end{algorithm}

\Cref{alg:GenerateHypParaboloidalDataSet} provides the general combinatorial procedure for constructing $\mathcal{D}_{hp}$.  The main idea is to set a target, maximal $\kappa_{hp}$.  Then, we should solve for its hyperbolic paraboloidal parameters drawn from an assorted pool of $a_{hp}:b_{hp}$ ratios and apply many random affine transformations to the geometry defined by every $(a_{hp},\, b_{hp})$ pair.  For each such instance of \cref{eq:HypParaboloidalLevelSetFunction}, we would examine the interface nodes and collect learning tuples using a probabilistic strategy to balance the $h\kappa^*$ distribution.

The formal parameters in \Cref{alg:GenerateHypParaboloidalDataSet} include the mesh resolution $\eta$ and the number of steps $\texttt{Nhk}$ to space out the maximally achievable $h\kappa^*$ between $h\kappa_{\max}^{low} \doteq \frac{1}{5}h\kappa_{\max}^*$ and $h\kappa_{\max}^* = 2/3$.  Incidentally, we have chosen $h\kappa_{\max}^{low} = 2/15 \approx 0.133333$ so that the overlap with the sinusoidal saddle data collected with \Cref{alg:GenerateSinusoidalDataSets,alg:CollectSinusoidalSamples} is minimal.  The routine also expects $\texttt{NR}$---the initial number of $a_{hp}:b_{hp}$ ratios in the range of $[1, r_{hp,\max}]$, where $r_{hp,\max} > 1$.  In our preliminary experiments, we have found that $\texttt{Nhk} = 150$ and $\texttt{NR} = 12$ work well with $\texttt{NT} = 30$, which denotes the least number of affine transformations to alter $\mathcal{C}_{hp}$. 

In addition, \Cref{alg:GenerateHypParaboloidalDataSet} requires a few probabilistic bounds and their support values for easing-based subsampling functions.  Among these, we need $\texttt{minIH2KGPr}$ and $\texttt{maxIH2KGPr}$, where $h^2\tilde{\kappa}_{G,\max}^{low}$ maps to the latter and $|h^2\kappa_{G,\min}^{ns}|$ to the former.  These Gaussian-curvature parameters are supplemented by probabilities associated with $h\kappa^*$, like $\texttt{hk0Pr}$ and $\texttt{hkNoErrorPr}$.  While $\texttt{hk0Pr}$ helps reduce the skewness in the mean-curvature distribution near zero, $\texttt{hkNoErrorPr}$ aids in minimizing the overrepresentation of small-error samples whose $|\bar{\varepsilon}| < h\kappa_{err}^{low}$.  Later, we will focus our discussion on these formal parameters in \Cref{alg:CollectHypParaboloidalSamples}.

The last input arguments in \Cref{alg:GenerateHypParaboloidalDataSet} are recurrent across our data-generative routines.  They are the number of redistancing steps, the amount of uniform random noise to perturb $\phi_{hp}(\cdot)$, and the $h$-normalized Gaussian-curvature decision boundary to identify non-saddle patterns.  For consistency and the benefit of our MLPs, we have set them again to $10$, $\eten{1}{-4}$, and $\eten{-7}{-6}$, respectively.

The outermost loop in \Cref{alg:GenerateHypParaboloidalDataSet} iterates over $\texttt{Nhk}$ uniformly distributed random $h\kappa^*$ values between $h\kappa_{\max}^{low}$ and $h\kappa_{\max}^{up}$.  Each represents a target non-dimensionalized maximal mean curvature $h\kappa_{\max}^{attain}$ that we can use to characterize \cref{eq:HypParaboloidalInterface}.  In other words, our goal is to solve for $a_{hp}$ and $b_{hp}$ such that $\kappa_{\max}^{attain} = \max_{u,v}|\kappa_{hp}(u,v)|$ (see  \cref{eq:HypParaboloidalCurvatures}).  This optimization problem is under-determined, but we can approach it by constraining $\vv{\qhp}(\cdot)$ so that $b_{hp} = r_{hp}a_{hp}$ or $a_{hp} = r_{hp}b_{hp}$, where $r_{hp} \geqslant 1$.  \Cref{alg:GenerateHypParaboloidalDataSet} alternates randomly between these two choices by drawing $r_{hp}$ from an $\texttt{Nr}$-element array whose size increases proportionally with $h\kappa_{\max}^{attain}$.  The purpose of this dynamic-size strategy is to make up for the growing difficulty of collecting high-curvature samples as $h\kappa_{\max}^{attain} \to h\kappa_{\max}^{up}$.

When $r_{hp} = 1$, $\kappa_{hp}(\cdot)$ has a saddle point at $\mathcal{C}_{hp}$'s origin and four global-extremum, nonzero critical points sitting on the $u$- or $v$-axis.  However, when $r_{hp} > 1$, the global mean-curvature extrema may be negative and appear symmetrically on the $u$-axis (if $b_{hp} > a_{hp}$), or they may be positive and lie symmetrically on the $v$-axis (if $a_{hp} > b_{hp}$).  In any of the latter cases, a bifurcation occurs when $r_{hp} = 3$, where the two global-extremum critical points collapse onto $(0, 0)$.  \Cref{alg:GenerateHypParaboloidalDataSet} handles all these conditions in \crefrange{alg:GenerateHypParaboloidalDataSet.ABSolutionStart}{alg:GenerateHypParaboloidalDataSet.ABSolutionEnd} when finding $a_{hp}$ and $b_{hp}$ for the desired $\kappa_{\max}^{attain}$.  Likewise, this block of statements ensures that such critical points are at least $1.5h$ apart from each other when $r_{hp} < 3$.  This restriction provides the sampling procedure with the ability to leave out degenerate scenarios where $h\kappa_{\max}^{attain}$ is unreachable due to $\mathcal{G}$'s resolution.  Part of the dynamic-sizing strategy discussed above for the $\texttt{ratios}$ array is also motivated by this constraint, alongside the increasing number of skipped $a_{hp}:b_{hp}$ ratios as $h\kappa_{\max}^{attain} \to h\kappa_{\max}^{up}$.

After defining $a_{hp}$ and $b_{hp}$, we allocate the buffer $\mathcal{B}_r$ to accumulate learning samples across variations of the same geometry.  Its purpose is to aid in class balancing via histogram-based subsampling at the intermediate stages of $\mathcal{D}_{hp}$'s construction.  Here too, a couple of other preparatory instructions determine the sampling radius $r_{sam} \geqslant 16h$ and the varying number $\texttt{Nt}$ of rotations (and translations).  $\texttt{Nt}$'s definition is identical to $\texttt{Nr}$'s, except that its lower bound is $\texttt{NT}$.  In line with $\texttt{Nr}$, $\texttt{Nt}$'s dynamic size works against the difficulty of garnering steep-mean-curvature data as $h\kappa_{\max}^{attain} \to h\kappa_{\max}^{up}$.

The innermost loop in \Cref{alg:GenerateHypParaboloidalDataSet} instantiates $\phi_{hp}(\cdot)$ $\texttt{Nt}$ times for a prescribed $(a_{hp}, b_{hp})$ pair.  Its structure is comparable to what we have described in \Cref{alg:GenerateSinusoidalDataSets}.  In each iteration, we first draw an angle $\theta_{hp}$ and a unit axis $\hat{\vv{e}}_{hp}$ from uniform random distributions to build a rotation matrix $R(\hat{\vv{e}}_{hp})$.  The {\tt RandomUnitAxis()} method, in particular, recreates \crefrange{alg:GenerateSphericalDataSet.PointOnSphereStart}{alg:GenerateSphericalDataSet.PointOnSphereEnd} in \Cref{alg:GenerateSphericalDataSet} to yield $\hat{\vv{e}}_{hp}$ from the unit sphere.  Afterward, we produce a random shift $\vv{x}_{hp}$ around $\mathcal{C}_0$'s origin to complete the definition of the affine transforms in \cref{eq:AffineTransform,eq:InverseAffineTransform}.

Next, we set up $\mathcal{B}_\Omega$ by finding the smallest cube enclosing the affine-transformed cylinder of radius $r_{sam}$, whose base is at $\max\left(-32h,\, \min\left(\vv{\qhp}(r_{sam}, 0),\, \vv{\qhp}(0, r_{sam})\right)\right) - 4h$ and top is at $\min\left(+32h,\, \max\left(\vv{\qhp}(r_{sam}, 0),\, \vv{\qhp}(0, r_{sam})\right)\right) + 4h$.  Our goal is to define $\Omega$ about $\vv{\qhp}(\cdot)$'s centroid with minimal but sufficient space for capturing $h\kappa_{\max}^{attain}$ at $(\pm u_j,0)$ or $(0, \pm v_j)$.  After that, $\mathcal{B}_\Omega$ and the corresponding {\tt HypParaboloidalLevelSet} object $\phi_{hp}(\cdot)$ are used to spawn $\mathcal{G}$ with a uniform band of half-width $3h$ around $\Gamma$.  As in \Cref{alg:GenerateSinusoidalDataSets}, we have also triangulated and organized the surface into a balltree for fast $\dist{\vv{x}}{\vv{\qhp}(\cdot)}$ computations during mesh refinement and partitioning.  Similarly, memoization has allowed us to speed up level-set evaluation when far away from the interface.  Near $\Gamma$, however, we have resorted to {\tt dlib}'s minimization tools to determine $\vv{x}_\mathcal{n}^\perp$ and calculate ``exact'' signed distances wherever the linear approximation to $\phi_{hp}(\vv{x}_\mathcal{n})$ does not exceed $3h\sqrt{3}$.

Upon evaluating \cref{eq:HypParaboloidalLevelSetFunction} into $\vv{\phi}$, we perform the typical random level-set perturbation followed by reinitialization.  Then, we trigger sample collection along $\Gamma$ by calling the {\tt CollectHypParaboloidalSamples()} subroutine shown in \Cref{alg:CollectHypParaboloidalSamples}.

The high-level logic of \Cref{alg:CollectHypParaboloidalSamples,alg:CollectSinusoidalSamples} is very much alike.  The only difference resides in how \Cref{alg:CollectHypParaboloidalSamples} filters saddle-pattern interface nodes lying inside the sampling sphere and not too close to the walls.  Here, we abridge the discussion by centering our attention on the conditional statement in \crefrange{alg:CollectHypParaboloidalSamples.FiltersStart}{alg:CollectHypParaboloidalSamples.FiltersEnd}.  Furthermore, we assume $\mathcal{n} \in \mathcal{N}$ is a candidate node with a regular, $h$-uniform stencil satisfying $h^2\kappa_G < h\kappa_{G,\min}^{ns}$.  To add $\mathcal{n}$'s learning tuples into the returning set $\mathcal{S}$, we verify three probabilistic criteria on $\mathcal{n}$'s curvature attributes.  First, we determine whether to keep or discard $\mathcal{n}$ by calculating $|h^2\kappa_G|$'s probability from an easing function (see \cref{eq:Ease}), where the support $[|h^2\kappa_{G,\min}|, \texttt{minIH2KGPr}]$ maps to $[\texttt{minIH2KGPr}, \texttt{maxIH2KGPr}] = [0, 1]$, and $\texttt{minIH2KGPr} = 0.01$.  This criterion ensures that higher-Gaussian-curvature samples are more likely to make it into $\mathcal{D}_{hp}$ than those with $\kappa_{G} \approx 0$.  If $\mathcal{n}$ succeeds in this test, we continue to check $|h\kappa|$.  In this case, the probability of keeping $\mathcal{n}$ depends on a distribution where $|h\kappa| = 0$ maps to $\texttt{hk0Pr} = 0.0025$ and $|h\kappa| = \frac{1}{2}h\kappa_{\max}^{attain}$ maps to $1$.  Such a distribution starts with a narrow support interval during the first iterations of the outermost loop in \Cref{alg:GenerateHypParaboloidalDataSet}.  Then, it widens as $|h\kappa_{\max}^{attain}| \to h\kappa_{\max}^{up}$, gradually reducing the selection probabilities around low-mean-curvature regions.

\begin{figure}[!t]
	\centering
	\includegraphics[width=7.25cm]{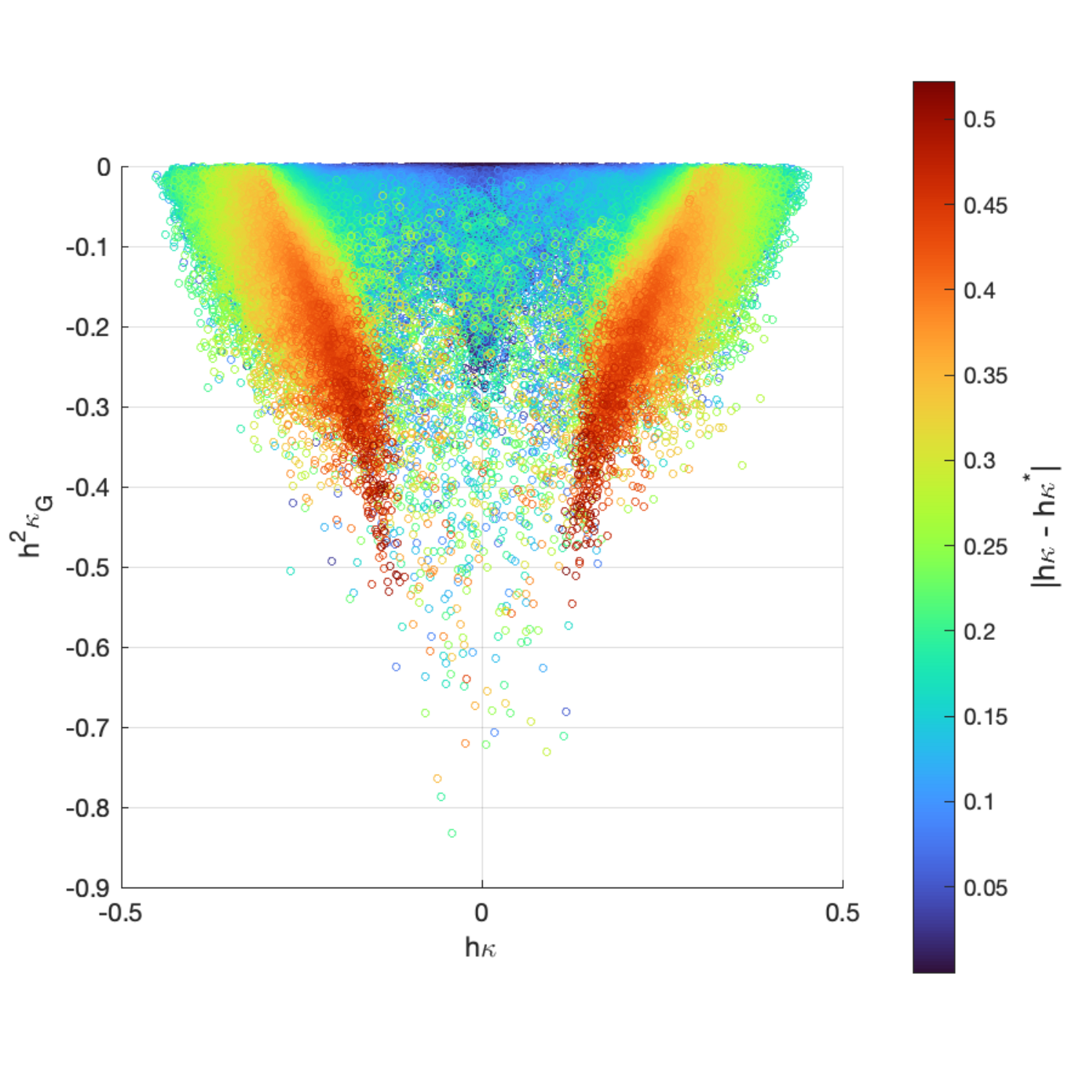}
	\caption{$|\bar{\varepsilon}|$ and its relation with dimensionless mean and Gaussian curvatures from saddle data collected with \Cref{alg:GenerateHypParaboloidalDataSet,alg:CollectHypParaboloidalSamples}.  (Color online.)}
	\label{fig:HypParaboloidalHeatMap}
\end{figure}

The last sampling criterion in \Cref{alg:CollectHypParaboloidalSamples} depends on $\mathcal{n}$'s mean-curvature error.  Its purpose is to increase the visibility of the saddle-pattern interface nodes where $|h\kappa^* - h\kappa| \geqslant h\kappa_{err}^{low}$.  To this end, we again compute $\mathcal{n}$'s selection probability with an easing function that maps the support interval $[0, h\kappa_{err}^{low}]$ to the range of $[\texttt{hkNoErrorPr}, 1]$.  In our implementation, $\texttt{hkNoErrorPr} = 0.005$ has worked well alongside $h\kappa_{err}^{low} = 0.1$ to mitigate low-error-sample overrepresentation.  In fact, we have found this error-based strategy helpful for trimming saddle data, given the complex relationship between $|\bar{\varepsilon}|$, $h\kappa$, and $h^2\kappa_G$, as shown in \cref{fig:HypParaboloidalHeatMap}.

All interface nodes succeeding through the internal filters in \Cref{alg:CollectHypParaboloidalSamples} contribute to $\mathcal{S}$ with six standard-formed-input learning tuples.  Eventually, we return this set to \Cref{alg:GenerateHypParaboloidalDataSet} on \cref{alg:GenerateHypParaboloidalDataSet.CallingCollectSamples}, where the orchestrating, innermost loop transfers its contents into $\mathcal{B}_r$.  The $h\kappa^*$ histogram in this auxiliary buffer grows steadily in the negative and positive directions, but is not always guaranteed to be well-balanced.  For this reason, we have interleaved histogram-based subsampling with every $r_{hp}$ by bucketing $\mathcal{B}_r$'s samples into $50$ to $100$ uniform slots (see \cref{eq:HistrogramBasedSubsampling}).  Thanks to this intermediate bucketing and balancing, \Cref{alg:GenerateHypParaboloidalDataSet} can populate $\mathcal{D}_{hp}$ efficiently.  Had we not done so, we could have run into memory-consumption issues, given the combinatorial nature of our routine.

As in \Cref{alg:GenerateSinusoidalDataSets}, the concluding statement in \Cref{alg:GenerateHypParaboloidalDataSet} levels up the final $h\kappa^*$ distribution in $\mathcal{D}_{hp}$.  Unlike $\mathcal{D}^{ns}$, however, we cannot build $\mathcal{D}^{sd}$ by just joining $\mathcal{D}_{hp}$ with $\mathcal{D}_{sn}^{sd}$.  The reason is that both data sets' $h\kappa^*$ histograms have varying-length supports and may exhibit disproportionate frequencies.  To address this problem, we have resorted to bucketing and balancing once more, using $100$ equally spaced intervals to categorize $|h\kappa^*|$.  Then, with $\mathcal{D}^{ns}$ and $\mathcal{D}^{sd}$ at hand, we can proceed to the next training step and optimize the error-correcting models needed in \Cref{alg:MLCurvature}.


\colorsubsection{Technical aspects}
\label{subsec:TechnicalAspects}

Our prior experience in \cite{Larios;Gibou;HybridCurvature;2021,Larios;Gibou;KECNet2D;2022} has provided a starting point for tackling the technical corners of the three-dimensional mean-curvature problem.  Most ideas, such as dimensionality reduction, z-scoring, regularization, $K$-fold cross-validation splitting, and even network design, have been readily adapted to our workflow to favor generalization.  Similarly, we have used performant, parallel libraries to deploy our neural models and \Cref{alg:MLCurvature} into the level-set framework.

First, we should point out that we have instrumented the algorithms in this report within our C++ implementation of the parallel level-set schemes proposed by Mirzadeh \etal \cite{Mirzadeh;etal:16:Parallel-level-set}.  At its core, our level-set codebase relies on {\tt p4est} \cite{Burstedde;Wilcox;Ghattas:11:p4est:-Scalable-Algo}, {\tt MPI} \cite{MPI14}, and {\tt PETSc} \cite{Balay;Brown;Buschelman;etal:12:PETSc-Web-page} to distribute numerical computations in a heterogenous system; however, we have augmented this codebase with {\tt OpenMP} \cite{OpenMP;2008} to improve performance when solving shortest-distance problems with {\tt dlib} \cite{dlib;2009}.  

Unlike the two-dimensional case, parallel processing is indispensable for making data-set construction viable in $\mathbb{R}^3$.  Consider \Cref{alg:GenerateSinusoidalDataSets}, for example.  Even with multiprocessing enabled, assembling $\mathcal{D}_{sn}^{ns}$ and $\mathcal{D}_{sn}^{sd}$ can take up to four days for one mesh resolution on a 3.7GHz 12-logical-core workstation with 64GB RAM.  Fortunately, the combinatorial design of \Crefrange{alg:GenerateSphericalDataSet}{alg:CollectHypParaboloidalSamples} has allowed us to split ranges of iterations among multiple computing nodes (e.g., on Stampede2 \cite{Stampede2;2022}).  By exploiting this feature, we have populated $\mathcal{D}^{ns}$ and $\mathcal{D}^{sd}$ and left them ready for training in less than five days.  In this context, it is also worth noting that sampling could have grown intractable for any strategy had we approached the problem with uniform grids.

In this research, we have optimized $\mathcal{F}_\kappa^{ns}(\cdot)$ and $\mathcal{F}_\kappa^{sd}(\cdot)$ with TensorFlow \cite{Tensorflow15} and Keras \cite{Keras15} in Python.  Moreover, we have adopted a conventional data splitting strategy, reserving $70\%$, $15\%$, and $15\%$ of the learning tuples in $\mathcal{D}^{ns}$ and $\mathcal{D}^{sd}$ for their training, testing, and validation subsets.  The validation subset is critical to prevent overfitting and guide model selection as we explore the hyperparameter space \cite{A18, Mehta19}.  

To reduce the likelihood of biased $h\kappa^*$ distributions in the training, testing, and validation subsets, we have employed Scikit-learn's {\tt split()} method from the {\tt StratifiedKFold} class \cite{scikit-learn11}.  $K$-fold cross-validation is a staple feature in machine learning for assessing the performance of regression and classification models.  It involves splitting a parent data set into $K$ subsets called \textit{folds} and fitting and evaluating a model $K$ times, choosing a different fold for testing in every iteration while training on the other $(K-1)$ segments \cite{A18, Hands-onMLwithScikit-LearnKerasAndTF19}.  A full-fledged $K$-fold cross-validation session is unaffordable in deep learning when the data sets are oversized.  However, the {\tt StratifiedKFold}'s {\tt split()} subroutine offers a convenient way of assembling randomized subsets with the same label distribution as $\mathcal{D}^{ns}$ or $\mathcal{D}^{sd}$.  To incorporate this method into our pipeline, we initially label all entries in $\mathcal{D}^{ns}$ and $\mathcal{D}^{sd}$ by bucketing their $|h\kappa^*|$ values into one hundred categories with Panda's {\tt cut()} function \cite{Pandas2010}.  Afterward, we partition the parent data sets into $K = 20$ folds by calling {\tt split()}.  Then, we group them accordingly to achieve the desired subset proportions.

Another essential component in our mean-curvature solver is the preprocessing module portrayed in \cref{fig:ECNet}.  Its role is to transform \cref{eq:DataPacket} into an amenable vector representation for our MLPs and rid the data of noisy, often harmful, signals.  \Cref{alg:Preprocess} lists the preprocessing operations performed on an incoming data packet $\mathcal{p}$ before sending it over to $\mathcal{F}_\kappa^{ns}(\cdot)$ or $\mathcal{F}_\kappa^{sd}(\cdot)$.  These transformations employ the statistics collected from the training subset during the fitting stage, and we assume they are available to the current routine through the $\texttt{Q}$ object.  First, \Cref{alg:Preprocess} $h$-normalizes the level-set values in $\mathcal{p}$.  Not only does this normalization release $\mathcal{p}$ from some particular mesh size, but it also minimizes the possibility of losing precision as $h \to 0$.  Then, we compute its \textit{z-scores} after constructing $\mathcal{p}$'s feature vector $\vv{f} \in \mathbb{R}^{110}$.  Standardization, or z-scoring, transforms $\vv{f}$ into $\vv{z}$ by centering it around the feature-wise mean and scaling it by the corresponding standard deviation \cite{Parker;CS170A;2016}.  During training, we have easily retrieved these first and second statistical moments using a {\tt StandardScaler} object from Scikit-learn.

The last section in \Cref{alg:Preprocess} performs dimensionality reduction and \textit{whitening} on $\vv{z} \in \mathbb{R}^{110}$ via \textit{principal component analysis} (PCA).  In contrast to plain z-scoring \cite{LALariosFGibou;LSCurvatureML;2021}, our preliminary research \cite{Larios;Gibou;HybridCurvature;2021, Larios;Gibou;ECNetSemiLagrangian;2021, Larios;Gibou;KECNet2D;2022} and the work by LeCun \etal \cite{LeCun;EfficientBackProp;98} have shown that centering, scaling, and uncorrelating data with these techniques can accelerate convergence and increase the predictive accuracy of the downstream estimators.  On one hand, PCA allows us to approximate a matrix by a sum of a few rank-one matrices \cite{Moler;NumComputingMatlab;2013}.  Also, it is the method of choice to uncover the intrinsic axes with the maximum spread of the data \cite{Parker;CS170A;2016}.  More succinctly, suppose $D \in \mathbb{R}^{n\times 110}$ is the $n$-sample matrix version of one of our training subsets.  Furthermore, let $C \doteq \frac{1}{n-1}(D-M)^T (D-M)$ be its 110-by-110 positive semi-definite covariance matrix, where $M \in \mathbb{R}^{n\times 110}$ contains $D$'s column-wise mean row vector $\vv{\mu}^T$ stacked $n$ times.  Then, we can find $D$'s \textit{principal components} by factoring $C = USV^T$, where the columns in $U = V = [\hat{\vv{v}}_0,\, \hat{\vv{v}}_1,\, \dots,\, \hat{\vv{v}}_{k-1},\, \hat{\vv{v}}_k,\, \dots,\, \hat{\vv{v}}_{109}]$ are $C$'s singular vectors, and $S = \texttt{diag(}[\sigma_0,\, \sigma_1,\, \dots,\, \sigma_{k-1},\, \sigma_k,\, \dots,\, \sigma_{109}]\texttt{)}$ holds the singular values.  Incidentally, $(D-M)\hat{\vv{v}}_0$ has the largest possible variance $\sigma_0$, $(D-M)\hat{\vv{v}}_1$ has the second largest variance $\sigma_1$, and so on.  Thus, PCA-transforming $\vv{z}$ into $\vv{r} \in \mathbb{R}^{m_\iota}$ in \Cref{alg:Preprocess} entails projecting $\vv{z}$ onto a $k$-dimensional space that closely approximates $D$'s underlying structure.  As seen in \cref{fig:ECNet}, $k = m_\iota < 110$ is a hyperparameter that we must adjust individually for $\mathcal{F}_\kappa^{ns}(\cdot)$ and $\mathcal{F}_\kappa^{sd}(\cdot)$.  Likewise, we should remark that $C$ is technically $D$'s correlation matrix in our case, since the columns in $D$ were standardized in the preceding step.  By using such a correlation matrix in our workflow, we have prevented PCA from picking up spurious correlations determined by the scale of the data \cite{Parker;CS170A;2016}.  

The whitening operation in \Cref{alg:Preprocess} is a simple normalization of the projected data by the resulting \textit{explained standard deviations} (i.e., $\sqrt{\sigma_i}$, for $i = 0, 1, \dots, m_\iota - 1$) \cite{scikit-learn11}.  Its role is to provide equal importance to uncorrelated concepts so that our MLPs can decide which of them to emphasize during the fitting process \cite{A18}.  In this work, we have exploited Scikit-learn's {\tt PCA} class to extract principal components and explained variances from the training subsets.  Similarly, we have exported the {\tt PCA} and {\tt StandardScaler} transformers into {\tt pickle}- and {\tt JSON}-formatted artifacts to facilitate offline and online inference in Python and C++.


\begin{algorithm}[!t]
\SetAlgoLined
\SetKwFunction{getstats}{GetStats}

\KwIn{$\mathcal{p}$: data packet to preprocess; $h$: mesh size.}
\KwResult{$\vv{r}$: vector form of the transformed data packet.}
\BlankLine

\tcp{$h$-normalize input level-set values}
Divide the 27 $\mathcal{p}.\phi_{ijk}$ features by $h$, where $i,j,k \in \{m, 0, p\}$\;
\BlankLine

\tcp{Computing z-scores}
{\tt Q} $\leftarrow$ \getstats{}\tcp*[r]{Retrieve training stats}
let $\vv{f} \in \mathbb{R}^{110}$ be $\mathcal{p}$'s feature vector form\;
let $\vv{\mu}_{STD}$ and $\vv{\sigma}_{STD}$ be the feature-wise mean and standard deviation vectors in {\tt Q.StandardScaler}\;
$\vv{z} \leftarrow (\vv{f} - \vv{\mu}_{STD}) \oslash \vv{\sigma}_{STD}$\tcp*[r]{$\oslash$ denotes a point-wise division}
\BlankLine

\tcp{Dimensionality reduction and whitening using $\texttt{Q}.m_\iota < 110$ components}
let $V_{PCA}$ and $\vv{\sigma}_{PCA}$ be the principal components matrix and singular values vector in {\tt Q.PCA}\;
$\vv{r} \leftarrow (V_{PCA}^T\vv{z}) \oslash \sqrt{\vv{\sigma}_{PCA}}$\;
\BlankLine

\Return $\vv{r}$\;

\caption{$\vv{r} \leftarrow$ {\tt Preprocess(}$\mathcal{p}$, $h${\tt )}: Preprocess a standard-formed data packet $\mathcal{p}$.}
\label{alg:Preprocess}
\end{algorithm}

The Keras module in TensorFlow 2.0 has provided us with a high-level functional API\footnote{Refer to TensorFlow's Python \href{https://www.tensorflow.org/versions/r2.0/api_docs/python/tf}{\textbf{A}pplication \textbf{P}rogramming \textbf{I}nterface} \cite{TensorFlowAPI2;2022}.} for assembling and monitoring the fitting process of our neural models prototyped in \cref{fig:ECNet}.  In their two-part input layers, $\mathcal{F}_\kappa^{ns}(\cdot)$ and $\mathcal{F}_\kappa^{sd}$ have $m_\iota + 1$ linearly activated neurons that transfer preprocessed feature vectors and the $h\kappa$ signal toward the hidden and output layers.  The first four hidden layers contain $N_{h}^i$ fully connected ReLU units, where $i = 1, \dots, 4$, and $\textrm{ReLU}(x) = \max{(0,x)}$.  ReLU functions are almost linear and share many of the properties that make linear models easy to optimize with (stochastic) gradient descent \cite{DeepLearning;Goodfellow-et-al;2016}.  Also, they are fast to evaluate, and their non-saturating feature counteracts the vanishing-gradient problem when training deep supervised neural networks \cite{DeepSparseRectifierNN;Glorot-et-al;2011}.  

After the nonlinear hidden layers in \cref{fig:ECNet}, a single linear neuron calculates the error-correction term $\bar{\varepsilon}$ and sends it over to the output unit that produces $h\kappa_\mathcal{F}$.  We can express a forward pass on either $\mathcal{F}_\kappa^{ns}([\vv{r}, h\kappa])$ or $\mathcal{F}_\kappa^{sd}([\vv{r}, h\kappa])$ more formally by the recurrence

\begin{equation}
\left\{\begin{array}{ll}
	\vv{h}_1 = W_1^T\vv{r} + \vv{b}_1                          & \textrm{from the input to the first hidden layer}, \\
	\vv{h}_{i+1} = \textrm{ReLU}(W_{i+1}^T\vv{h}_i + \vv{b}_{i+1}),\; 1 \leqslant i \leqslant 3 & \textrm{from the }i\textrm{th to the }(i+1)\textrm{th hidden layer}, \\
	h\kappa_\mathcal{F} = (W_5^T\vv{h}_4 + \vv{b}_5) + h\kappa & \textrm{from the fourth hidden layer to the output}
\end{array}\right.
\label{eq:ECNetForwardPass}
\end{equation}
where $W_1 \in \mathbb{R}^{m_\iota\times N_h^1}$, $W_{i+1} \in \mathbb{R}^{N_h^i\times N_h^{i+1}}$, and $W_5 \in \mathbb{R}^{N_h^4\times 1}$ are weight matrices, and $\vv{b}_i$ is a bias vector.

As in \cite{Larios;Gibou;KECNet2D;2022}, we have used TensorFlow's backpropagation and its Adam optimizer \cite{Adam;2015} to adapt the parameters in \cref{eq:ECNetForwardPass}.  In particular, we have chosen to minimize the \textit{root mean squared error} loss (RMSE) over the training subset.  Initially, the weights are set randomly to a uniform Glorot distribution \cite{Glorot;Bengio;2010}, and the biases start at zero.  Then, stochastic gradient descent periodically updates the neural parameters after evaluating batches of 64 samples for up to one thousand epochs.  At the same time, we monitor the \textit{mean absolute error} (MAE) to halt optimization whenever the validation subset's MAE stagnates for fifty iterations.  This validation error helps stabilize the learning process, too.  In Keras, we have attained the latter by integrating a plateau-detecting callback function that halves the learning rate from $\eten{1.5}{-4}$ to $\eten{1}{-5}$ every fifteen epochs of no MAE improvement.  In like manner, Keras has allowed us to introduce hidden-layer-wise kernel {\tt L2} regularizers.  Here, we have adopted that type of penalization\footnote{Also known as \textit{Tikhonov regularization}.} because it has been shown to boost generalization by acting as a weight decay or gradual forgetting mechanism during the parameter updates \cite{A18}.  For brevity, we have omitted a thorough presentation of the mathematical aspects and the inner workings of standard backpropagation; however, we refer the interested reader to \cite{A18, DeepLearning;Goodfellow-et-al;2016, Tensorflow15, Keras15} for additional information and tutorials.

To conclude, we comment on the third-party utilities employed to realize \Cref{alg:MLCurvature,alg:Preprocess}.  These are Lohmann's {\tt json} library \cite{json;2021}, Hermann's {\tt frugally-deep} \cite{frugally-deep;2021}, and OpenBLAS \cite{openblas;2021}.  First, the {\tt json} library and a selected group of {\tt frugally-deep} methods have helped us import the network parameters and $\texttt{Q}$ from Python into C++.  Once there, we used OpenBLAS to carry out neural inference efficiently as 32-bit {\tt sgemm}\footnote{A \textbf{s}ingle-precision \textbf{ge}neral \textbf{m}atrix-\textbf{m}atrix multiply operation of the form $C = \alpha AB + \beta C$.  In our case, $\alpha = 1$ and $\beta = 0$.  See \cite{gemm;intel;2022} for more information.} matrix operations.  Indeed, single-precision computations are faster than 64-bit operations, both at inference time and during training.  Similarly, single-precision has been essential in this study for saving disk space by lowering the 110-feature storage cost by half.


\FloatBarrier
\colorsection{Results}
\label{sec:Results}

In this section, a series of experiments illustrate how neural models constructed with the workflow above can improve numerical mean-curvature estimations in $\mathbb{R}^3$.  Unless otherwise stated, we consider $\eta = 6$ as a base case (i.e., $h = 2^{-6} = 0.015625$).  But, any user-defined $\eta$ should lead to comparable results.  In fact, we have discovered that level-set $h$-normalization and dimensionality reduction on the correlation matrix have made it possible to deploy our neural networks transparently across grid resolutions with marginal accuracy variations.  This observation is also important because it allows the researcher to use $\mathcal{F}_\kappa^{ns}(\cdot)$ and $\mathcal{F}_\kappa^{sd}(\cdot)$ even in uniform grids where $h \neq 2^{-\eta}$ and $\eta \in \mathbb{Z}^+$.

\begin{table}[!t]
	\centering
	\small
	\bgroup
	\def\arraystretch{1.1}%
	\begin{tabular}{cccccccc}
		\hline
		\makecell{Neural\\function} & $m_\iota$ & $N_h^i$ & {\tt L2} factor & \makecell{Training\\subset size} & \makecell{Testing\\subset size} & \makecell{Validation\\subset size} & \makecell{Number of\\parameters} \\
		\hline \hline
		$\mathcal{F}_\kappa^{ns}(\cdot)$ & 72   & 140 & $\eten{2}{-6}$ & 4,491,984 & 962,739 & 962,694 & 69,581 \\
		\hline
		$\mathcal{F}_\kappa^{sd}(\cdot)$ & 80   & 140 & $\eten{5}{-6}$ & 3,680,753 & 788,812 & 788,786 & 70,701 \\
		\hline
	\end{tabular}
	\egroup
	\caption{Architectural data for $\mathcal{F}_\kappa^{ns}(\cdot)$ and $\mathcal{F}_\kappa^{sd}(\cdot)$ with $\eta = 6$.  Here, $i = 1, \dots, 4$ enumerates the first hidden layers, as shown in \cref{fig:ECNet}.}
	\label{tbl:results.train.config}
\end{table}

The resource-intensive, generative routines in \Cref{sec:Methodology} led to 12.3GB and 6.2GB worth of non-saddle and saddle data for $\eta = 6$.  Afterward, we randomly took 67\% of the spherical data and 60\% of the non-saddle sinusoidal samples to assemble a more practical $\mathcal{D}^{ns}$ while keeping 100\% of the saddle data for $\mathcal{D}^{sd}$.  Columns 5 to 7 in \cref{tbl:results.train.config} outline the number of learning tuples in each subset.  With these data, we then optimized $\mathcal{F}_\kappa^{ns}(\cdot)$ and $\mathcal{F}_\kappa^{sd}(\cdot)$, settling at the network configurations shown in the rest of \cref{tbl:results.train.config}.  In the end, these optimal architectures produced the error statistics displayed for $\mathcal{D}^{ns}$ and $\mathcal{D}^{sd}$ in \cref{tbl:results.train.stats}.  Clearly, our MLPs can reduce the \textit{maximum absolute error} (MaxAE) at least by a factor of five for non-saddle stencils and four for saddle patterns.

\begin{table}[!t]
	\centering
	\small
	\bgroup
	\def\arraystretch{1.1}%
	\begin{tabular}{llcccc}
		\hline
		Type of stencils            & Method & MAE & MaxAE & RMSE & Epochs \\
		\hline \hline
		\multirow{2}{*}{Non-saddle} & $\mathcal{F}_\kappa^{ns}(\cdot)$ & $\eten{6.084535}{-4}$ & $\eten{4.837263}{-2}$ & $\eten{9.240574}{-4}$ & 602 \\
		                            & Baseline                         & $\eten{4.158100}{-2}$ & $\eten{2.475311}{-1}$ & $\eten{6.910001}{-2}$ & - \\
		\hline
		\multirow{2}{*}{Saddle}     & $\mathcal{F}_\kappa^{sd}(\cdot)$ & $\eten{2.056113}{-3}$ & $\eten{1.317899}{-1}$ & $\eten{3.008280}{-3}$ & 485 \\
		                            & Baseline                         & $\eten{1.183315}{-1}$ & $\eten{5.379409}{-1}$ & $\eten{1.547914}{-1}$ & - \\
		\hline
	\end{tabular}
	\egroup
	\caption{Training $h\kappa$ and $h\kappa_\mathcal{F}$ error statistics for $\eta = 6$ over $\mathcal{D}^{ns}$ and $\mathcal{D}^{sd}$.}
	\label{tbl:results.train.stats}
\end{table}

\begin{figure}[!t]
	\centering
	\begin{subfigure}[!t]{6cm}
		\includegraphics[width=\textwidth]{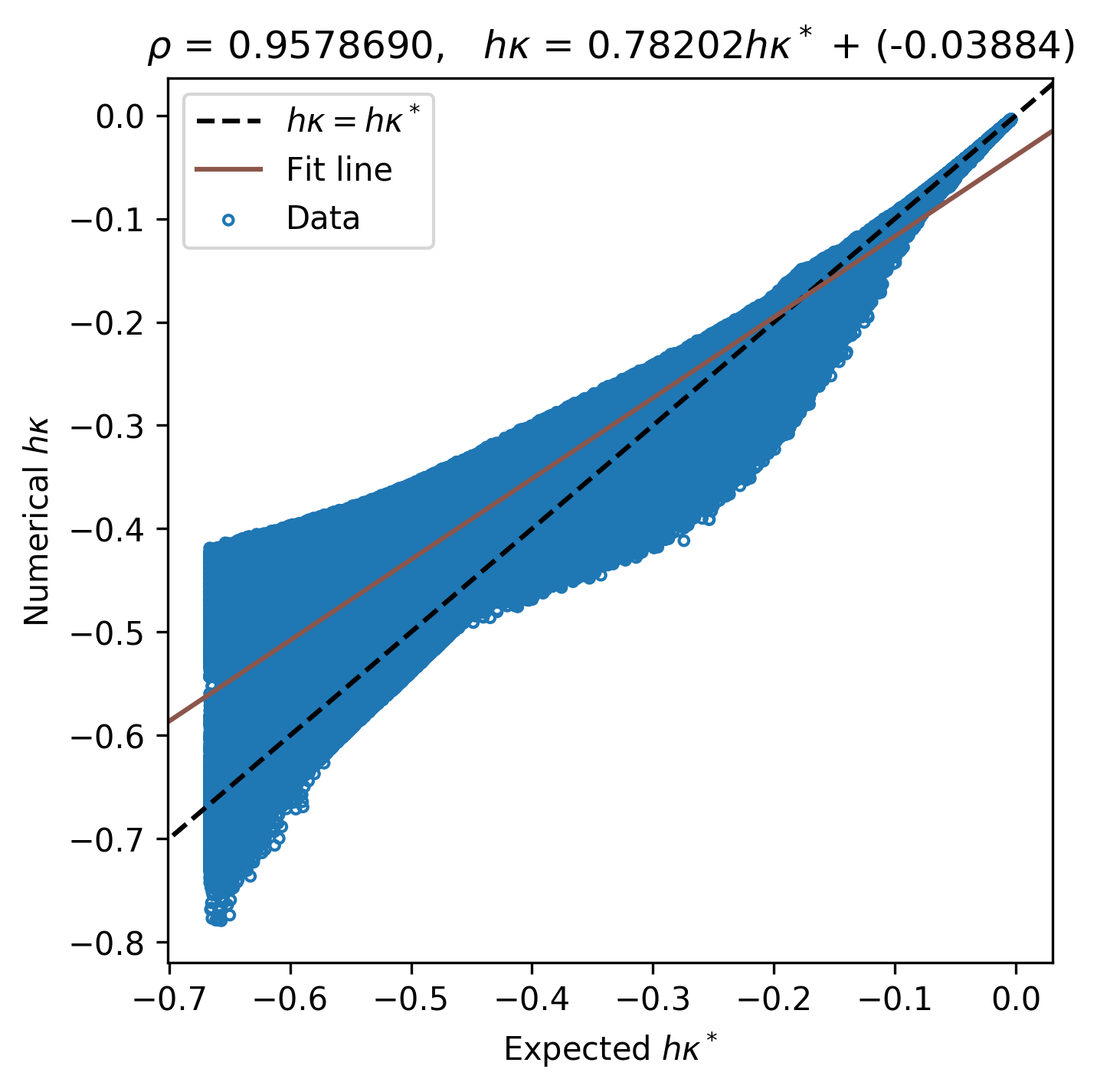}
	\end{subfigure}
	~
	\begin{subfigure}[!t]{6cm}
		\includegraphics[width=\textwidth]{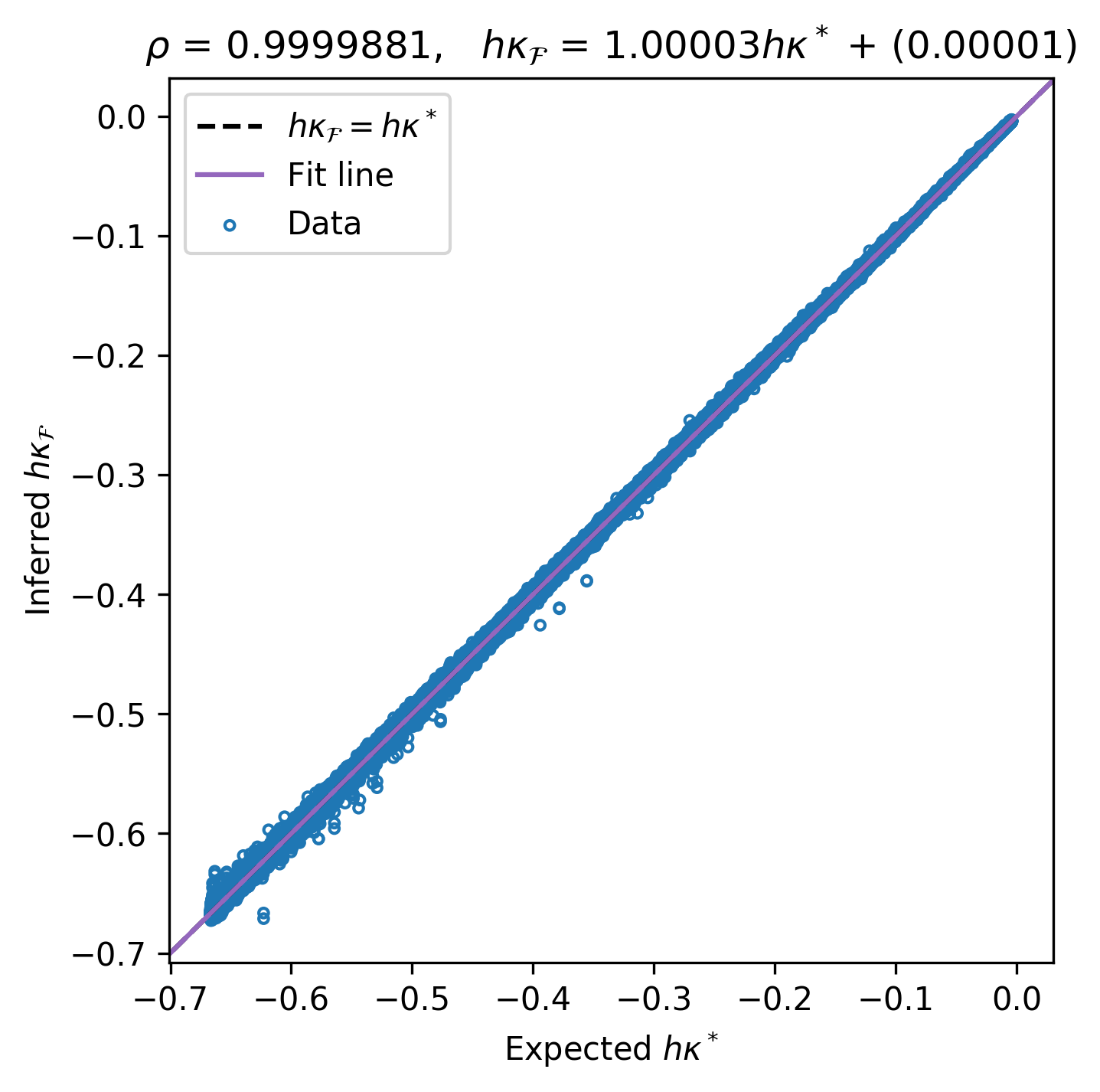}
	\end{subfigure}
	\\
	\begin{subfigure}[!t]{\textwidth}
		\caption{\footnotesize Non-saddle stencils}
		\label{fig:results.train.charts.6.non-saddle}
    \end{subfigure}
	\\
    
	\begin{subfigure}[!t]{6cm}
		\includegraphics[width=\textwidth]{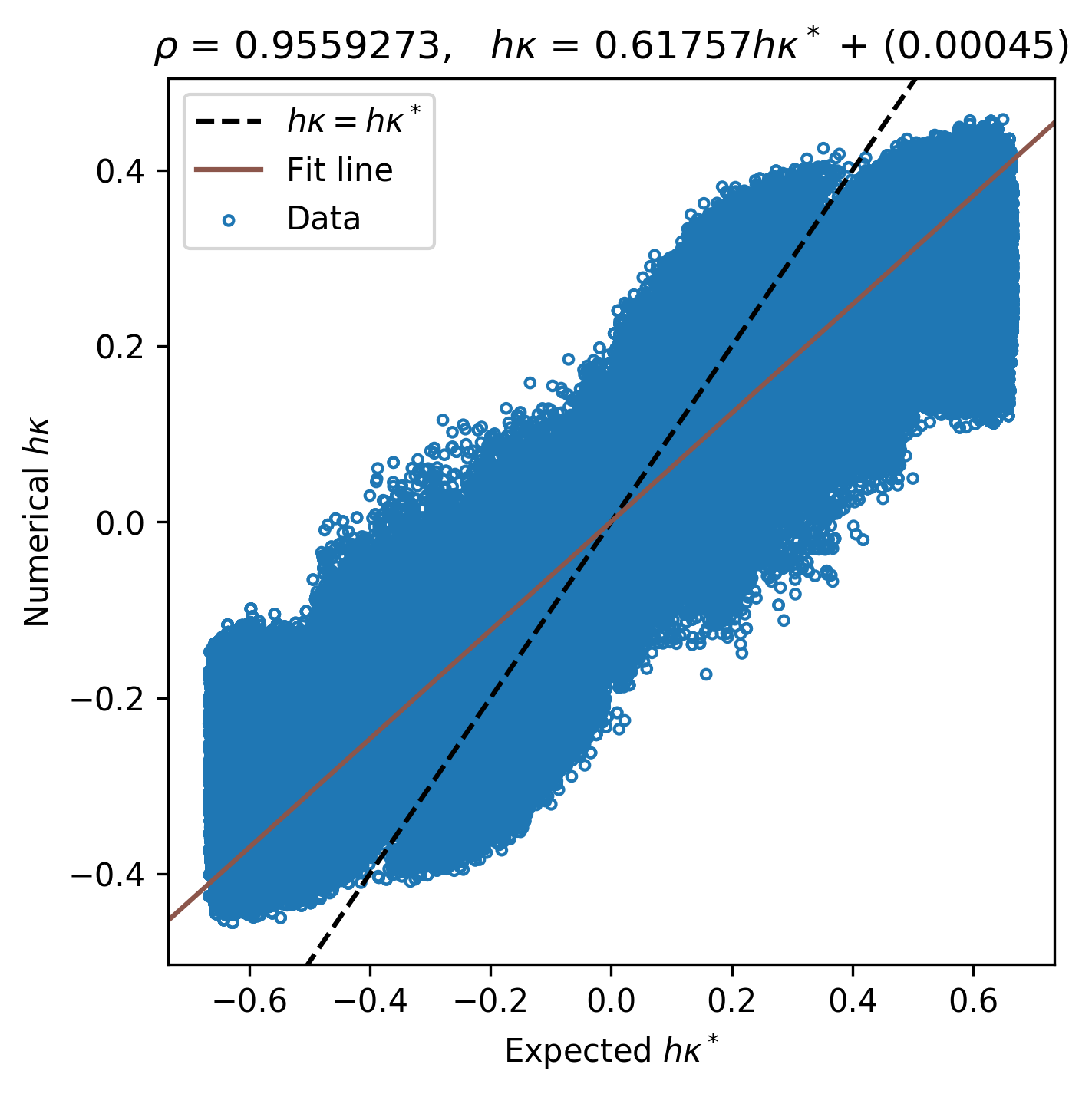}
	\end{subfigure}
	~
	\begin{subfigure}[!t]{6cm}
		\includegraphics[width=\textwidth]{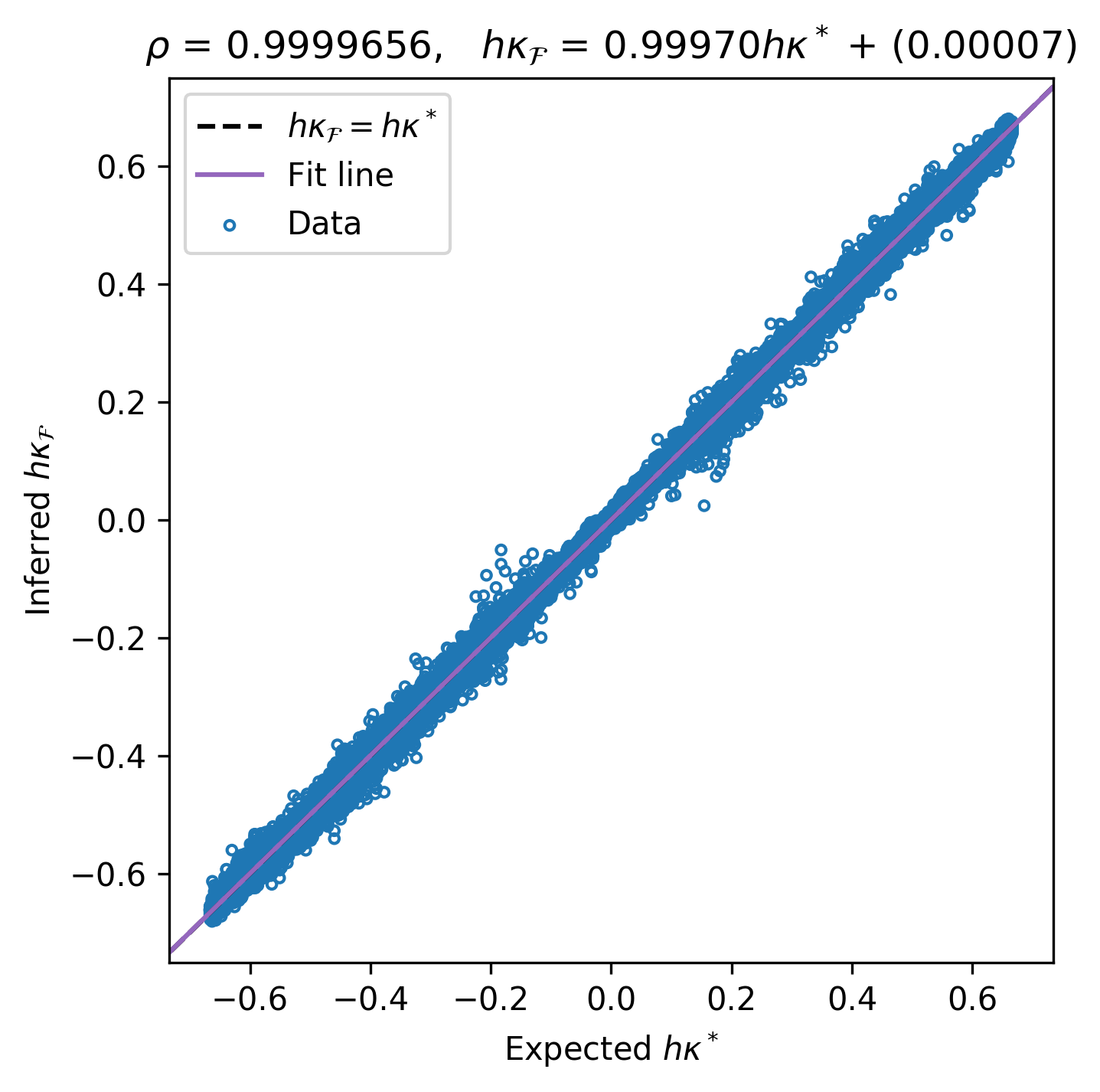}
	\end{subfigure}
	\\
	\begin{subfigure}[!t]{\textwidth}
		\caption{\footnotesize Saddle stencils}
		\label{fig:results.train.charts.6.saddle}
	\end{subfigure}
	\\
	\caption{Correlation plots showing the training outcomes for $\eta = 6$ on (a) non-saddle and (b) saddle stencils.  The left column illustrates the quality of the numerical fit (i.e., the $h\kappa$ baseline), while the right column portrays the $h\kappa_\mathcal{F}$ estimations over the entire $\mathcal{D}^{ns}$ and $\mathcal{D}^{sd}$.  Each chart has its $\rho$ correlation factor and the equation of the best regression line through the data.  (Color online.)}
	\label{fig:results.train.charts}
\end{figure}

\Cref{fig:results.train.charts} validates the results in \cref{tbl:results.train.stats} by contrasting the quality of the fit of our neural functions and the numerical baseline.  In particular, the higher $\rho$ correlation factors of $\mathcal{F}_\kappa^{ns}(\cdot)$ and $\mathcal{F}_\kappa^{sd}(\cdot)$ demonstrate the ability of our error-quantifiers to fix $h\kappa$ along under-resolved and steep interface regions.  For the interested reader, we have released these and the neural networks trained for other mesh resolutions as {\tt HDF5} and {\tt Base64}-encoded {\tt JSON} files at \url{https://github.com/UCSB-CASL/Curvature_ECNet_3D}.  In the same repository, we have provided the preprocessing objects from \Cref{alg:Preprocess} as {\tt JSON} and {\tt pickle} files alongside testing data and their supplementary documentation.

\colorsubsection{Geometrical tests}
\label{subsec:GeometricalTests}

\begin{figure}[!t]
	\centering
	\begin{subfigure}[!t]{0.32\textwidth}
		\includegraphics[width=\textwidth]{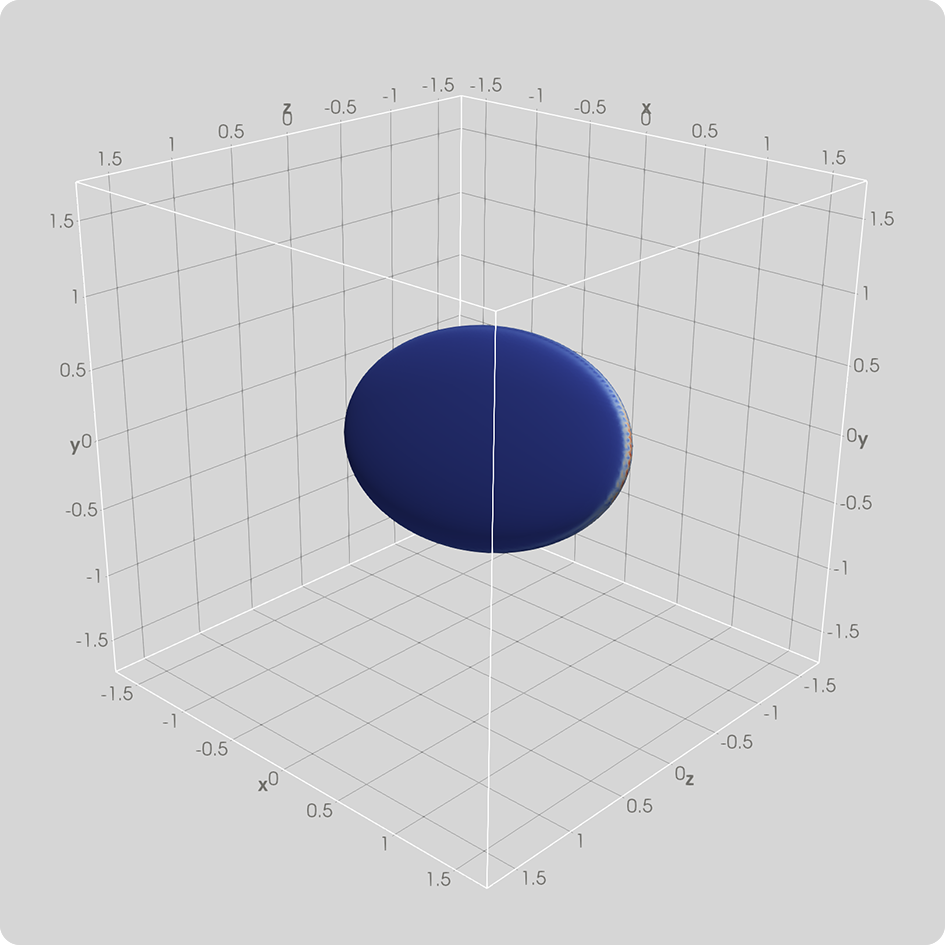}
		\caption{\footnotesize Ellipsoidal interface}
		\label{fig:results.geometries.ellipsoid}
	\end{subfigure}
	~
	\begin{subfigure}[!t]{0.32\textwidth}
		\includegraphics[width=\textwidth]{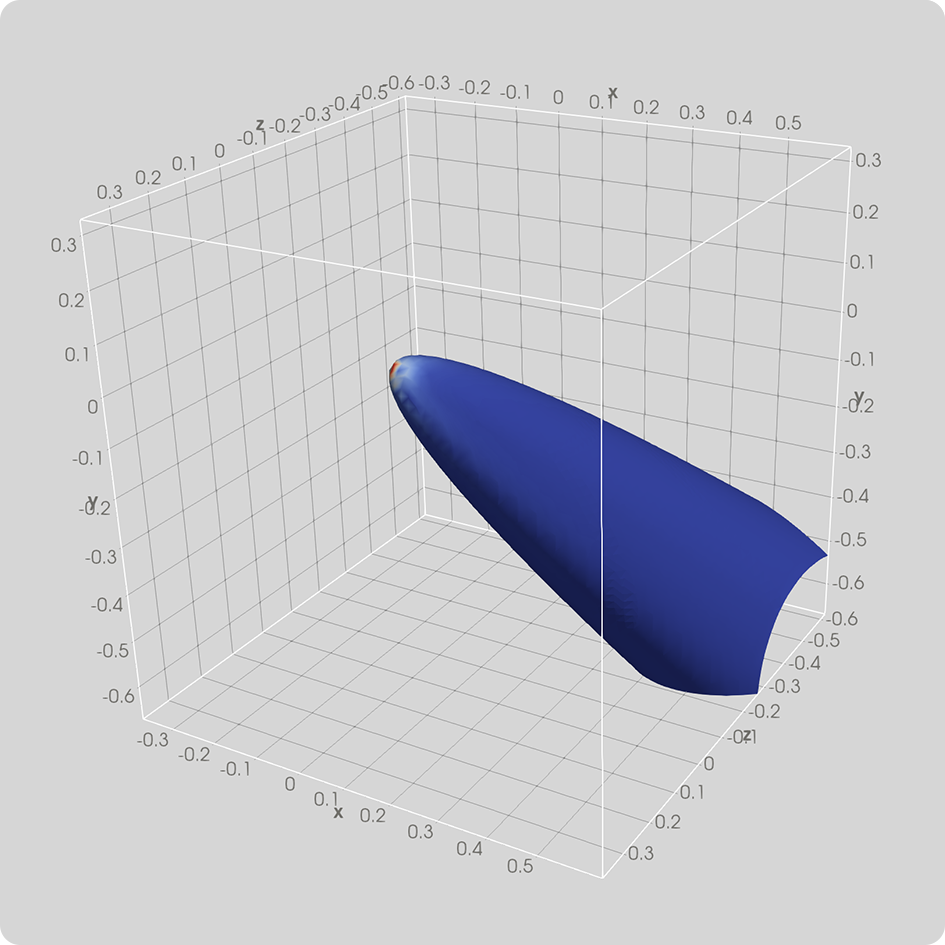}
		\caption{\footnotesize Paraboloidal interface}
		\label{fig:results.geometries.paraboloid}
	\end{subfigure}
	~
	\begin{subfigure}[!t]{0.32\textwidth}
		\includegraphics[width=\textwidth]{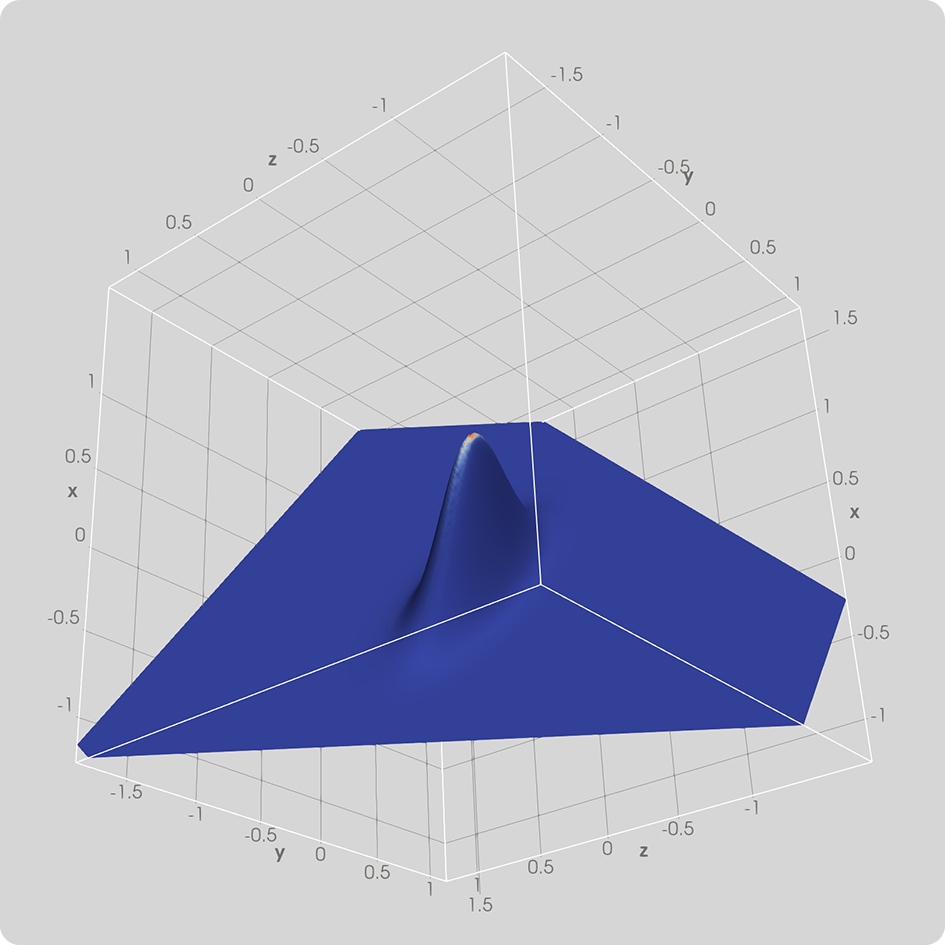}
		\caption{\footnotesize Gaussian interface}
		\label{fig:results.geometries.gaussian}
	\end{subfigure}
	
	\caption{Stationary geometries tested in \Crefrange{subsubsec:AnEllipsoidalInterface}{subsubsec:AGaussianInterface}.  We have shaded the interfaces in dark blue with diverging transitions toward red in regions of high mean-curvature error.  (Color online.)}
	\label{fig:results.geometries}
\end{figure}

In this section, we evaluate our hybrid strategy with three stationary geometries not accounted for during training.  These geometries---an ellipsoid, a paraboloid, and a Gaussian shown in \cref{fig:results.geometries}---represent the zero-isosurfaces of their respective level-set functions.  Then, we validate the {\tt MLCurvature()} routine on a sphere morphing into an ellipsoid with large mean curvatures at its semi-axes.  In principle, our starting point is $\eta = 6$ for all scenarios, but we have introduced $\eta = 7$ and $\eta = 8$ when appropriate to compare the conventional approach and \Cref{alg:MLCurvature}.

Before addressing each case, we briefly describe the general experimental setup.  First, we prescribe a random affine transformation (see \cref{eq:AffineTransform,eq:InverseAffineTransform}) and discretize $\Omega$ with adaptive Cartesian grids accordingly.  Such a discretization must enforce a uniform band of half-width $3h$ around the affine-transformed interface, where $h = 2^{-\eta}$.  Then, we evaluate $\phi(\cdot)$ by calculating ``exact'' signed distances to $\Gamma$ at least within a (linearly approximated) shell of half-width $3h\sqrt{3}$.  These normal distances have resulted from solving the nearest-location problem with {\tt dlib}, as in \Cref{alg:GenerateSinusoidalDataSets,alg:GenerateHypParaboloidalDataSet}.  After that, we perturb the nodal level-set values with uniform random noise in the range of $\epsilon_{rnd}\cdot[-h, +h]$, where $\epsilon_{rnd} = \eten{1}{-4}$.  Finally, we redistance our level-set function with $\nu = 10$ steps and estimate $h\kappa$ at $\Gamma$ with \Cref{alg:MLCurvature} for all the interface nodes.  Note that we have continued to use $h\kappa_{G,\min}^{ns} = \eten{-7}{-6}$ to distinguish non-saddle from saddle stencils.


\colorsubsubsection{An ellipsoidal interface}
\label{subsubsec:AnEllipsoidalInterface}

Consider an ellipsoid parametrized in spherical coordinates by

\begin{equation}
\vv{\elp}(\vartheta, \varphi) \doteq [a_{el} \cos\varphi \cos\vartheta,\, b_{el} \cos\varphi \sin\vartheta,\, c_{el}\sin\varphi]^T,
\label{eq:EllipsoidalSurface}
\end{equation}
where $-\frac{\pi}{2} \leqslant \varphi \leqslant \frac{\pi}{2}$ is the reduced latitude (as measured from the $uv$-plane), $0 \leqslant \vartheta < 2\pi$ is the longitude, and $a_{el}$, $b_{el}$, and $c_{el}$ are the semi-axis lengths in its canonical frame $\mathcal{C}_{el}$ \cite{Nurnberg;DistToEllipsoid;2006}.  Furthermore, suppose $\vv{p} = [u,\, v,\, w]^T = \vv{\elp}(\vartheta, \varphi)$ is some query point on the surface.  Then, we can derive and evaluate the mean and Gaussian curvatures at $\vv{p}$ as

\begin{equation}
\kappa_{el}(\vv{p}) = \frac{-\left(\frac{u^2}{a_{el}^6}+\frac{v^2}{b_{el}^6}+\frac{w^2}{c_{el}^6}\right) + \left(\frac{u^2}{a_{el}^4}+\frac{v^2}{b_{el}^4}+\frac{w^2}{c_{el}^4}\right)\left(\frac{1}{a_{el}^2}+\frac{1}{b_{el}^2}+\frac{1}{c_{el}^2}\right)}{2\left(\frac{u^2}{a_{el}^4}+\frac{v^2}{b_{el}^4}+\frac{w^2}{c_{el}^4}\right)^{3/2}} \quad \textrm{and} \quad
\kappa_{G,el}(\vv{p}) = \frac{1}{a_{el}^2 b_{el}^2 c_{el}^2 \left(\frac{u^2}{a_{el}^4}+\frac{v^2}{b_{el}^4}+\frac{w^2}{c_{el}^4}\right)^2}
\label{eq:EllipsoidalCurvatures}
\end{equation}
by using \cref{eq:GaussianCurvature.Compact} \cite{CurvatureFormulasForImplicit;Goldman;2005} on the ellipsoid's implicit form

\begin{equation}
I_{el}(x,y,z) = \frac{x^2}{a_{el}^2} + \frac{y^2}{b_{el}^2} + \frac{z^2}{c_{el}^2} - 1.
\label{eq:EllipsoidalSurfaceImplicit}
\end{equation}

Putting together \cref{eq:EllipsoidalSurface,eq:EllipsoidalSurfaceImplicit} leads us to this test's level-set function

\begin{equation}
\phi_{el}(\vv{x}) = \left\{\begin{array}{ll}
	-\dist{\vv{x}}{\vv{\elp}(\cdot)} & \textrm{if } I_{el}(\vv{x}) < 0 \\
	0                                & \textrm{if } I_{el}(\vv{x}) = 0 \;(\textrm{i.e., } \vv{x} \in \Gamma), \\
	+\dist{\vv{x}}{\vv{\elp}(\cdot)} & \textrm{if } I_{el}(\vv{x}) > 0, 
\end{array}\right.
\label{eq:EllipsoidalLevelSetFunction}
\end{equation}
where $\dist{\cdot}{\cdot}$ yields the shortest distance between $\vv{x} \in \mathbb{R}^3$ (expressed in terms of $\mathcal{C}_{el}$) and the surface.  This definition partitions the computational domain in such a way that $\Omega^-$ lies inside $\Gamma \equiv \vv{\elp}(\cdot)$, and $\Omega^+$ corresponds to the exterior space.  Moreover, since $\kappa_{G,el}(\cdot) > 0$, we expect mostly non-saddle stencils enabling $\mathcal{F}_\kappa^{ns}(\cdot)$ in \Cref{alg:MLCurvature}.

In this experiment, we have chosen $a_{el} = 1.65$, $b_{el} = 0.75$, and $c_{el} = 0.2$, so the $h\kappa^*$ local extrema at the $u$-, $v$-, and $w$-intercepts are 0.345182, 0.148637, and $\eten{3.351699}{-3}$.  \Cref{fig:results.geometries.ellipsoid} pictures the ellipsoid as $\phi_{el}(\vv{x})$'s zero-isosurface while emphasizing regions along $\Gamma$ where $|\bar{\varepsilon}|$ is large.  The statistics for our hybrid method and the numerical baseline appear in \cref{tbl:results.ellipsoid.stats}.  For reference, we have included the error metrics and costs for estimating mean curvatures numerically at twice and four times the grid resolution (i.e., at $\eta = 7, 8$).  Furthermore, we have wall-timed\footnote{In C++ 14, we enabled compiling optimization with the options {\tt -O2 -O3 -march=native}.} the evaluations using three {\tt MPI} tasks on a MacBook Pro with six 2.2 GHz Intel i7 cores and 16 GB RAM.  We have selected the best runtimes out of ten trials in all the assessments in this section.

\begin{table}[!t]
	\centering
	\small
	\bgroup
	\def\arraystretch{1.1}%
	\begin{tabular}{l|l|ccc|ccc|cc}
		\cline{3-10}
		\multicolumn{2}{c|}{} & \multicolumn{3}{c|}{MAE} & \multicolumn{3}{c|}{MaxAE} & \multicolumn{2}{c}{Performance} \\
		\hline
		Method & $\eta$ & $h\kappa$ & $\kappa$ & \makecell{Improv.\\Factor} & $h\kappa$ & $\kappa$ & \makecell{Improv.\\Factor} & \makecell{Time\\(secs.)} & \makecell{Cost\\(\%)} \\
		\hline \hline
		{\tt MLCurvature()}       & 6 & $\eten{1.2291}{-4}$ & $\eten{7.8663}{-3}$     &    - 
									  & $\eten{8.5979}{-3}$ & $\eten{5.5027}{-1}$     &    - 
									  & 6.5107 &        - \\
		\hline
		\multirow{3}{*}{Baseline} & 6 & $\eten{1.1972}{-3}$ & $\eten{7.6621}{-2}$     & 9.74 
		                   			  & $\eten{7.4814}{-2}$ & $4.7881$                & 8.70 
								 	  & 4.7991 & $+35.67$ \\
		                          & 7 & $\eten{1.7360}{-4}$ & $\eten{2.2221}{-2}$     & 2.82 
									  & $\eten{1.5320}{-2}$ & $1.9610$                & 3.56 
									  & 19.3433 & $-66.34$ \\
								  & 8 & $\eten{3.7271}{-5}$ & $\eten{9.5415}{-3}$     & 1.21 
									  & $\eten{2.6118}{-3}$ & $\eten{6.6863}{-1}$     & 1.22 
									  & 85.8928 & $-92.42$ \\
		\hline
	\end{tabular}
	\egroup
	\caption{Error and performance statistics for the ellipsoidal-surface test in \Cref{subsubsec:AnEllipsoidalInterface}.}
	\label{tbl:results.ellipsoid.stats}
\end{table}

\begin{figure}[!t]
	\centering
	\begin{subfigure}[!t]{6cm}
		\includegraphics[width=\textwidth]{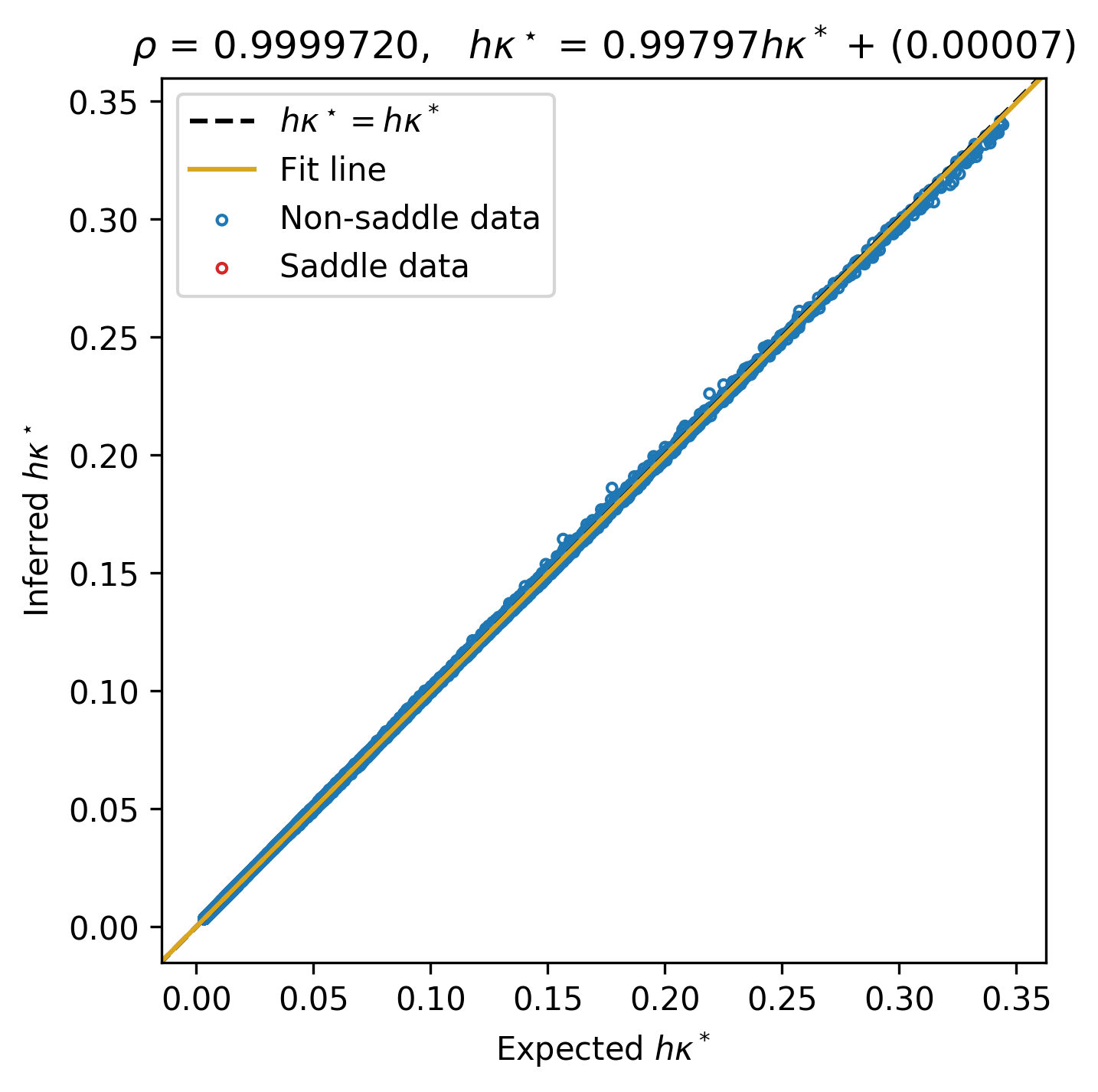}
		\caption{\footnotesize {\tt MLCurvature()} ($h = 2^{-6}$)}
		\label{fig:results.ellipsoid.charts.6.nnet}
	\end{subfigure}
	~
	\begin{subfigure}[!t]{6cm}
		\includegraphics[width=\textwidth]{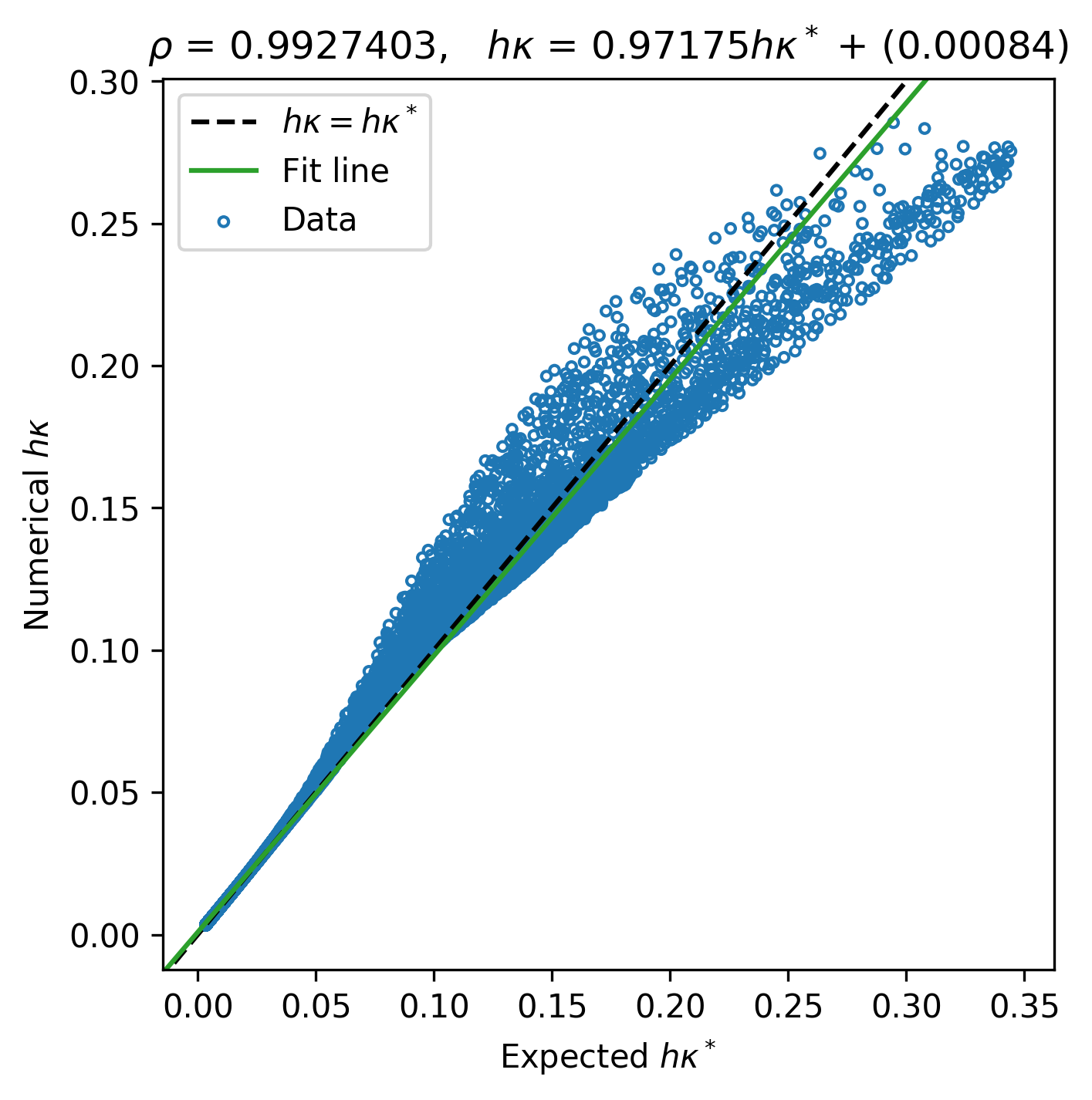}
		\caption{\footnotesize Baseline ($h = 2^{-6}$)}
		\label{fig:results.ellipsoid.charts.6.num}
	\end{subfigure}
	\\
    
	\begin{subfigure}[!t]{6cm}
		\includegraphics[width=\textwidth]{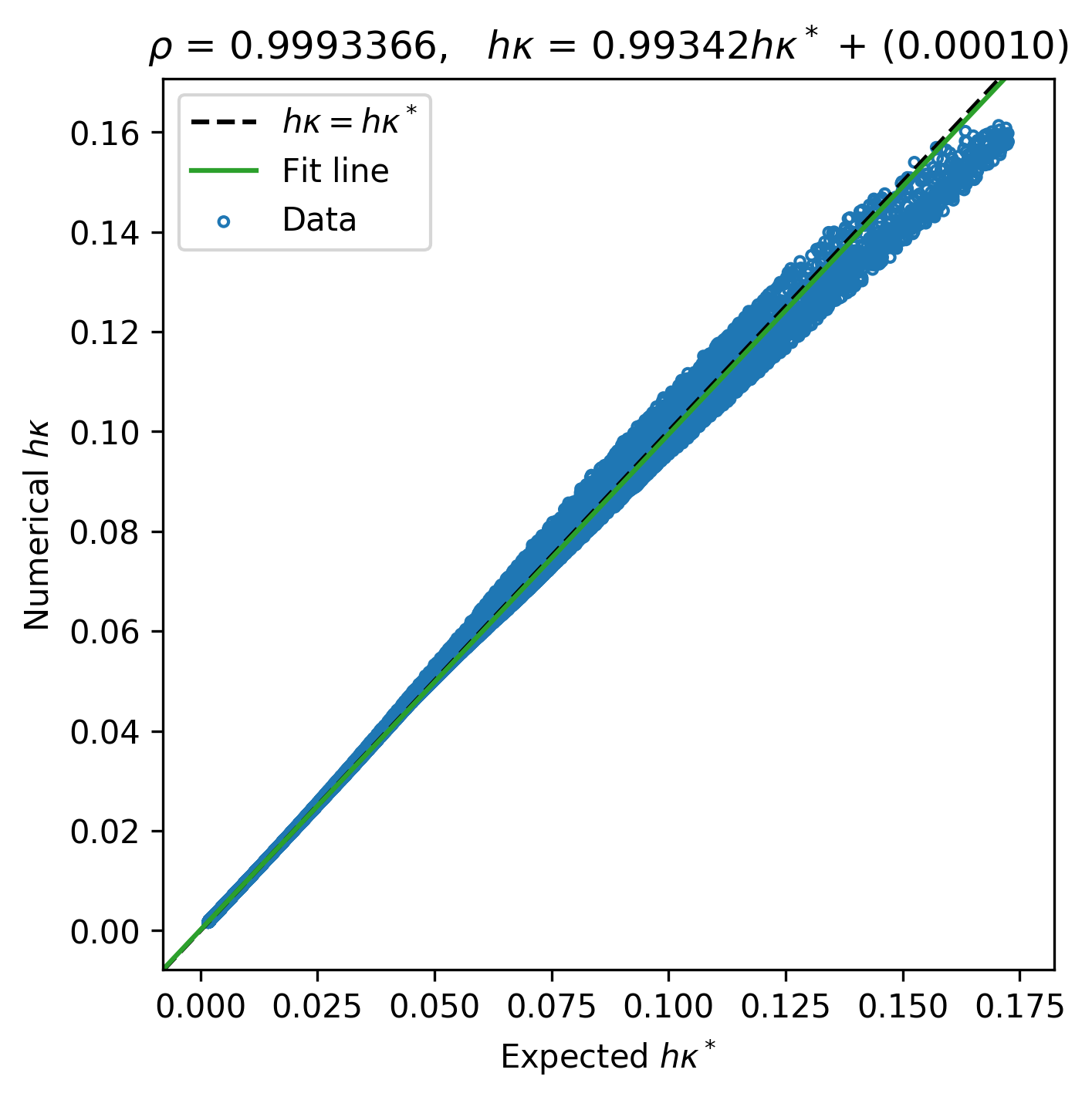}
		\caption{\footnotesize Baseline ($h = 2^{-7}$)}
		\label{fig:results.ellipsoid.charts.7.num}
	\end{subfigure}
	~
	\begin{subfigure}[!t]{6cm}
		\includegraphics[width=\textwidth]{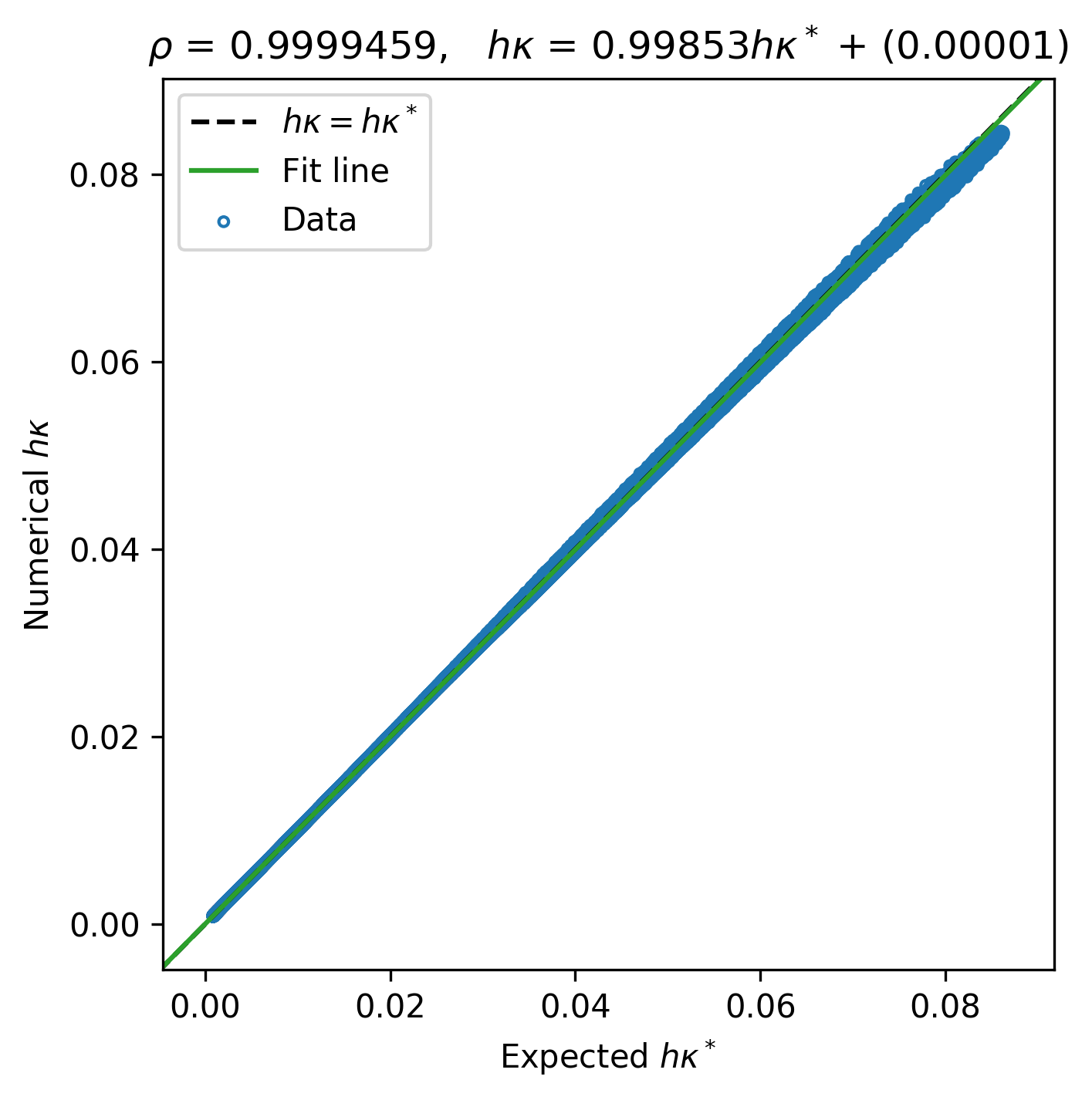}
		\caption{\footnotesize Baseline ($h = 2^{-8}$)}
		\label{fig:results.ellipsoid.charts.8.num}
	\end{subfigure}
	
	\caption{Correlation plots for estimated $h\kappa$ along an ellipsoid, as described in \Cref{subsubsec:AnEllipsoidalInterface}.  We display the quality of the approximations of \Cref{alg:MLCurvature} in (a) and the numerical baseline in (b) for $\eta = 6$.  The bottom row provides the corresponding charts for the baseline at (c) $\eta = 7$ and (d) $\eta = 8$.  In (a), non-saddle stencils appear in blue (\textcolor{bluematlab}{$\vv{\circ}$}) and saddle data in red (\textcolor{redmatlab}{$\vv{\circ}$}).  (Color online.)}
	\label{fig:results.ellipsoid.charts}
\end{figure}

\Cref{tbl:results.ellipsoid.stats} provides the $L^1$ and $L^\infty$ error norms for $h\kappa$ and $\kappa$ to ease the comparisons between the solvers at the same and different mesh sizes.  In particular, the \textit{improvement factor} column indicates the $\kappa$-error ratio from the baseline to our hybrid inference system.  Similarly, the \textit{cost} column shows how much more ($+\%$) or less ($-\%$) time it takes for the {\tt MLCurvature()} routine to solve the same problem as the numerical approach.  At $\eta = 6$, for example, our strategy can reduce the estimation error by a factor of at least $8.7$ while incurring an additional overhead of $35.67\%$.  This inference cost might seem discouraging, but we can justify it by analyzing the rows of $\eta = 7$ and $8$.  In this regard, not only does increasing the grid resolution improve the numerical accuracy, but it also has a detrimental effect on the baseline performance.  Consider $\eta = 7$, for instance.  In that case, \cref{tbl:results.ellipsoid.stats} reveals that the conventional schemes alone cannot guarantee the same precision as {\tt MLCurvature()}, which now requires one-third of the cost.  In addition, the statistics in this table suggest that our proposed method can attain a comparable level of precision to the baseline at four times the mesh size but at $7.58\%$ of the numerical cost.  \Cref{fig:results.ellipsoid.charts} complements these results by contrasting the quality of the evaluated approximations.  Clearly, our solver offers an inexpensive way to offset mean-curvature errors, with no need for further mesh refinement.

\colorsubsubsection{A paraboloidal interface}
\label{subsubsec:AParaboloidalInterface}

Next, consider a paraboloid denoted by the Monge patch

\begin{equation}
\vv{\qpr}(u,v) \doteq (u, v, \qpr(u,v)) \quad \textrm{with} \quad \qpr(u,v) = a_{pr}u^2 + b_{pr}v^2,
\label{eq:ParaboloidalInterface}
\end{equation}
where $a_{pr}$ and $b_{pr}$ are positive shape parameters.  Expressing the paraboloidal surface with \cref{eq:ParaboloidalInterface} allows us to find the exact mean and Gaussian curvatures at any query point $\vv{p} = [u,\, v]^T \in \mathbb{R}^2$ with

\begin{equation}
\kappa_{pr}(u,v) = \frac{(1+4b_{pr}^2v^2)a_{pr} + (1+4a_{pr}^2u^2)b_{pr}}{(1 + 4a_{pr}^2u^2 + 4b_{pr}^2v^2)^{3/2}} \quad \textrm{and} \quad
\kappa_{G,pr}(u,v) = \frac{4a_{pr}b_{pr}}{(1 + 4a_{pr}^2u^2 + 4b_{pr}^2v^2)^2},
\label{eq:ParaboloidalCurvatures}
\end{equation}
where $\kappa_{G,pr} > 0$ for all $\vv{p}$, at least theoretically.  In practice, the latter implies more frequent activations of the non-saddle path in \Cref{alg:MLCurvature}.

The patch definition in \cref{eq:ParaboloidalInterface} leads us to the level-set function

\begin{equation}
\phi_{pr}(\vv{x}) = \left\{\begin{array}{ll}
	-\dist{\vv{x}}{\vv{\qpr}(\cdot)} & \textrm{if } z > \qpr(x,y), \\
	0                                & \textrm{if } z = \qpr(x,y) \;(\textrm{i.e., } \vv{x} \in \Gamma), \\
	+\dist{\vv{x}}{\vv{\qpr}(\cdot)} & \textrm{if } z < \qpr(x,y),
\end{array}\right.
\label{eq:ParaboloidalLevelSetFunction}
\end{equation}
where $\vv{x} \in \mathbb{R}^3$ is given in the representation of the local coordinate system $\mathcal{C}_{pr}$.  Like $\phi_{hp}(\cdot)$ in \cref{eq:HypParaboloidalLevelSetFunction}, $\phi_{pr}(\cdot)$ partitions the computational domain into $\Omega^-$ above $\Gamma \equiv \vv{\qpr}(\cdot)$ and $\Omega^+$ right beneath it.

In this test, we have selected $a_{pr} = 25.6$ and $b_{pr} = 12.8$ so that $h\kappa^* = 0.6$ at the trough. In addition, we have introduced a customized parameter $c_{pr} = 0.5$ to set the paraboloid's height.  This parameter has helped us bound $\Omega$ and keep the entire affine-transformed surface from $w = 0$ to $w = c_{pr}$ within the computational domain.  \Cref{fig:results.geometries.paraboloid} illustrates the paraboloidal interface in the world coordinate system.  In analogy to \Cref{subsubsec:AnEllipsoidalInterface}, the mean-curvature accuracy results and the performance metrics for this experiment appear in \cref{tbl:results.paraboloid.stats}.

\begin{table}[!t]
	\centering
	\small
	\bgroup
	\def\arraystretch{1.1}%
	\begin{tabular}{l|l|ccc|ccc|cc}
		\cline{3-10}
		\multicolumn{2}{c|}{} & \multicolumn{3}{c|}{MAE} & \multicolumn{3}{c|}{MaxAE} & \multicolumn{2}{c}{Performance} \\
		\hline
		Method & $\eta$ & $h\kappa$ & $\kappa$ & \makecell{Improv.\\Factor} & $h\kappa$ & $\kappa$ & \makecell{Improv.\\Factor} & \makecell{Time\\(secs.)} & \makecell{Cost\\(\%)} \\
		\hline \hline
		{\tt MLCurvature()}       & 6 & $\eten{1.5456}{-4}$ & $\eten{9.8921}{-3}$ &    - 
									  & $\eten{3.3952}{-3}$ & $\eten{2.1729}{-1}$ &    - 
									  & 0.6854 &        - \\
		\hline
		\multirow{3}{*}{Baseline} & 6 & $\eten{1.6780}{-3}$ & $\eten{1.0739}{-1}$ & 10.86 
		                   			  & $\eten{1.3264}{-1}$ & $8.4893$            & 39.07 
								 	  & 0.5663 & $+21.04$ \\
		                          & 7 & $\eten{2.1512}{-4}$ & $\eten{2.7535}{-2}$ & 2.78 
									  & $\eten{2.6035}{-2}$ & $3.3324$            & 15.34 
									  & 1.4223 & $-51.81$ \\
								  & 8 & $\eten{4.0019}{-5}$ & $\eten{1.0245}{-2}$ & 1.04 
									  & $\eten{3.4807}{-3}$ & $\eten{8.9106}{-1}$ & 4.10 
									  & 4.2628 & $-83.92$ \\
		\hline
	\end{tabular}
	\egroup
	\caption{Error and performance statistics for the paraboloidal-surface test in \Cref{subsubsec:AParaboloidalInterface}.}
	\label{tbl:results.paraboloid.stats}
\end{table}

\begin{figure}[!t]
	\centering
	\begin{subfigure}[!t]{6cm}
		\includegraphics[width=\textwidth]{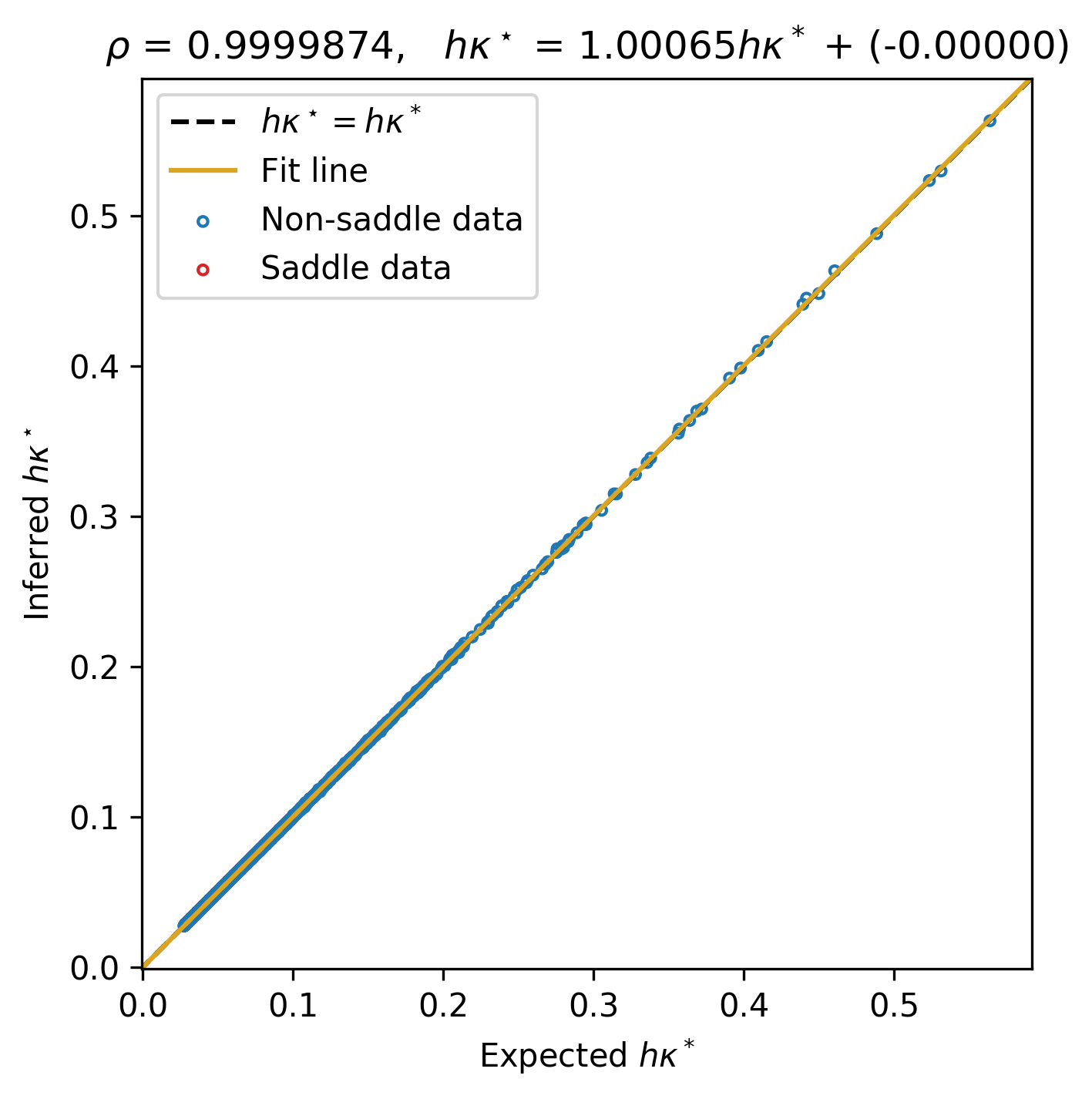}
		\caption{\footnotesize {\tt MLCurvature()} ($h = 2^{-6}$)}
		\label{fig:results.paraboloid.charts.6.nnet}
	\end{subfigure}
	~
	\begin{subfigure}[!t]{6cm}
		\includegraphics[width=\textwidth]{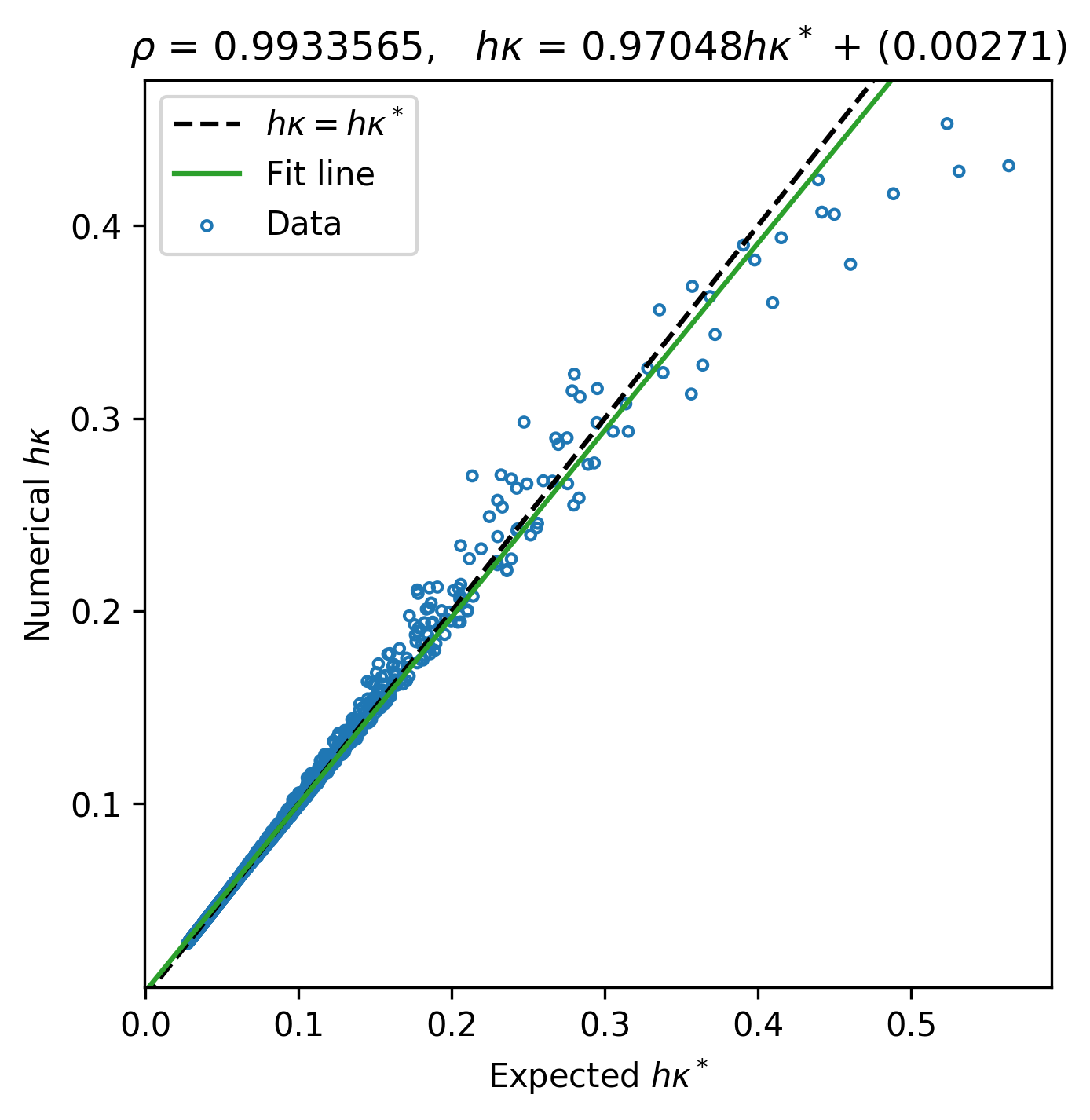}
		\caption{\footnotesize Baseline ($h = 2^{-6}$)}
		\label{fig:results.paraboloid.charts.6.num}
	\end{subfigure}
	\\
    \begin{subfigure}[!t]{6cm}
		\includegraphics[width=\textwidth]{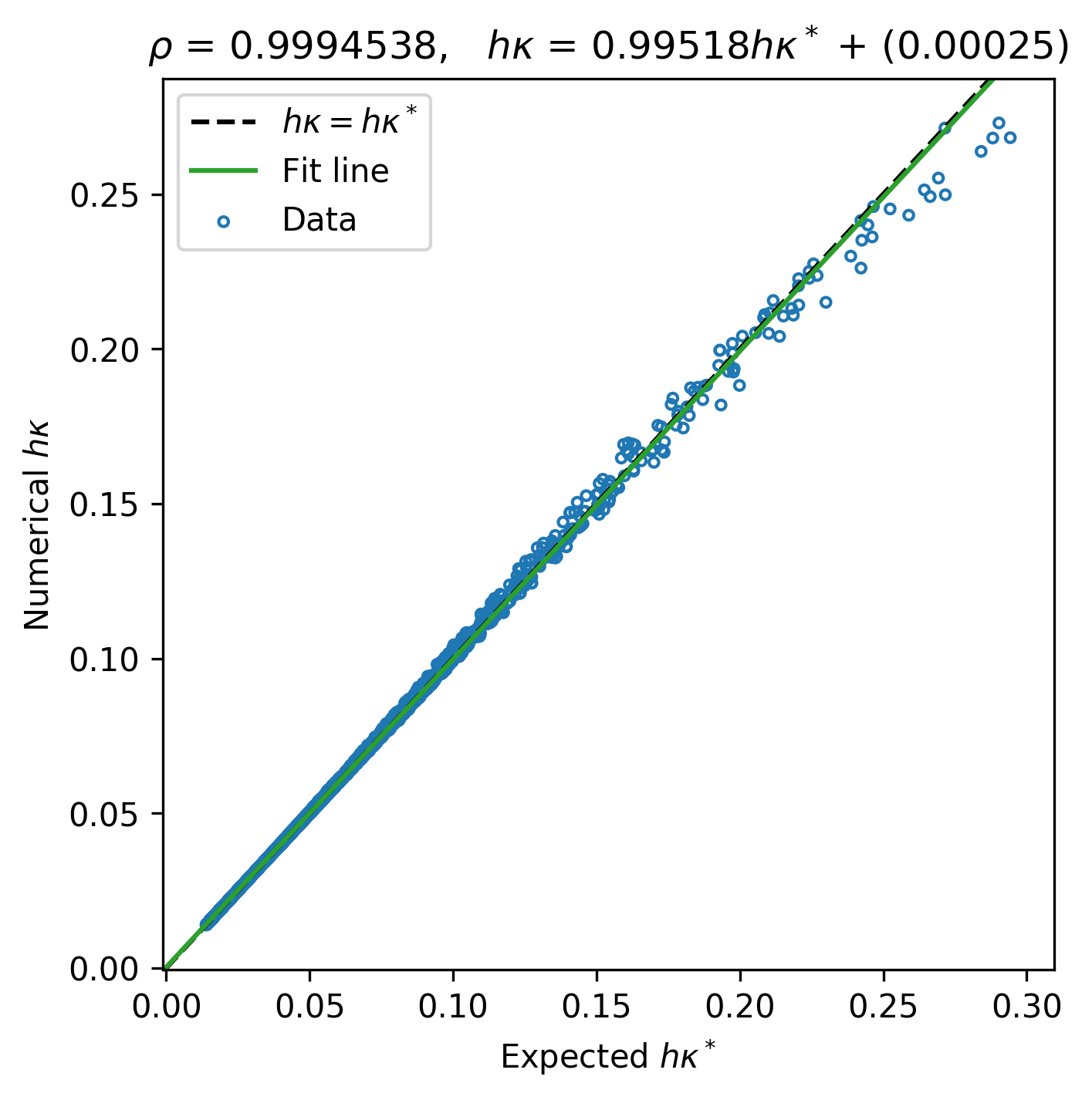}
		\caption{\footnotesize Baseline ($h = 2^{-7}$)}
		\label{fig:results.paraboloid.charts.7.num}
	\end{subfigure}
	~
	\begin{subfigure}[!t]{6cm}
		\includegraphics[width=\textwidth]{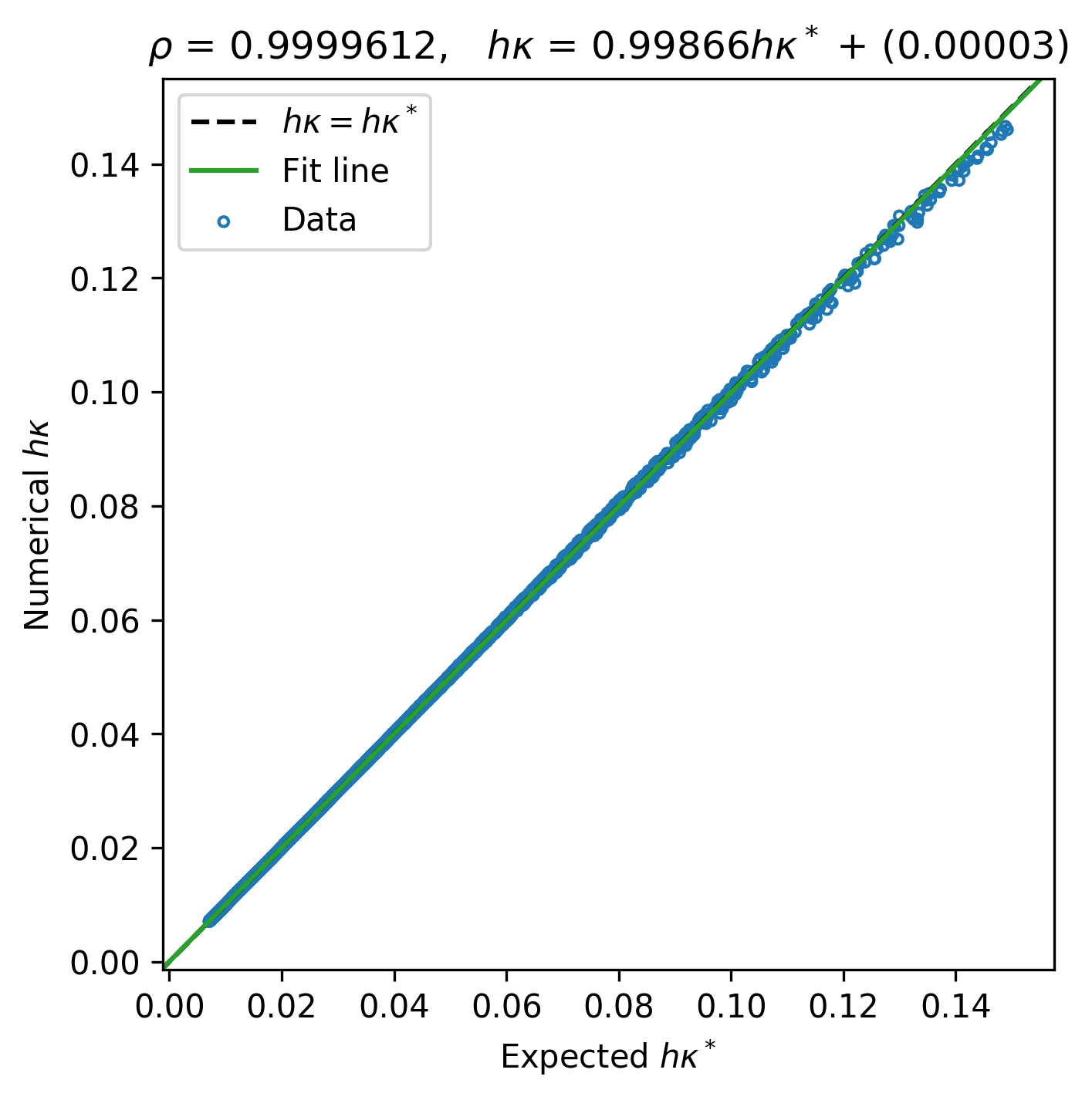}
		\caption{\footnotesize Baseline ($h = 2^{-8}$)}
		\label{fig:results.paraboloid.charts.8.num}
	\end{subfigure}
	
	\caption{Correlation plots for estimated $h\kappa$ along a paraboloid, as described in \Cref{subsubsec:AParaboloidalInterface}.  We display the quality of the approximations of \Cref{alg:MLCurvature} in (a) and the numerical baseline in (b) for $\eta = 6$.  The bottom row provides the corresponding charts for the baseline at (c) $\eta = 7$ and (d) $\eta = 8$.  In (a), non-saddle stencils appear in blue (\textcolor{bluematlab}{$\vv{\circ}$}) and saddle data in red (\textcolor{redmatlab}{$\vv{\circ}$}).  (Color online.)}
	\label{fig:results.paraboloid.charts}
\end{figure}

\Cref{tbl:results.paraboloid.stats}, once more, confirms the ability of our data-driven strategy to reduce mean-curvature errors significantly.  For example, when the paraboloid is embedded in a discretized grid with $\eta = 6$, we can get error-reduction factors of at least one order of magnitude and up to forty in the $L^1$ and $L^\infty$ norms.  Likewise, as discussed in \Cref{subsubsec:AnEllipsoidalInterface}, increasing $\eta$ to 7 and 8 improves the baseline mean-curvature approximations, but not at the same level as the {\tt MLCurvature()} routine when $\eta = 6$.  Indeed, our hybrid solver yields better estimations than the numerical approach at twice the resolution but requires only half of the cost.  \Cref{tbl:results.paraboloid.stats} also corroborates that \Cref{alg:MLCurvature} can attain a comparable level of precision to the standard level-set approach at four times the base mesh size, but at $16.08\%$ of the numerical cost.  Lastly, \cref{fig:results.paraboloid.charts} offers another perspective on the quality of the neural and numerical curvature computations by plotting them against $h\kappa^*$.  As expected, the baseline's entries look noisier than ours at $\eta = 6, 7$, particularly when handling steep interface regions.

\colorsubsubsection{A Gaussian interface}
\label{subsubsec:AGaussianInterface}

Now, we evaluate the accuracy of the mean-curvature estimations along a Gaussian Monge patch given by

\begin{equation}
\vv{\qgs}(u,v) \doteq (u, v, \qgs(u,v)) \quad \textrm{with} \quad \qgs(u,v) = a_{gs} e^{\frac{1}{2}\left(\frac{u^2}{\sigma_u^2} + \frac{v^2}{\sigma_v^2}\right)},
\label{eq:GaussianInterface}
\end{equation}
where $a_{gs} > 0$ is the height, and the nonzero $\sigma_u^2$ and $\sigma_v^2$ parameters control the spread of the data in the $u$ and $v$ directions.  After applying the procedures from \cite{ModernDifferentialGeometry;2006} on \cref{eq:GaussianInterface}, we arrive at the analytical forms

\begin{equation}
\begin{split}
\kappa_{gs}(u,v) &= -\qgs(u,v)\frac{\qgs^2(u,v)\left(\sigma_u^2v^2 + \sigma_v^2u^2\right) + \sigma_u^4\left(\sigma_v^2-v^2\right) + \sigma_v^4\left(\sigma_u^2-u^2\right)}{\sigma_u^4\sigma_v^4\left(1 + \qgs^2(u,v)\left(\frac{u^2}{\sigma_u^4} + \frac{v^2}{\sigma_v^4}\right)\right)^{3/2}} \quad \textrm{and} \\
\kappa_{G,gs}(u,v) &= \qgs^2(u,v)\frac{\left(\sigma_u^2-u^2\right)\left(\sigma_v^2-v^2\right) - u^2v^2}{\sigma_u^4\sigma_v^4\left(1 + \qgs^2(u,v)\left(\frac{u^2}{\sigma_u^4} + \frac{v^2}{\sigma_v^4}\right)\right)^2}
\label{eq:GaussianCurvatures}
\end{split}
\end{equation}
to calculate the mean and Gaussian curvatures at any $\vv{p} = [u,\, v]^T \in \mathbb{R}^2$.  Unlike \cref{eq:EllipsoidalCurvatures,eq:ParaboloidalCurvatures}, $\kappa_{G,gs}(\cdot)$ may be positive or negative.  This property enables the non-saddle and saddle paths in \Cref{alg:MLCurvature} and thus allows us to test our solver's response when both $\mathcal{F}_\kappa^{ns}(\cdot)$ and $\mathcal{F}_\kappa^{sd}(\cdot)$ partake in the mean-curvature computations.

As before, we can use \cref{eq:GaussianInterface} to construct the level-set function

\begin{equation}
\phi_{gs}(\vv{x}) = \left\{\begin{array}{ll}
	-\dist{\vv{x}}{\vv{\qgs}(\cdot)} & \textrm{if } z > \qgs(x,y), \\
	0                                & \textrm{if } z = \qgs(x,y) \;(\textrm{i.e., } \vv{x} \in \Gamma), \\
	+\dist{\vv{x}}{\vv{\qgs}(\cdot)} & \textrm{if } z < \qgs(x,y),
\end{array}\right.
\label{eq:GaussianLevelSetFunction}
\end{equation}
where $\vv{x} \in \mathbb{R}^3$ is expressed in terms of the local coordinate frame $\mathcal{C}_{gs}$.  \Cref{eq:GaussianLevelSetFunction} partitions the computational space into $\Omega^-$ above $\Gamma \equiv \vv{\qgs}(\cdot)$ and $\Omega^+$ right beneath it.

In this experiment, we consider the shape parameters $a_{gs} = 1$, $\sigma_u^2 = \eten{1.302083}{-1}$, and $\sigma_v^2 = \eten{1.446759}{-2}$ so that the desired $h\kappa^*$ at the peak reaches $0.6$.  Also, we have ensured that the randomly affine-transformed Gaussian surface inscribed in an elliptical cylinder of semi-axes $u_{gs}^{lim} = |u_{gs}^0| + \sigma_u$ and $v_{gs}^{lim} = |v_{gs}^0| + \sigma_v$ is fully contained within the discretized $\Omega$.  Here, $u_{gs}^0$ and $v_{gs}^0$ denote the locations where $\kappa_{gs}(|u_{gs}^0|, 0) = \kappa_{gs}(0, |v_{gs}^0|) = 0$ (i.e., they define the four $\qgs(\cdot)$'s saddle points in the $u$ and $v$ directions).  We have approximated these critical points with the Newton--Raphson subroutine provided by the Boost C++ library \cite{Boost;2019}.  Furthermore, we have only accounted for stencils lying inside the limiting cylinder when sampling and gleaning error statistics to keep flat-region overrepresentation at bay.

\begin{table}[!t]
	\centering
	\small
	\bgroup
	\def\arraystretch{1.1}%
	\begin{tabular}{l|l|ccc|ccc|cc}
		\cline{3-10}
		\multicolumn{2}{c|}{} & \multicolumn{3}{c|}{MAE} & \multicolumn{3}{c|}{MaxAE} & \multicolumn{2}{c}{Performance} \\
		\hline
		Method & $\eta$ & $h\kappa$ & $\kappa$ & \makecell{Improv.\\Factor} & $h\kappa$ & $\kappa$ & \makecell{Improv.\\Factor} & \makecell{Time\\(secs.)} & \makecell{Cost\\(\%)} \\
		\hline \hline
		{\tt MLCurvature()}       & 6 & $\eten{3.3748}{-4}$ & $\eten{2.1599}{-2}$ &    - 
									  & $\eten{1.5914}{-2}$ & $1.0185$            &    - 
									  & 7.7799 &        - \\
		\hline
		\multirow{3}{*}{Baseline} & 6 & $\eten{1.2920}{-3}$ & $\eten{8.2689}{-2}$ & 3.83 
		                   			  & $\eten{1.8863}{-1}$ & $\eten{1.2072}{1}$  & 11.85 
								 	  & 6.4389 & $+20.83$ \\
								  & 7 & $\eten{1.9042}{-4}$ & $\eten{2.4374}{-2}$ & 1.13 
									  & $\eten{4.9294}{-2}$ & $6.3096$            & 6.19 
									  & 27.1916 & $-71.39$ \\
								  & 8 & $\eten{3.8896}{-5}$ & $\eten{9.9573}{-3}$ & 0.46 
									  & $\eten{8.1923}{-3}$ & $2.0972$            & 2.06 
									  & 136.3310 & $-71.39$ \\
		\hline
	\end{tabular}
	\egroup
	\caption{Error and performance statistics for the Gaussian-surface test in \Cref{subsubsec:AGaussianInterface}.}
	\label{tbl:results.gaussian.stats}
\end{table}

\begin{figure}[!t]
	\centering
	\begin{subfigure}[!t]{6cm}
		\includegraphics[width=\textwidth]{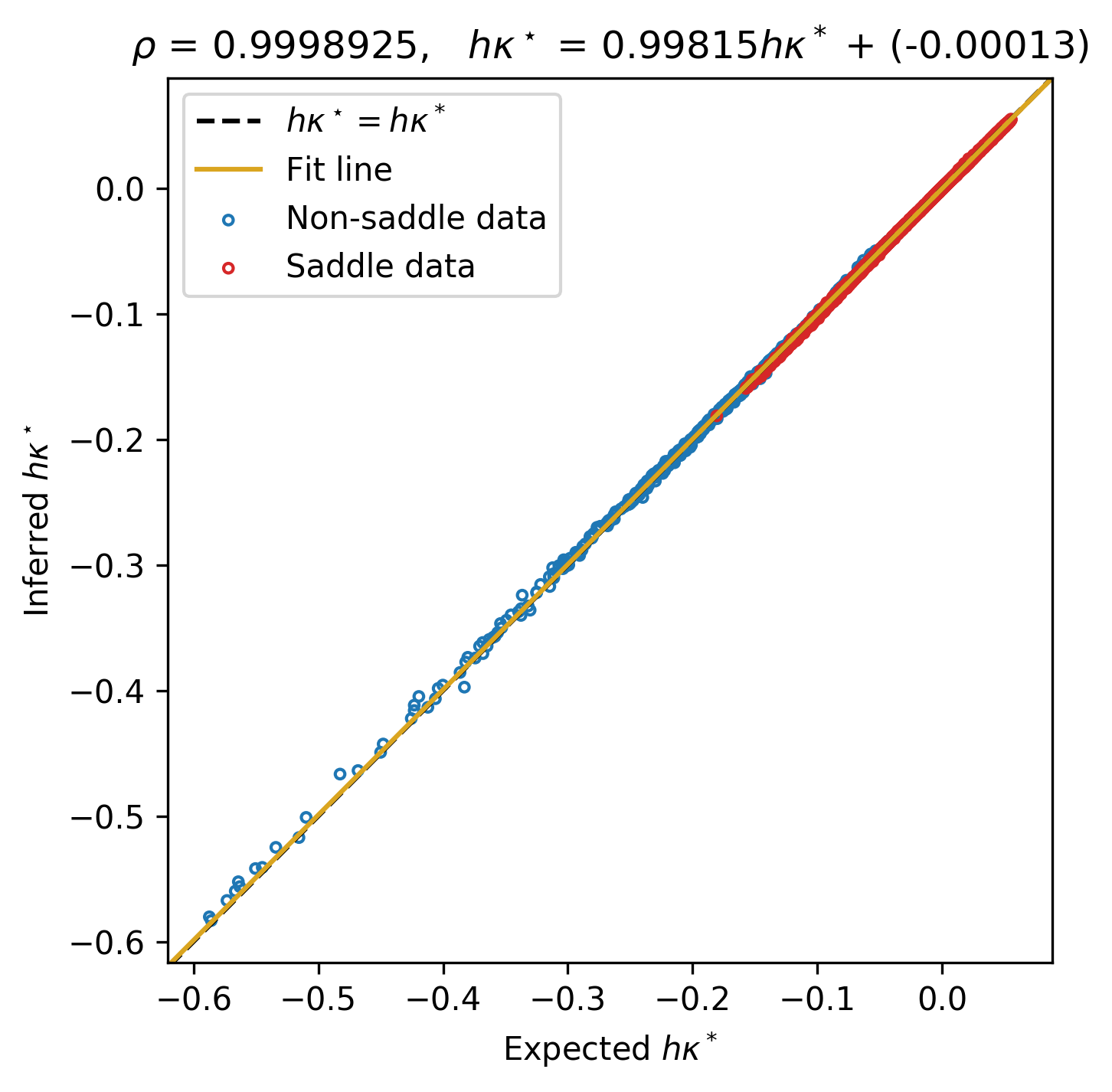}
		\caption{\footnotesize {\tt MLCurvature()} ($h = 2^{-6}$)}
		\label{fig:results.gaussian.charts.6.nnet}
	\end{subfigure}
	~
	\begin{subfigure}[!t]{6cm}
		\includegraphics[width=\textwidth]{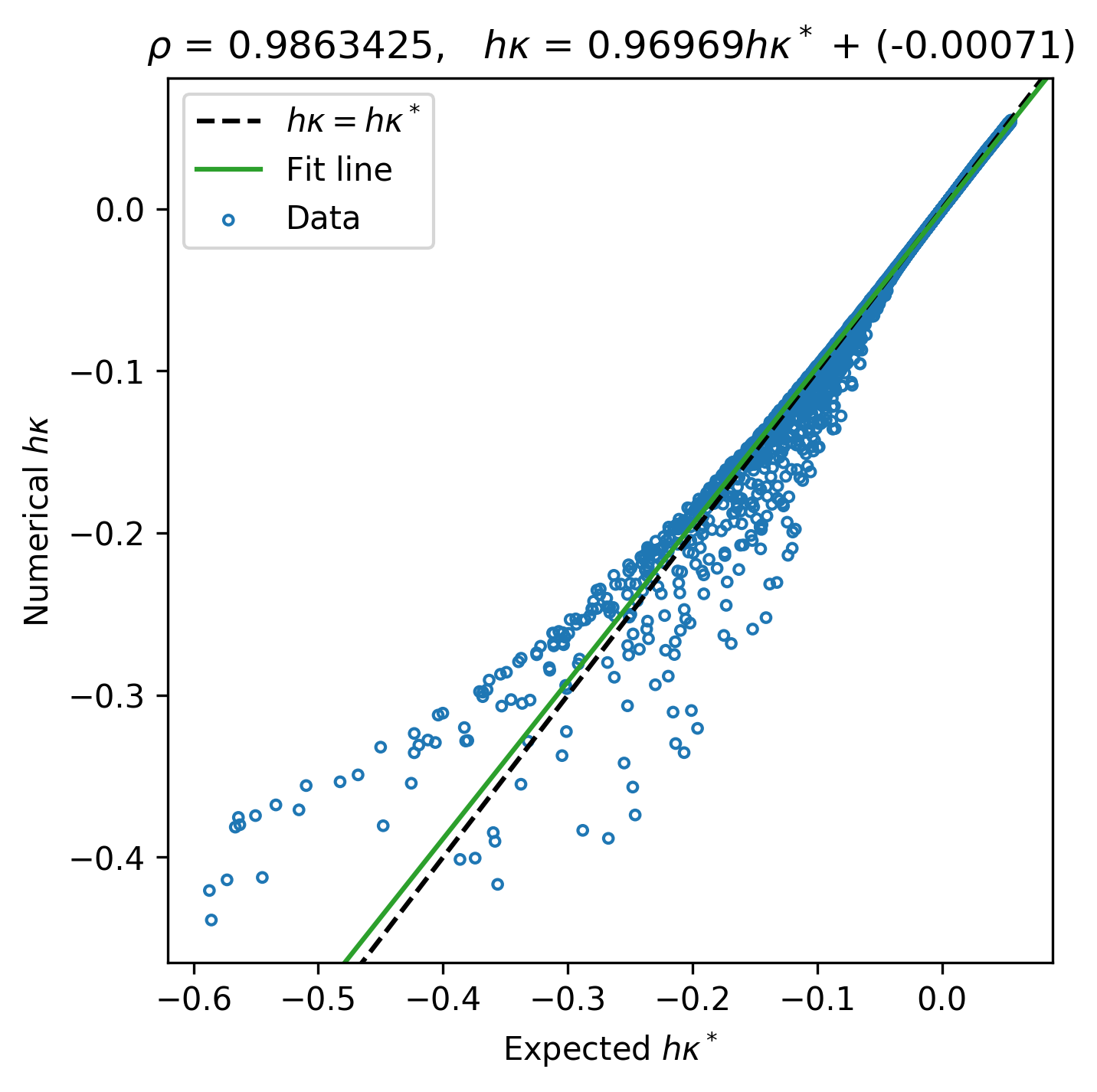}
		\caption{\footnotesize Baseline ($h = 2^{-6}$)}
		\label{fig:results.gaussian.charts.6.num}
	\end{subfigure}
	\\
	\begin{subfigure}[!t]{6cm}
		\includegraphics[width=\textwidth]{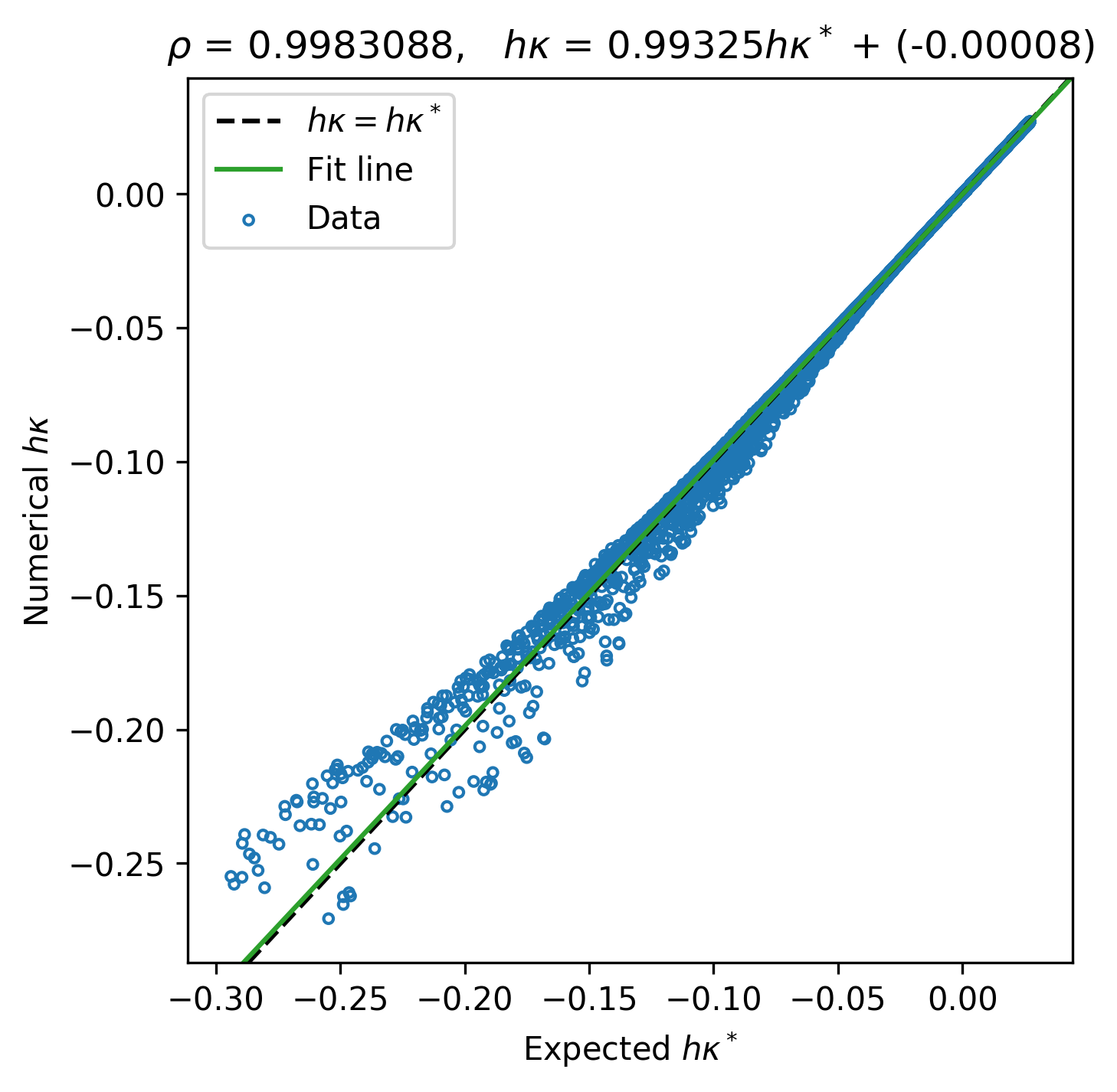}
		\caption{\footnotesize Baseline ($h = 2^{-7}$)}
		\label{fig:results.gaussian.charts.7.num}
	\end{subfigure}
	~
	\begin{subfigure}[!t]{6cm}
		\includegraphics[width=\textwidth]{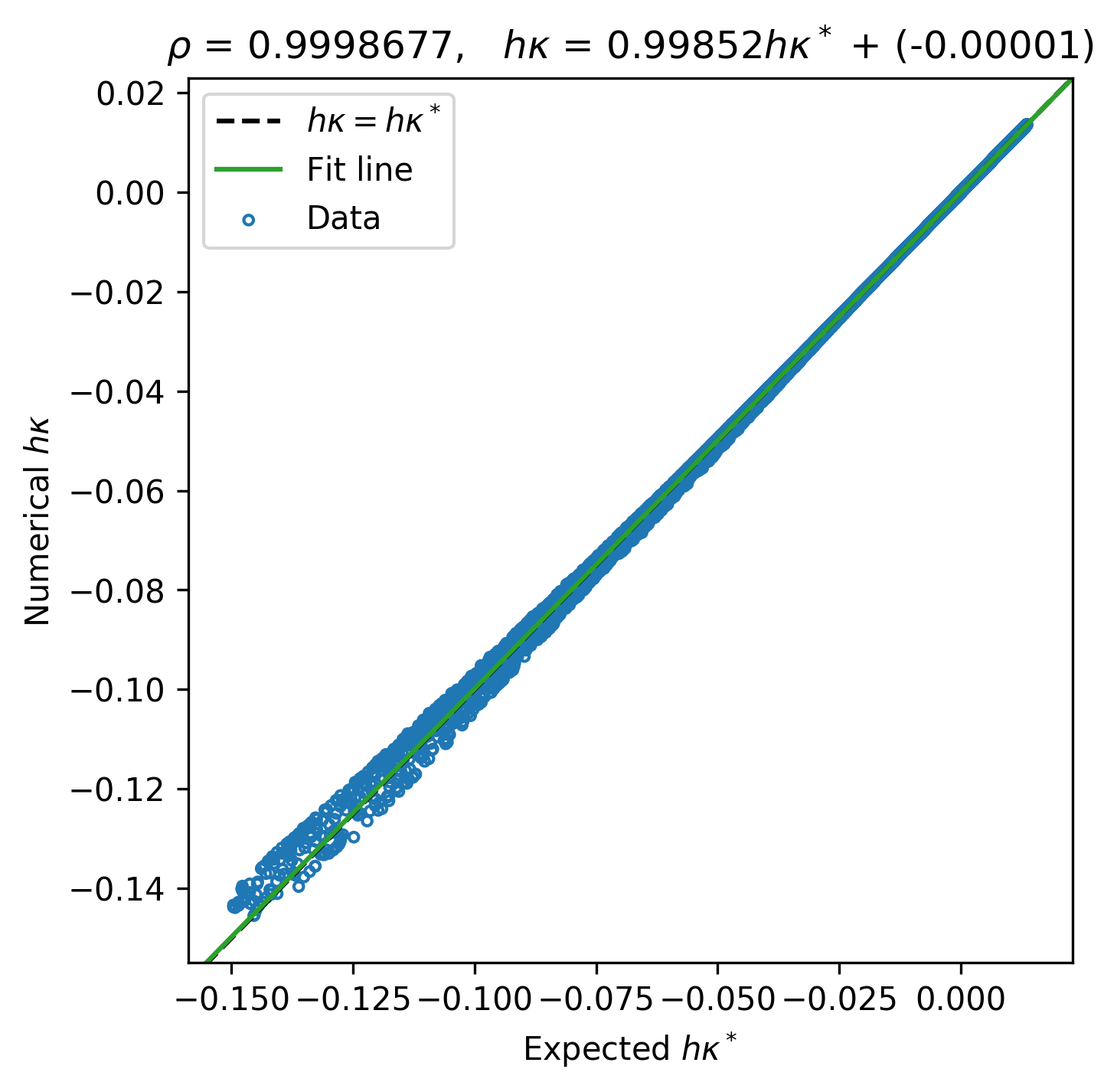}
		\caption{\footnotesize Baseline ($h = 2^{-8}$)}
		\label{fig:results.gaussian.charts.8.num}
	\end{subfigure}
	\\
	
	\caption{Correlation plots for estimated $h\kappa$ along a Gaussian surface, as described in \Cref{subsubsec:AGaussianInterface}.  We display the quality of the approximations of \Cref{alg:MLCurvature} in (a) and the numerical baseline in (b) for $\eta = 6$.  The bottom row provides the corresponding charts for the baseline at grid resolutions with (c) $\eta = 7$ and (d) $\eta = 8$.  In (a), non-saddle stencils appear in blue (\textcolor{bluematlab}{$\vv{\circ}$}) and saddle data in red (\textcolor{redmatlab}{$\vv{\circ}$}).  (Color online.)}
	\label{fig:results.Gaussian.charts}
\end{figure}

Following the above assumptions, we have summarized the results in \cref{tbl:results.gaussian.stats} for the Gaussian interface shown in \cref{fig:results.geometries.gaussian}.  These metrics reveal that our approach is again superior to the numerical baseline by at least one order of magnitude in the $L^\infty$ norm when $\eta = 6$.  Surely, one can narrow this gap by doubling the grid resolution to $\eta = 7$, but this only quadruples the cost with unsatisfactory results as $|h\kappa^*| \to 0.3$ (see \cref{fig:results.gaussian.charts.7.num}).  Likewise, reducing the mesh size to $h = 2^{-8}$ enhances the numerical precision significantly, but it raises the computational cost by 17.5 times compared to our \Cref{alg:MLCurvature} at $\eta = 6$.  Our strategy is thus more resilient to high-mean-curvature (non-saddle and saddle) regions and can fix such mean-curvature errors without increasing mesh refinement.  More importantly, training with saddle surfaces has endowed our solver with the ability to handle a larger variety of instantiated level-set stencils than those considered by other systems tailored only for non-saddle patterns \cite{VOFCurvature3DML19}.  To conclude, \cref{fig:results.Gaussian.charts} corroborates the statistics in \cref{tbl:results.gaussian.stats} and facilitates the error analysis and contrast between the baseline and the proposed method.  The improvements are remarkable.


\colorsubsubsection{A sphere morphing into an ellipsoid}
\label{subsubsec:ASphereMorphingIntoAnEllipsoid}

We conclude this series of geometrical experiments with a spherical interface morphing into an ellipsoid.  Initially, we consider a level-set function, such as \cref{eq:SphericalLevelSetFunction}, with a randomly shifted spherical zero-isosurface of radius $r_{sp} = 0.06$.  Then, we steadily interpolate from $a_{el} = b_{el} = c_{el} = r_{sp}$ to $a_{el} = 1.45$, $b_{el} = 0.51$, and $c_{el} = 0.17$ so that $\phi_{sp}(\cdot)$ transforms into $\phi_{el}(\cdot)$ in \cref{eq:EllipsoidalLevelSetFunction} after 51 steps.  In each stage, we have perturbed the nodal level-set values with random uniform noise and reinitialized $\vv{\phi}$ with ten iterations, as noted in the preamble of \Cref{subsec:GeometricalTests}.

\begin{figure}[!t]
	\centering
	\begin{subfigure}[b]{4cm}
		\includegraphics[width=\textwidth]{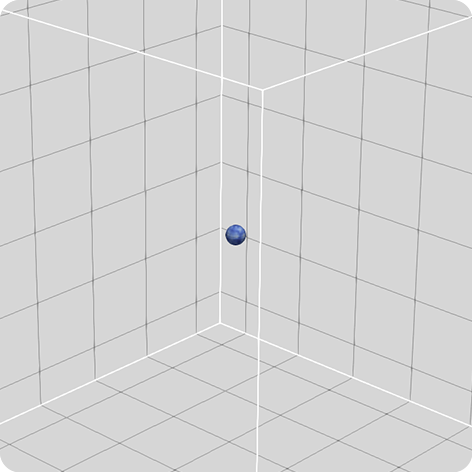}
        \caption{\footnotesize Step 0}
        \label{fig:results.morphing.step.0}
    \end{subfigure}
    ~
   	\begin{subfigure}[b]{4cm}
		\includegraphics[width=\textwidth]{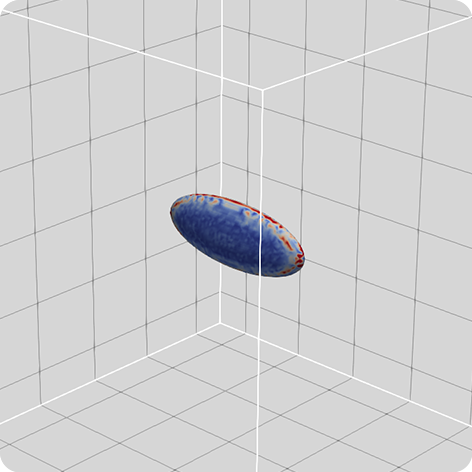}
        \caption{\footnotesize Step 17}
        \label{fig:results.morphing.step.17}
    \end{subfigure}
    ~
	\begin{subfigure}[b]{4cm}
		\includegraphics[width=\textwidth]{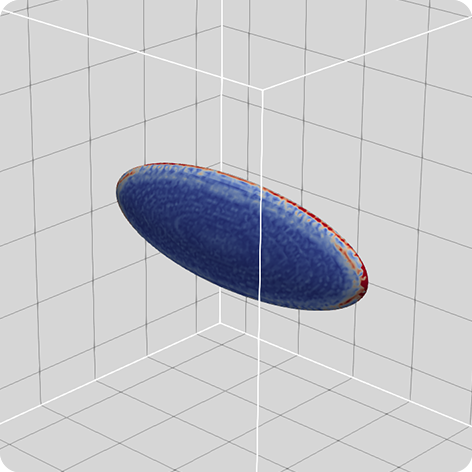}
        \caption{\footnotesize Step 34}
        \label{fig:results.morphing.step.34}
    \end{subfigure}
    ~
	\begin{subfigure}[b]{4cm}
		\includegraphics[width=\textwidth]{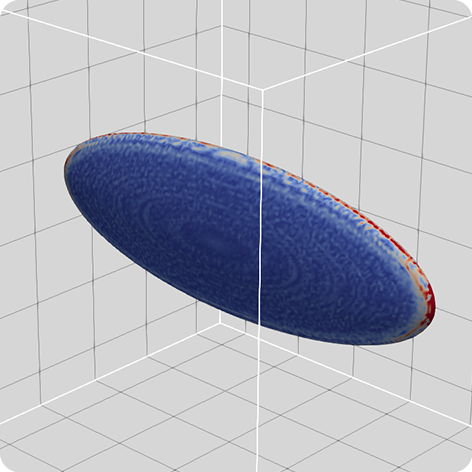}
        \caption{\footnotesize Step 51}
        \label{fig:results.morphing.step.51}
    \end{subfigure}    
	
	\begin{subfigure}[b]{8.25cm}
		\includegraphics[width=\textwidth]{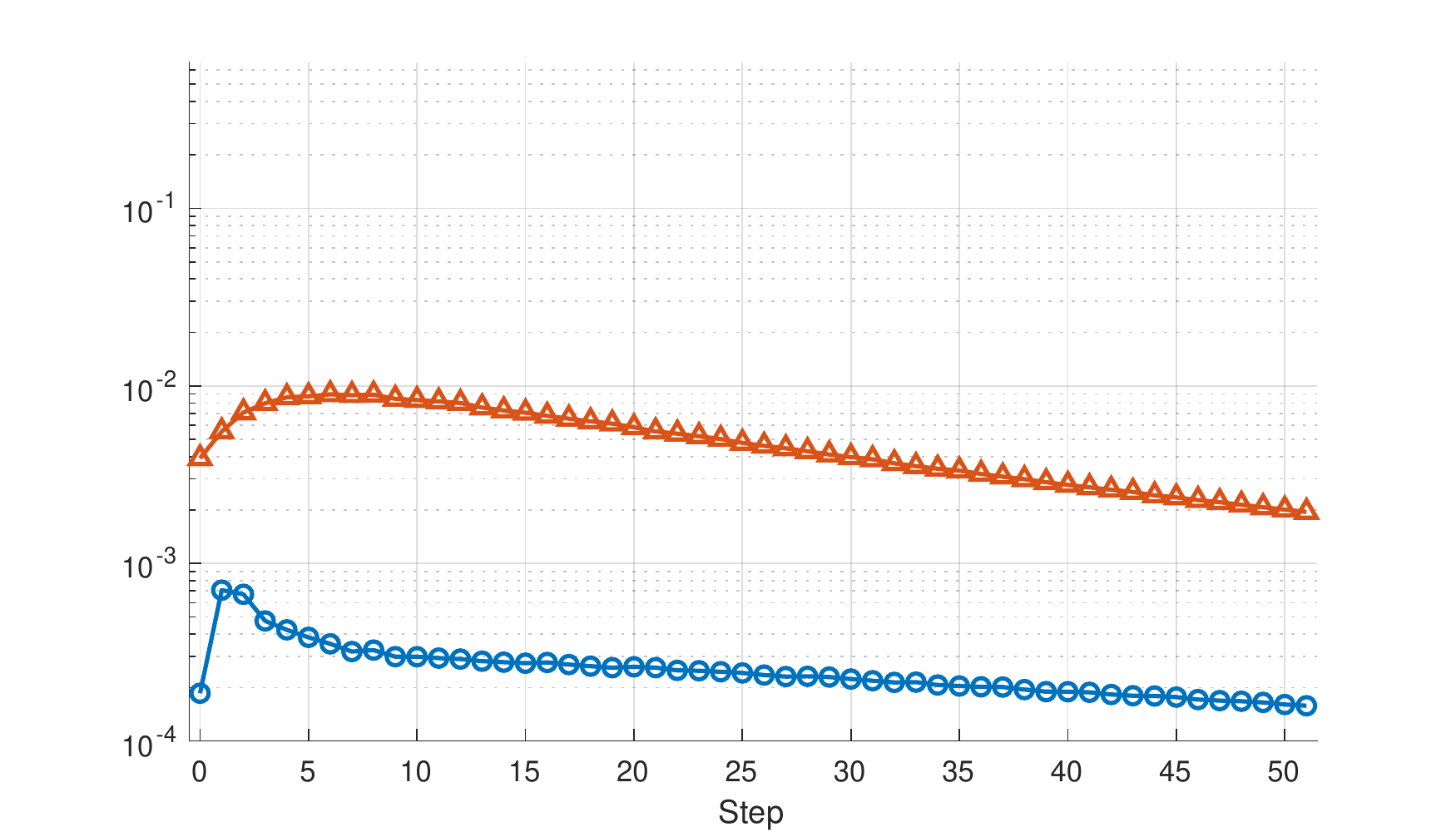}
        \caption{\footnotesize MAE}
        \label{fig:results.morphing.mae}
    \end{subfigure}
    ~
	\begin{subfigure}[b]{8.25cm}
		\includegraphics[width=\textwidth]{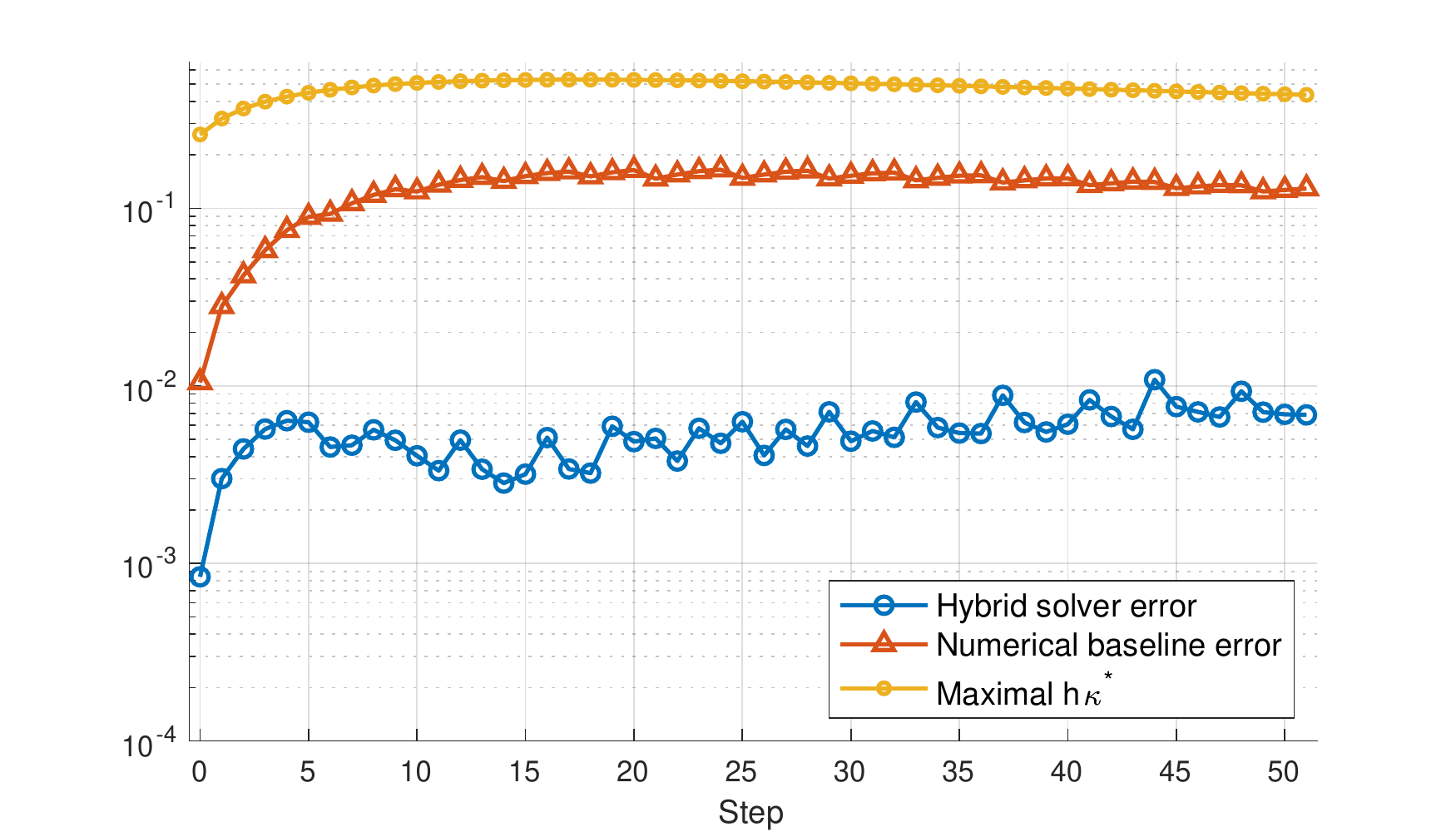}
		\caption{\footnotesize MaxAE and maximal $h\kappa^*$ per step}
		\label{fig:results.morphing.maxae}
	\end{subfigure}
	   
	\caption{A pictorial analysis of the experiment in \Cref{subsubsec:ASphereMorphingIntoAnEllipsoid}.  Top: Sphere morphing into an ellipsoid, emphasizing regions with high mean-curvature error in red.  Bottom: Evolution of the dimensionless mean-curvature $L^1$ (e) and $L^\infty$ (f) error norms.  For reference, (f) shows the maximal $h\kappa^*$ in each step.  (Color online.)}
	\label{fig:results.morphing}
\end{figure}

The top row in \cref{fig:results.morphing} depicts the evolution of the morphing surface with a few snapshots.  In particular, we have emphasized the regions where the baseline schemes exhibit significant mean-curvature errors.  \Cref{fig:results.morphing} also asserts the ability of our neural networks to reduce $\bar{\varepsilon}$ by contrasting the numerical and hybrid $L^1$ and $L^\infty$ error norms in a couple of charts.  These plots mostly demonstrate a sustained accuracy improvement of about one order of magnitude.  However, the hybrid solver's MaxAE growing trend in \cref{fig:results.morphing.maxae} reveals an underlying problem with $\mathcal{F}_\kappa^{ns}(\cdot)$ and $\mathcal{F}_\kappa^{sd}(\cdot)$ that we have not analyzed yet.  In the next section, we address this issue from the perspective of two convergence use cases.


\colorsubsection{Convergence tests}
\label{subsec:ConvergenceTests}

Next, we verify the accuracy and analyze the convergence of \Cref{alg:MLCurvature} on spherical interfaces.  The first test deals with octree-based meshes, where $h = 2^{-\eta}$ and $\eta \in \mathbb{Z}^+$.  After that, we validate our strategy on arbitrary uniform grids where the latter condition no longer holds.


\colorsubsubsection{Case 1: Relative errors on compatible adaptive grids}
\label{subsubsec:Sphere1}

First, we consider the convergence test proposed by Karnakov \etal in Section \href{https://www.sciencedirect.com/science/article/pii/S0301932219306615#sec0017}{3.1} in \cite{Karnakov;etal;HybridParticleMthdVOFCuravture;2020}.   Karnakov and coauthors have developed a hybrid mean-curvature solver for the VOF representation using strings of particles in equilibrium and height functions.  Their method has been shown to outperform the combination of height functions and parabolic fitting \cite{Popinet:09:An-accurate-adaptive} for under-resolved interfaces in multiphase flows. 

In this test, we define a spherical-interface level-set function of radius $R$, as in \cref{eq:SphericalLevelSetFunction}.  Then, we vary the number of cells per radius, $R/h$, and evaluate one hundred randomly shifted instances of the same surface.  The procedure for generating these instances is analogous to \Cref{alg:GenerateSphericalDataSet}.  Here, we extract their centers (component by component) from a uniform random distribution between $-h/2$ and $+h/2$.  Afterward, we discretize $\Omega$ with an adaptive grid $\mathcal{G}$, enforcing a uniform band of half-width $3h$ around $\Gamma$.  Next, we find the exact signed distances to $\Gamma$, reinitialize $\phi_{sp}(\cdot)$ with $\nu=10$ iterations, and compute mean curvatures along the surface.  For each $R/h = 2^i$, we have constructed $\mathcal{G}$ with one cubic octree so that $\eta = 6 + i - 1$, $h = 2^{-\eta}$, and $i = 1, 2, \dots, 5$.  The relative metrics collected for every ratio $R/h$ include

\begin{equation}
\tilde{L}^2 = \left(\frac{1}{|\mathcal{N}|} \sum_{\mathcal{n} \in \mathcal{N}}\left(\frac{\kappa - \kappa^*}{\kappa^*}\right)^2\right)^{1/2} \quad
\textrm{and} \quad
\tilde{L}^\infty = \max_{\mathcal{n} \in \mathcal{N}}\left|\frac{\kappa - \kappa^*}{\kappa^*}\right|,
\end{equation}
where $\mathcal{N}$ is the set of all interface nodes.

Incidentally, we have chosen $R = 2/64$ because it is almost the smallest sphere we can resolve for $\eta = 6$.  Furthermore, we have gathered results using the MLPs optimized for $\eta = 6$ and the models trained individually for the other resolutions.  To fit these additional estimators, we have followed the methodologies presented in \Cref{sec:Methodology} by changing $\eta$ where appropriate.  The reader may find all these neural networks at \url{https://github.com/UCSB-CASL/Curvature_ECNet_3D}.

\begin{table}[!t]
	\centering
	\small
	\bgroup
	\def\arraystretch{1.1}%
	\begin{tabular}{r|cr|cr|cr}
		\hline
		$R/h$ & \makecell{{\tt MLCurvature()}\\($\eta = 6$)} & Order & \makecell{{\tt MLCurvature()}\\($\eta = *$)} & Order & Baseline              & Order\\
		\hline \hline
		$2$   & $\eten{6.305417}{-4}$ &     - & $\eten{6.305417}{-4}$ &     - & $\eten{3.807297}{-2}$ &    - \\
		\hline
		$4$   & $\eten{9.280946}{-4}$ & -0.56 & $\eten{9.790485}{-4}$ & -0.63 & $\eten{1.569252}{-2}$ & 1.28 \\
		\hline
		$8$   & $\eten{1.065464}{-3}$ & -0.20 & $\eten{8.709580}{-4}$ &  0.17 & $\eten{5.192317}{-3}$ & 1.60 \\
		\hline
		$16$  & $\eten{1.495525}{-3}$ & -0.49 & $\eten{1.582009}{-3}$ & -0.86 & $\eten{1.548557}{-3}$ & 1.75 \\
		\hline
		$32$  & $\eten{2.471534}{-3}$ & -0.72 & $\eten{3.144518}{-3}$ & -0.99 & $\eten{4.659419}{-4}$ & 1.73 \\
		\hline
	\end{tabular}
	\egroup
	\caption{Convergence analysis on the mean-curvature $\tilde{L}^2$ error norm for the spherical-interface test in \Cref{subsubsec:Sphere1}.}
	\label{tbl:results.convergence1.rrmse}
\end{table}

\begin{table}[!t]
	\centering
	\small
	\bgroup
	\def\arraystretch{1.1}%
	\begin{tabular}{r|cr|cr|cr}
		\hline
		$R/h$ & \makecell{{\tt MLCurvature()}\\($\eta = 6$)} & Order & \makecell{{\tt MLCurvature()}\\($\eta = *$)} & Order & Baseline              & Order\\
		\hline \hline
		$2$   & $\eten{5.011141}{-3}$ &     - & $\eten{5.011141}{-3}$ &     - & $\eten{9.442257}{-2}$ &    - \\
		\hline
		$4$   & $\eten{5.728841}{-3}$ & -0.19 & $\eten{5.726814}{-3}$ & -0.19 & $\eten{4.443665}{-2}$ & 1.09 \\
		\hline
		$8$   & $\eten{4.540026}{-3}$ &  0.34 & $\eten{3.367722}{-3}$ &  0.77 & $\eten{1.411050}{-2}$ & 1.66 \\
		\hline
		$16$  & $\eten{8.826733}{-3}$ & -0.96 & $\eten{9.011447}{-3}$ & -1.42 & $\eten{3.890664}{-3}$ & 1.86 \\
		\hline
		$32$  & $\eten{2.445215}{-2}$ & -1.47 & $\eten{1.783329}{-2}$ & -0.98 & $\eten{1.636096}{-3}$ & 1.25 \\
		\hline
	\end{tabular}
	\egroup
	\caption{Convergence analysis on the mean-curvature $\tilde{L}^\infty$ error norm for the spherical-interface test in \Cref{subsubsec:Sphere1}.}
	\label{tbl:results.convergence1.rmaxae}
\end{table}

\begin{figure}[!t]
	\centering
	\begin{subfigure}[b]{5cm}
		\includegraphics[width=\textwidth]{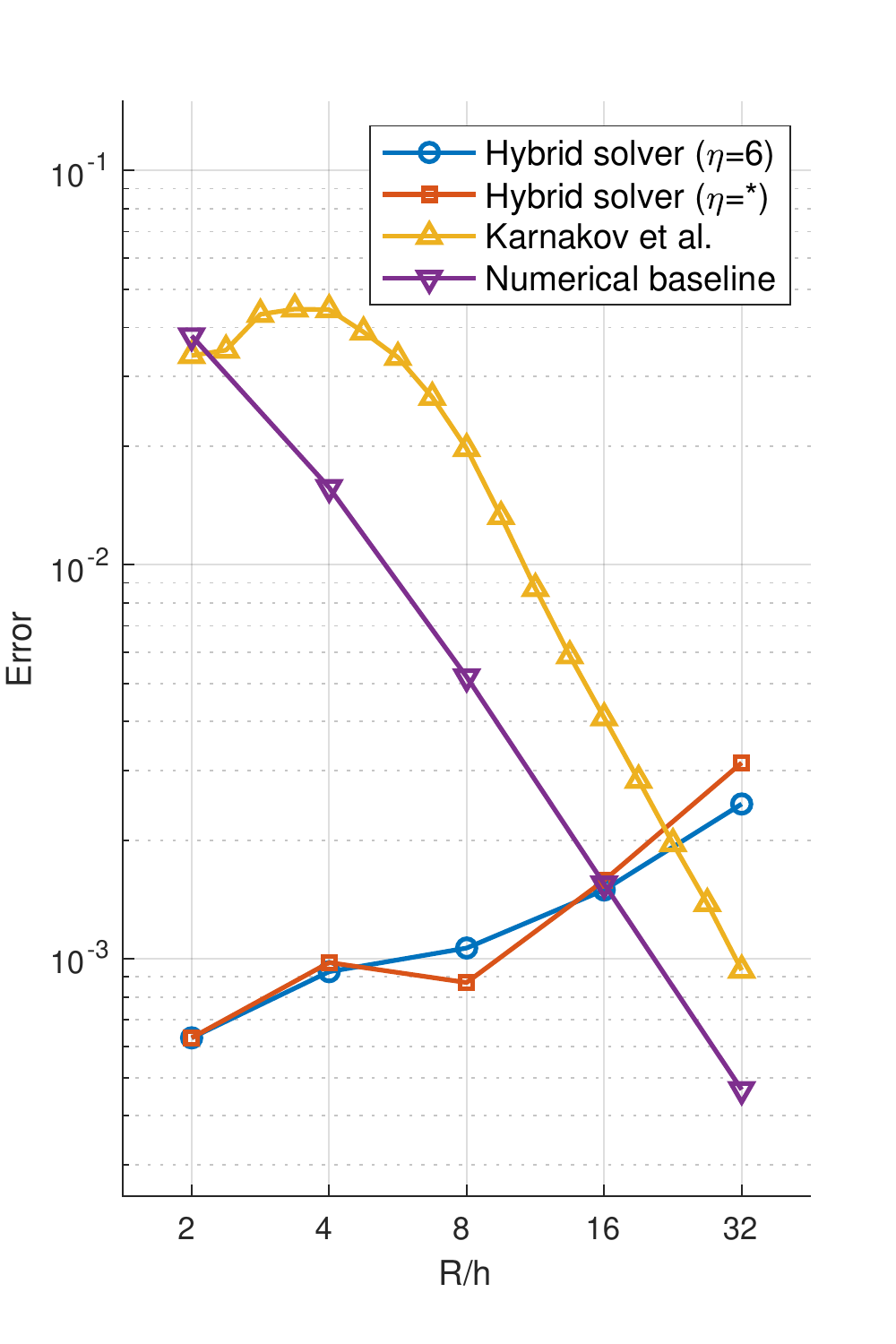}
        \caption{\footnotesize $\tilde{L}^2$}
        \label{fig:results.convergence1.accuracy.rrmse}
    \end{subfigure}
    ~
	\begin{subfigure}[b]{5cm}
		\includegraphics[width=\textwidth]{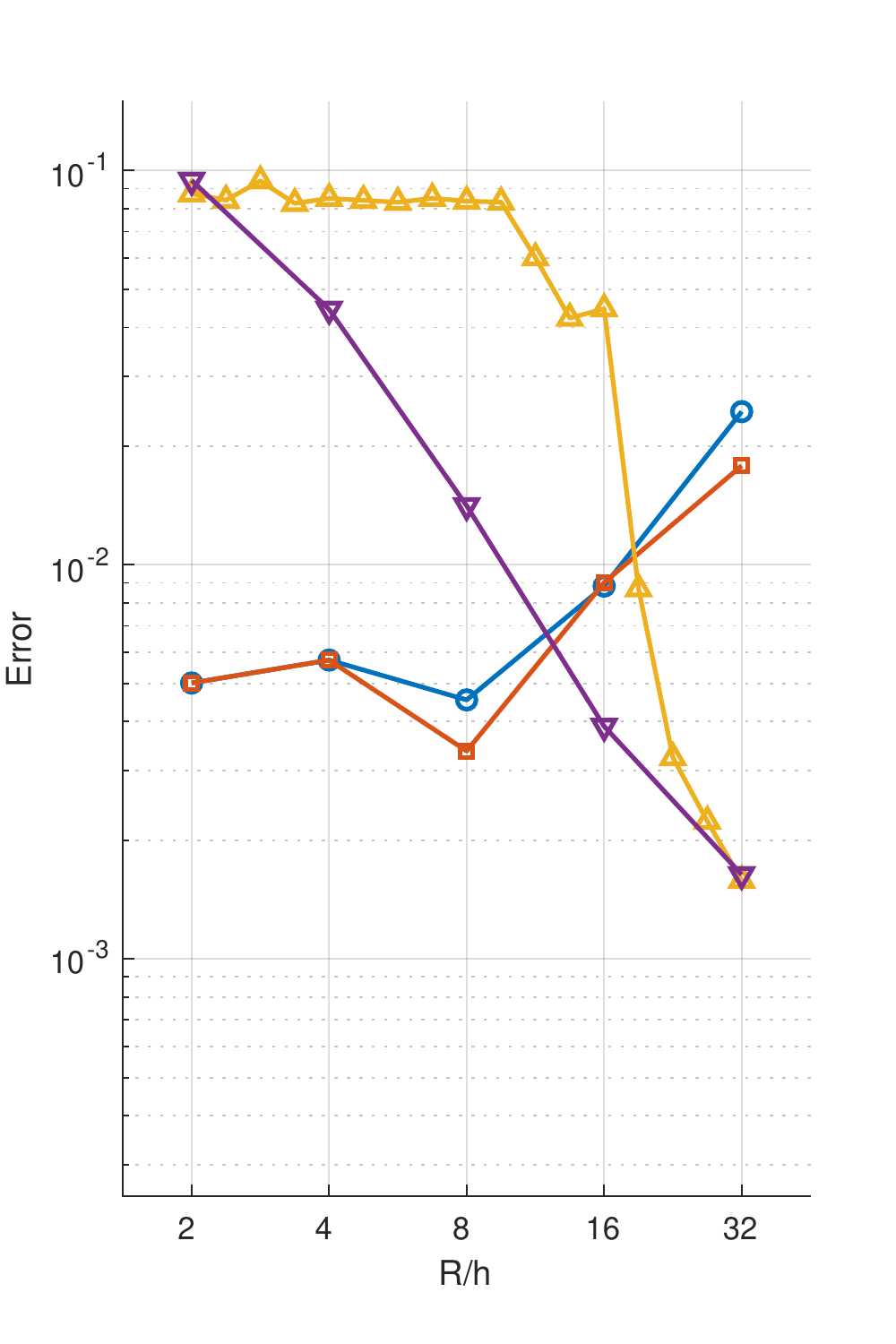}
		\caption{\footnotesize $\tilde{L}^\infty$}
		\label{fig:results.convergence1.accuracy.rmaxae}
	\end{subfigure}
	   
	\caption{Convergence analysis on the mean-curvature relative (a) RMSE and (b) MaxAE for the spherical-interface test in \Cref{subsubsec:Sphere1}.  (Color online.)}
	\label{fig:results.convergence1.accuracy}
\end{figure}

\Cref{tbl:results.convergence1.rrmse,tbl:results.convergence1.rmaxae} summarize the relative errors and the order of convergence of our hybrid inference system and the baseline.  To distinguish the neural networks used with \Cref{alg:MLCurvature}, we have labeled their training resolution with $\eta = 6$ and $\eta = *$.  The $*$ wildcard denotes matching training and testing mesh sizes.  \Cref{tbl:results.convergence1.rrmse,tbl:results.convergence1.rmaxae} and the accompanying plots in \cref{fig:results.convergence1.accuracy} reveal the superiority of our error-correcting strategy over the baseline and Karnakov's schemes along under-resolved interface sectors.  For $R/h = 2, 4$, for example, the $\tilde{L}^2$ and $\tilde{L}^\infty$ error norms can decrease at least by a factor of 7.7.  In addition, these results demonstrate that neural networks fitted to data from a prescribed $h$ can be used with negligible accuracy variations in other discretizations.  This observation is important because the methodology outlined in \Cref{sec:Methodology} can free the scientist from building data sets and training pairs of neural models for each distinct $\eta$ they plan to support.  Consequently, our strategy is more cost-effective than constructing resolution-dependent dictionaries of MLPs, as proposed in \cite{LALariosFGibou;LSCurvatureML;2021, Larios;Gibou;HybridCurvature;2021}.

\Cref{tbl:results.convergence1.rrmse,tbl:results.convergence1.rmaxae} and \cref{fig:results.convergence1.accuracy} also corroborate the lack of convergence in the machine learning solution, as noted in prior studies for $\mathbb{R}^2$ \cite{Larios;Gibou;HybridCurvature;2021, Larios;Gibou;KECNet2D;2022}.  In fact, our relative accuracy deteriorates rather quickly as $h \to 0$, thus suggesting the need for a better blending strategy between $h\kappa$ and $h\kappa_\mathcal{F}$ (see \crefrange{alg:MLCurvature.BlendStart}{alg:MLCurvature.BlendEnd} \Cref{alg:MLCurvature}).  There are several possibilities in this regard.  Karnakov and coauthors, for instance, have used conventional mechanisms to switch between their particle approximation and height functions when computing curvatures in \cite{Karnakov;etal;HybridParticleMthdVOFCuravture;2020}.  In contrast, Buhendwa \etal have resorted to neural networks to decide when to reconstruct a level-set interface via machine learning or with consistent numerical procedures \cite{Buhendwa;Bezgin;Adams;IRinLSwithML;2021}.  Enforcing consistency in our {\tt MLCurvature()} routine is a topic that deserves further investigation, possibly incorporating classifiers like those in \cite{TroubledCellIndicator18, ShockDetector20, Buhendwa;Bezgin;Adams;IRinLSwithML;2021}.

\colorsubsubsection{Case 2: Absolute errors on arbitrary uniform grids}
\label{subsubsec:Sphere2}

To conclude, we demonstrate that \Cref{alg:MLCurvature} works on uniform grids where $h$ is unconstrained.  This last test recreates the assessment in Section 5.3 in \cite{Chene;Min;Gibou:08:Second-order-accurat}, which considers the domain $\Omega \equiv [-1, +1]^3$ and the level-set field defined by

\begin{equation}
\phi_{nsp}(x, y, z) = x^2 + y^2 + z^2 - (0.2222)^2.
\label{eq:NSDSphereTestLevelSetFunction}
\end{equation}

\Cref{eq:NSDSphereTestLevelSetFunction} has a spherical interface of radius $0.2222$ centered at the origin, but it is not a signed distance function.  Consequently, we have reinitialized $\phi_{nsp}(\cdot)$ with $\nu = 80$ iterations, as in \cite{Chene;Min;Gibou:08:Second-order-accurat}, and then computed mean curvatures at $\Gamma$ with four uniform grid resolutions.  \Cref{tbl:results.convergence2.mae,tbl:results.convergence2.maxae} summarize the convergence results when using our models for $\eta = 6$ and an initial mesh with $19^3$ cubic cells.  Note we have not perturbed the level-set values or the geometry as before.  In addition, we provide our statistics in terms of $\kappa$ (not $h\kappa$).

\begin{table}[!t]
	\centering
	\small
	\bgroup
	\def\arraystretch{1.1}%
	\begin{tabular}{r|cr|cr}
		\hline
		Cell count & {\tt MLCurvature()}   & Order & Baseline              & Order\\
		\hline \hline
		$19^3$     & $\eten{1.193669}{-2}$ &     - & $\eten{1.006602}{-1}$ &    - \\
		\hline
		$38^3$     & $\eten{6.485115}{-3}$ &  0.88 & $\eten{6.141582}{-2}$ & 0.71 \\
		\hline
		$76^3$     & $\eten{6.204561}{-3}$ &  0.06 & $\eten{1.842037}{-2}$ & 1.74 \\
		\hline
		$152^3$    & $\eten{6.402527}{-3}$ & -0.05 & $\eten{5.401022}{-3}$ & 1.77 \\
		\hline
	\end{tabular}
	\egroup
	\caption{Convergence analysis on the mean-curvature $L^1$ error norm for the spherical-interface test in \Cref{subsubsec:Sphere2}.}
	\label{tbl:results.convergence2.mae}
\end{table}

\begin{table}[!t]
	\centering
	\small
	\bgroup
	\def\arraystretch{1.1}%
	\begin{tabular}{r|cr|cr}
		\hline
		Cell count & {\tt MLCurvature()}   & Order & Baseline              & Order\\
		\hline \hline
		$19^3$     & $\eten{1.891856}{-2}$ &     - & $\eten{1.876860}{-1}$ &    - \\
		\hline
		$38^3$     & $\eten{2.396634}{-2}$ & -0.34 & $\eten{1.653421}{-1}$ & 0.18 \\
		\hline
		$76^3$     & $\eten{1.681145}{-2}$ &  0.51 & $\eten{4.190414}{-2}$ & 1.98 \\
		\hline
		$152^3$    & $\eten{2.955420}{-2}$ & -0.81 & $\eten{1.356495}{-2}$ & 1.63 \\
		\hline
	\end{tabular}
	\egroup
	\caption{Convergence analysis on the mean-curvature $L^\infty$ error norm for the spherical-interface test in \Cref{subsubsec:Sphere2}.}
	\label{tbl:results.convergence2.maxae}
\end{table}

\begin{figure}[!t]
	\centering
	\begin{subfigure}[b]{5cm}
		\includegraphics[width=\textwidth]{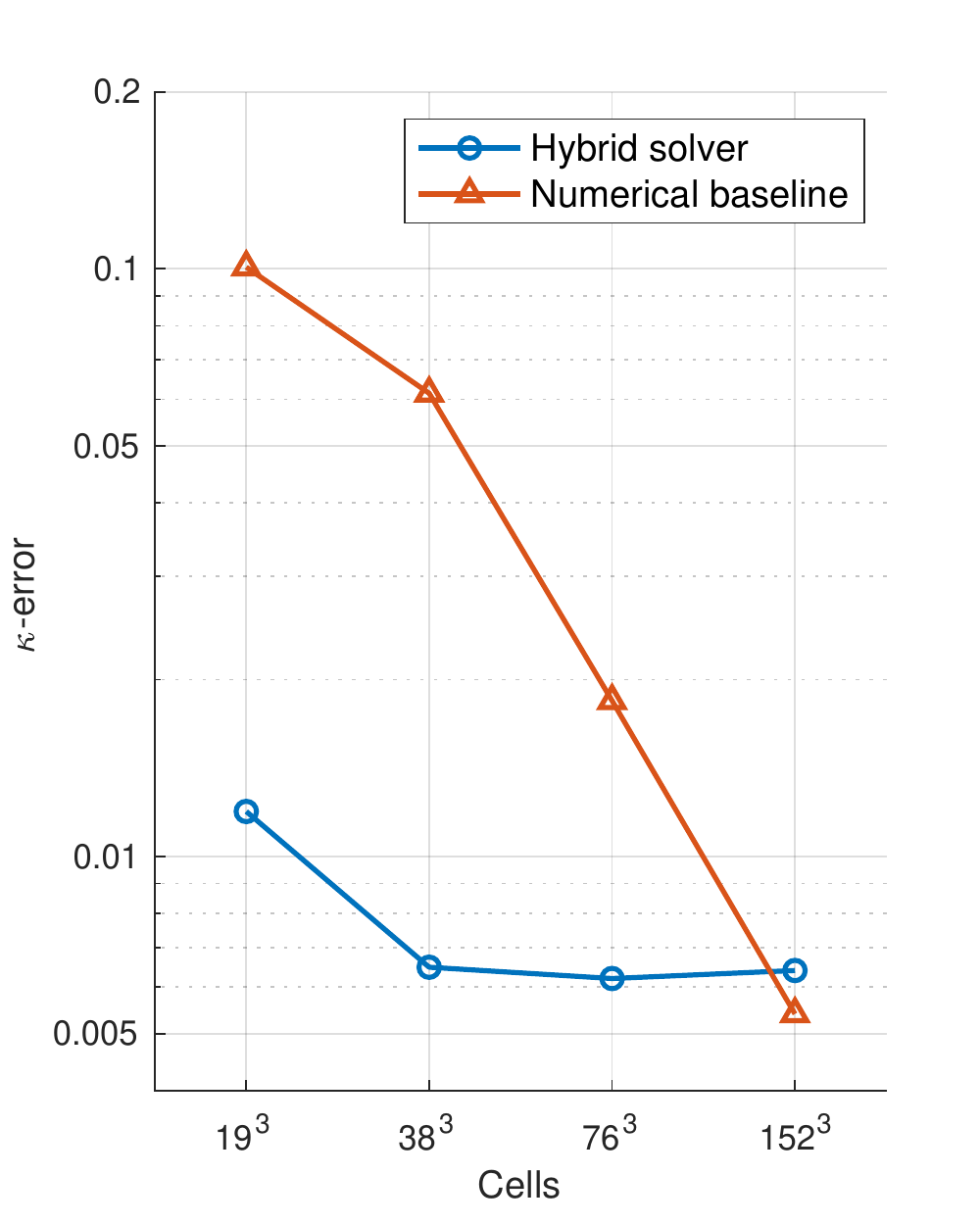}
        \caption{\footnotesize MAE}
        \label{fig:results.convergence2.accuracy.mae}
    \end{subfigure}
    ~
	\begin{subfigure}[b]{5cm}
		\includegraphics[width=\textwidth]{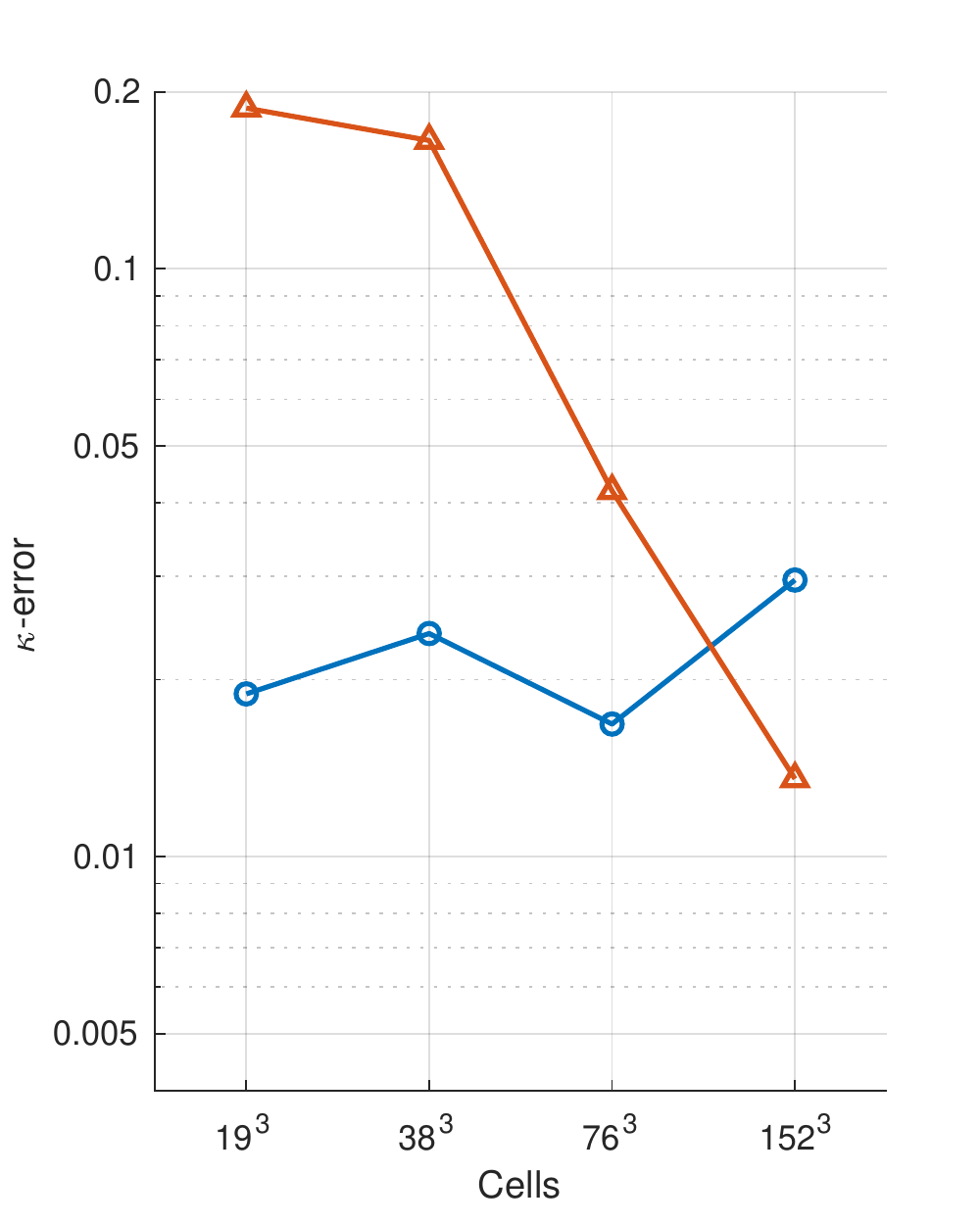}
		\caption{\footnotesize MaxAE}
		\label{fig:results.convergence2.accuracy.maxae}
	\end{subfigure}
	   
	\caption{Convergence analysis on the mean-curvature (a) $L^1$ and (b) $L^\infty$ error norms for the spherical-interface test in \Cref{subsubsec:Sphere2}.  (Color online.)}
	\label{fig:results.convergence2.accuracy}
\end{figure}

\Cref{tbl:results.convergence2.mae,tbl:results.convergence2.maxae} show that our hybrid inference system works transparently across grid resolutions after training $\mathcal{F}_\kappa^{ns}(\cdot)$ and $\mathcal{F}_\kappa^{sd}(\cdot)$.  They also confirm that our strategy is superior to the baseline when resolving steep mean curvatures.  In particular, these metrics indicate that we can improve the numerical estimations for up to one order of magnitude in the $L^1$ and $L^\infty$ norms for relatively coarse meshes.  However, these statistics and \cref{fig:results.convergence2.accuracy} reiterate that our solver is not consistent.  For a better hybrid method, we should be able to detect well-resolved stencils with reliable mean-curvature estimations.  Then, we could fall back to the standard level-set approach and converge to the correct answer as $h \to 0$.

\FloatBarrier
\colorsection{Conclusions}
\label{sec:Conclusions}

We have introduced a machine learning strategy to enhance mean-curvature computations in the level-set representation in $\mathbb{R}^3$.  Our approach entails optimizing and deploying error-quantifying neural networks that improve numerical mean-curvature estimations, $\kappa$, at the interface, regardless of the mesh size.  Unlike \cite{VOFCurvature3DML19} in the VOF framework, these models leverage local level-set and gradient, as well as other geometrical information within a novel hybrid inference system.  In line with \cite{Buhendwa;Bezgin;Adams;IRinLSwithML;2021}, our solver also enforces symmetry by blending multiple inferred $\kappa$ estimations and  the standard approximation selectively.

To minimize our solver's time-space complexity, we have tackled the mean-curvature problem from two perspectives.  These depend on whether the Gaussian curvature $\kappa_G$ associated with an interface node is (approximately) nonnegative or negative.  The former (non-saddle) case is simpler to handle \cite{Larios;Gibou;HybridCurvature;2021, Larios;Gibou;KECNet2D;2022} because its baseline error grows monotonically with $|\kappa|$.  This feature has helped increase neural capacity by training only on the negative mean-curvature spectrum.  Likewise, it has allowed us to convex-combine the machine-learning solution and baseline approximations in \Cref{alg:MLCurvature} when $\kappa \approx 0$.  Saddle patterns (i.e., $\kappa_G \lessapprox 0$), on the other hand, do not offer the same flexibility.  Hence, we have fitted separate estimators, using the full $\kappa$ spectrum for our saddle model.  Pattern separability has thus led us to design data-generative routines that yield learning tuples from spherical, sinusoidal, and hyperbolic-paraboloidal interfaces.  In this regard, \Cref{sec:Methodology} has introduced the corresponding procedures to build well-balanced data sets with a randomized strategy.  Similarly, we have presented other technical aspects to train and realize our two kinds of neural functions and port their preprocessing artifacts into level-set applications.

Experiments with stationary, steep-$\kappa$ geometries have confirmed our models' ability to fix the numerical error in the conventional level-set approach.  In particular, we have observed accuracy improvements of around one order of magnitude in the $L^\infty$ norm along under-resolved interface regions.  Also, we have shown that our proposed solver can estimate sharp mean curvatures at an unparalleled level of precision, even if one doubles the mesh resolution to reduce the baseline mean-curvature error.  Our relative accuracy, however, can deteriorate as stencils become more well-resolved.  On this matter, the convergence tests in the second half of \Cref{sec:Results} have underlined the lack of consistency in \Cref{alg:MLCurvature} as $h \to 0$.  Making the {\tt MLCurvature()} routine consistent is thus the natural next step in our research.  Although various possibilities exist, we have narrowed down this problem to identifying well-resolved patterns so that we can fall back to the standard mean-curvature estimations when appropriate.  Such an identification function could be as simple as a quality metric \cite{Macklin;Lowengrub:05:Evolving-interfaces-, Macklin;Lowengrub;ImprovedCurvatureAppTumorGrowth;2006, Lervag;CalcCurvatureLSM;2014, Ervik;Lervag;Munkejord;LOLEX;2014} or as sophisticated as a neural binary classifier \cite{TroubledCellIndicator18, ShockDetector20, Buhendwa;Bezgin;Adams;IRinLSwithML;2021}.

To conclude, we comment on a potential enhancement to the data-generating strategy from \Crefrange{subsubsec:SphericalInterfaceDataSetConstruction}{subsubsec:HypParaboloidalInterfaceDataSetConstruction}.  Although comprehensive, our learning data sets are only well-balanced with respect to $h\kappa^*$.  Other latent distributions, like those of $h\kappa$ and $\bar{\varepsilon}$, could be heavily skewed and impair generalization.  For this reason, we have entertained the possibility of using two-dimensional histograms to level up sample frequencies along $h\kappa^*$ and either $|h\kappa - h\kappa^*|$ or $h\kappa$.  Then, one would expect more robust neural responses across broader error and mean-curvature spectra.


\sffamily\color{navy}\section*{Acknowledgements}\color{black}\rmfamily

This research was funded by ONR MURI N00014-11-1-0027 and NSF CBET-EPSRC 2054894.  This work used the Extreme Science and Engineering Discovery Environment (XSEDE), which is supported by National Science Foundation grant number ACI-1053575, for resources at the \href{http://www.tacc.utexas.edu}{Texas Advanced Computing Center} (TACC).



{\footnotesize
\biboptions{sort&compress}
\bibliographystyle{unsrt}
\bibliography{references}}

\end{document}